\documentclass{article}

\usepackage[preprint]{neurips_2026}

\usepackage[utf8]{inputenc} % allow utf-8 input
\usepackage[T1]{fontenc}    % use 8-bit T1 fonts
\usepackage[colorlinks]{hyperref}
\hypersetup{
 colorlinks=True,
 linkcolor=nhblue,
 citecolor=gray,
 urlcolor=nhblue}
\usepackage{url}            % simple URL typesetting
\usepackage{booktabs}       % professional-quality tables
\usepackage{amsfonts}       % blackboard math symbols
\usepackage{nicefrac}       % compact symbols for 1/2, etc.
\usepackage{microtype}      % microtypography
\usepackage{xcolor}         % colors
\usepackage{amsmath}
\usepackage{graphicx}
\usepackage{xcolor}
\usepackage[format=plain,
            labelfont=it,
            labelsep=period]{caption}
\usepackage{makecell}
\usepackage[most]{tcolorbox}

\newtcolorbox{takeawaybox}{
  enhanced,
  colback=bmcorange!10,
  colframe=bmcorange!80!black,
  boxrule=0.7pt,
  arc=2pt,
  left=7pt,
  right=7pt,
  top=8pt,
  bottom=7pt,
  parskip=2pt,
  title=Key Takeaways,
  coltitle=white,
  colbacktitle=bmcorange!85!black,
  fonttitle=\bfseries,
  attach boxed title to top left={xshift=8pt, yshift=-2.5mm},
  boxed title style={
    colback=bmcorange!85!black,
    colframe=bmcorange!85!black,
    arc=2pt,
    boxrule=0pt,
    left=6pt,
    right=6pt,
    top=2pt,
    bottom=2pt
  }
}

\definecolor{dmred}{HTML}{CE2029}
\definecolor{dmyellow}{HTML}{FFB70B} % for marking
\definecolor{nhblue}{HTML}{28679E}

\definecolor{uniformgray}{HTML}{8C8C8C}
\definecolor{dynamicblue}{HTML}{5A8DEE}
\definecolor{thompsonpurple}{HTML}{8172B2}
\definecolor{treered}{HTML}{C44E52}
\definecolor{bmcorange}{HTML}{DD8452}

\newcommand{\colorbullet}[1]{\indent\textcolor{#1}{$\bullet$}}

\title{Manifold Bandits:\\Bayesian Curriculum Learning over the Latent Geometry of Large Language Models}

\author{
Darrien McKenzie$^{1}$ \quad Nicklas Hansen$^{1,\dagger}$ \quad Xiaolong Wang$^{1,\dagger}$\\
$^{1}$University of California, San Diego \quad $^{\dagger}$Equal advising.\\
\texttt{\href{https://darrienmckenzie.com/manifold-bandits}{darrienmckenzie.com/manifold-bandits}}
}

\begin{document}

\maketitle
\begin{figure}[!h]
    \centering
    \vspace{-0.24in} 
    \includegraphics[width=0.84\textwidth]{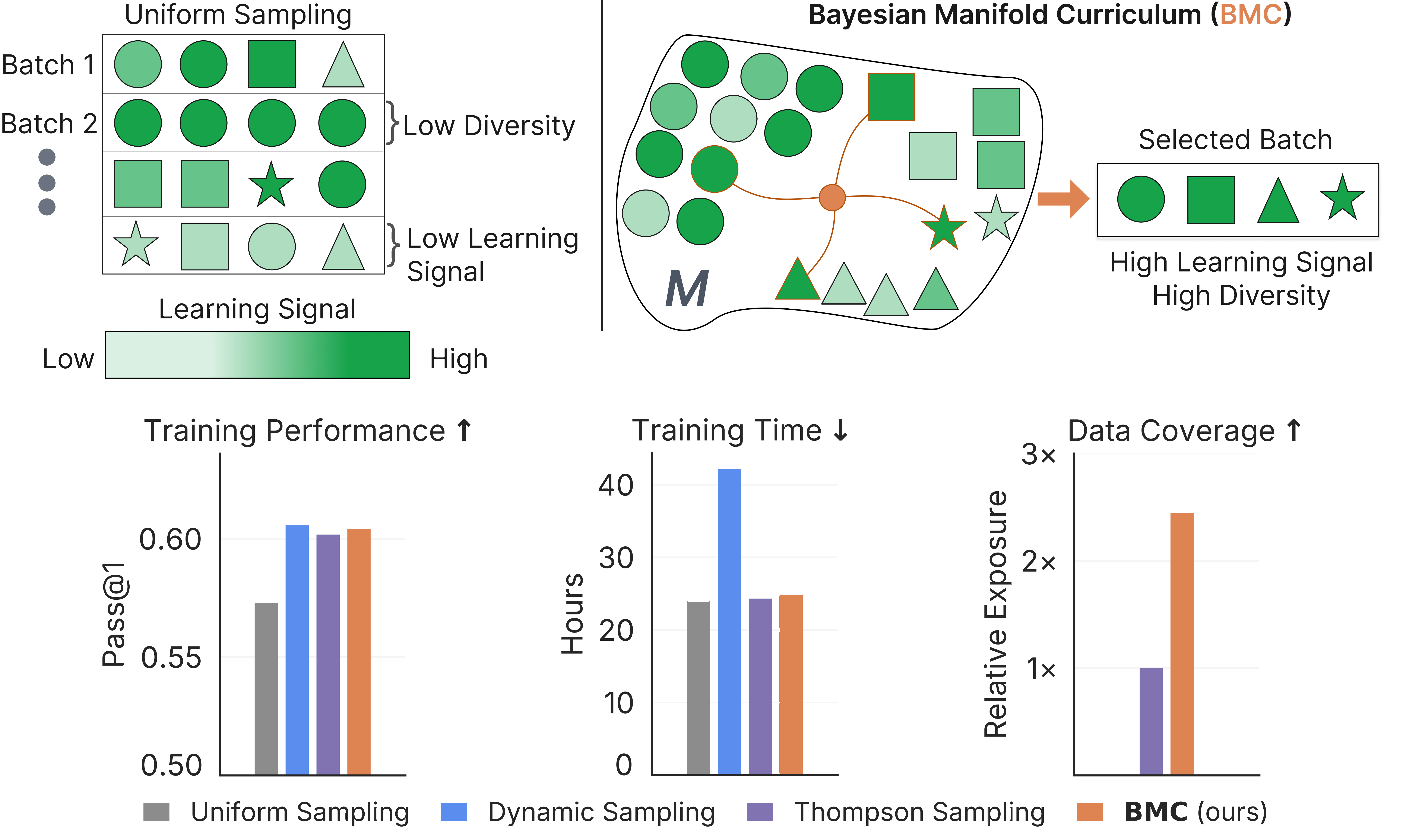}
    \caption{\textbf{Bayesian Manifold Curriculum (BMC).}
    BMC uses Bayesian learning over the policy model's latent task manifold to construct training batches from productive and diverse problem types. BMC improves the trade-off between training efficiency and broad data coverage.}

    \label{fig:teaser}
    %\vspace{0.1in}
\end{figure}

\begin{abstract}
Reinforcement learning (RL) is a central approach for improving reasoning capabilities in large language models (LLMs), where training efficiency depends critically on how problems are sampled during optimization. Existing adaptive curriculum learning methods typically prioritize prompts of intermediate difficulty, treating problem selection as a standard bandit problem with independent arms and overlooking the structured, heterogeneous nature of the task space. In this work, we frame problem sampling as a manifold-structured bandit problem with endogenous non-stationarity: problems are related through the model's latent representation space, and sampling decisions can steer how learning signals evolve across that space. To operationalize this perspective, we introduce \textbf{Bayesian Manifold Curriculum (BMC)}, a structure-aware framework that organizes problems into a hierarchical task tree and applies Bayesian learning to guide sampling. Empirically, we find that different sampling strategies induce non-trivial tradeoffs between \emph{productivity} (learning signal), \emph{diversity} (coverage of the task manifold), and \emph{utility} (evaluation relevance). These results show that prioritizing difficulty alone is insufficient for strong downstream performance, highlighting the importance of incorporating structure and type-awareness into problem sampling.
\end{abstract}

\section{Introduction}
Reinforcement Learning (RL) has emerged as a powerful paradigm for improving the reasoning capabilities of large language models (LLMs). Following the seminal work of DeepSeek-R1 \citep{deepseek_r1_2025_pdf}, a dominant class of RL algorithms for training LLMs consists of group-relative optimization methods, including GRPO \citep{shao2024deepseekmath}, DAPO \citep{yu2025dapo}, and GSPO \citep{zheng2025group}. These methods estimate policy improvements by comparing multiple rollouts generated from the same prompt and assigning credit based on relative performance across trajectories. Unlike supervised learning, where each training example typically contributes a direct learning signal, group-relative methods rely critically on variability between rollout outcomes. When rollouts produce identical rewards, the relative advantage collapses to zero, yielding no policy gradient signal. As a result, training efficiency depends strongly on presenting prompts that induce meaningful reward variation \citep{foster2025learning}. Prompts that are either consistently solved or consistently failed by the model contribute no learning signal, leading to wasted computation.

To address this challenge, recent work has explored adaptive sampling strategies designed to prioritize prompts that yield informative reward variation. One of the most widely used approaches, Dynamic Sampling, introduced in DAPO \citep{yu2025dapo}, regenerates prompts until batches contain only examples with non-zero reward variance. While effective at increasing learning signal, this strategy incurs substantial wall-clock overhead due to repeated batch regeneration. To mitigate this overhead, subsequent work explores \emph{adaptive curriculum learning} methods that attempt to predict which prompts are most likely to yield informative learning signals as the model evolves \citep{chen2025self, wang2025dump, qu2025can, zeng2025cures}. More formally, these methods tend to either explicitly or implicitly frame optimal batch formation as a \emph{non-stationary multi-arm bandit problem}, where the goal is to select arms (individual prompts) that maximize expected policy improvement.

While the standard non-stationary bandit framing is principled and has shown promise, it does not fully capture several key aspects of LLM training dynamics. In this setting, non-stationarity is not purely exogenous (i.e., external to the model): as the model improves, the reward distribution over prompts evolves as a direct consequence of the model's own updates and sampling decisions. This induces an inherently \emph{endogenous} feedback loop, where the choice of arms actively shapes the future reward landscape. Moreover, problems are not independent. Problems of similar type often share reasoning patterns or other underlying features, implying that observations from one prompt can inform beliefs about others. Ignoring these relationships can lead to inefficient exploration. 

Effective training also requires balancing not only difficulty, but also diversity. Training datasets, even within a single domain, are highly heterogeneous and often conceptually imbalanced. Focusing solely on prompts that maximize immediate learning signal, without accounting for diversity, can therefore lead to narrow coverage of the problem space, potentially harming generalization. While some approaches group problems by type or difficulty, these groupings are typically defined externally (e.g., via human annotation) and may not align with the model's internal representation of the problem space. Ideally, curriculum strategies should not only adapt to the model's evolving notion of difficulty, but also to its latent organization of the problem space.

To leverage these dynamics, we frame optimal problem sampling as a latent manifold-structured bandit problem: sampling decisions act as approximate interventions whose effects can propagate across related problems and shape future reward distributions. To operationalize 
this perspective, we introduce three key contributions:

\begin{itemize}
    \item \textbf{Latent Task Trees.} A hierarchical method for approximating the structure of the task manifold directly from model embeddings, producing a multi-scale representation of relationships between prompts.

    \item \textbf{Bayesian Manifold Curriculum (BMC).} A structured adaptive curriculum learning method that performs Bayesian decision-making over Latent Task Trees, enabling efficient exploration while accounting for non-stationary, model-induced dynamics.

    \item \textbf{Productivity--Diversity--Utility Trade-offs.} A diagnostic analysis showing that adaptive curricula cannot be fully characterized by downstream evaluation performance alone. We distinguish three axes of problem selection: productivity, diversity, and utility, and show that these axes are not interchangeable. This motivates \textbf{BMC-T}, a utility-aware extension that biases sampling toward target-relevant regions while preserving BMC's structure-aware exploration.
\end{itemize}

\section{Preliminaries}

\textbf{Reinforcement Learning for LLMs.}
Reinforcement Learning from Verifiable Rewards (RLVR) frames the training of an autoregressive LLM as a reinforcement learning problem: a policy $\pi_\theta$ maps each prompt $x$ to a sampled response $y=(y_1,\dots,y_T)$, which is evaluated by an automatic verifier to produce a scalar reward $r(x,y)$. Training seeks to maximize expected reward over the prompt distribution, optionally with additional regularization or constraints on policy updates.

\textbf{Group-Relative Policy Optimization.}
A widely used class of RL methods for LLMs consists of \emph{group-relative} policy optimization algorithms, including GRPO~\citep{shao2024deepseekmath}, DAPO~\citep{yu2025dapo}, and GSPO~\citep{zheng2025group}. For each prompt, a group of $G$ responses $\{y_i\}_{i=1}^G \sim \pi_\theta(\cdot \mid x)$ is sampled and evaluated, and a contrastive advantage $\hat{A}_i$ is constructed using only within-group statistics, such as subtracting the group mean. This formulation avoids the need for a learned critic and instead updates the policy based on \emph{relative} differences among responses conditioned on the same prompt. Recent variants retain the group-relative formulation while modifying details such as clipping, importance weighting, or length normalization to improve training stability.

Since these methods rely on reward variation within a group, prompts whose rollouts exhibit zero reward variance contribute no learning signal. This motivates sampling procedures that prioritize prompts likely to produce informative reward variance across rollouts. One prominent example is Dynamic Sampling, introduced in DAPO~\citep{yu2025dapo}, which regenerates prompts until sufficient reward variation is obtained, at the cost of increased wall-clock time.

\textbf{Structured, Causal, and Non-Stationary Bandits.}
In the classical multi-armed bandit setting, each arm is treated independently, and the learner estimates its reward distribution in isolation. \textbf{Structured bandits} relax this assumption by exploiting relationships between arms, for example by assuming that rewards vary smoothly with respect to arm features or embeddings \citep{lattimore_szepesvari_2020, slivkins2024introductionmultiarmedbandits}. When such structure is present, observations from one arm can inform beliefs about related arms, improving sample efficiency over independent-arm formulations.

A second relevant class is \textbf{causal bandits}, where arms are interpreted as interventions and the learner exploits a causal model to reason about how actions affect rewards~\citep{lattimore2016causalbanditslearninggood}. This perspective is relevant because prompt sampling in RL training is not merely a passive measurement: sampled prompts induce policy updates, which can change future reward distributions over related prompts.

Our setting combines these ideas with learning dynamics. Related problems can share information, as in structured bandits; sampling decisions are intervention-like, as in causal bandits; and reward distributions evolve over time, as in non-stationary bandits. The key distinction is that this non-stationarity is \emph{endogenous}: it is partly caused by the learner's own sampling decisions. We argue that the underlying latent task structure connects these phenomena: it provides a prior over where information and learning effects are likely to transfer across the task space.

\textbf{Latent Geometry and the Manifold Hypothesis.}
Representations learned by large language models exhibit structured organization in high-dimensional space, often well approximated by lower-dimensional geometric structure~\citep{valeriani2023geometry, modell2025originsrepresentationmanifoldslarge}; see \hyperref[app:manifold_representations]{Appendix~\ref*{app:manifold_representations}} for additional related work and discussion. This perspective is related to the manifold hypothesis, which posits that complex data distributions concentrate near lower-dimensional manifolds embedded in high-dimensional spaces~\citep{fefferman2013testingmanifoldhypothesis}.

\section{Latent Task Trees}

We first approximate the policy's latent task manifold with a hierarchy over prompts.
Intuitively, problem types are not flat: broad categories such as mathematics, coding,
or science can often be decomposed into progressively finer subtypes. Rather than treating each problem as an independent arm, we use the policy model's own representations to organize prompts into regions that reflect how the model
perceives the task space. The guiding intuition is that representational similarity should partly predict behavioral similarity: prompts that are nearby in the policy's latent space are more likely to share solution strategies, failure modes, and dynamics relevant to learning signal. The resulting hierarchy provides the scaffold on which BMC performs structured sampling and information sharing.

The need for a hierarchy also arises geometrically: LLM representation spaces are arguably not well described by a single global partition or parametric model. Their local structure can vary substantially across the data distribution, with different regions exhibiting different densities, intrinsic dimensions, and modes of organization~\citep{li2025unraveling, shafran2026directionsregionsdecomposingactivations}. We therefore construct a \emph{Latent Task Tree} using recursive local manifold approximation: each region is analyzed for remaining substructure and subdivided only when finer organization can be identified. This yields a multi-scale, computationally tractable representation of the latent task manifold that preserves local structure without imposing a fixed global partition or smooth global manifold assumption.

\begin{figure}[t]
  \centering
  \includegraphics[width=\textwidth]{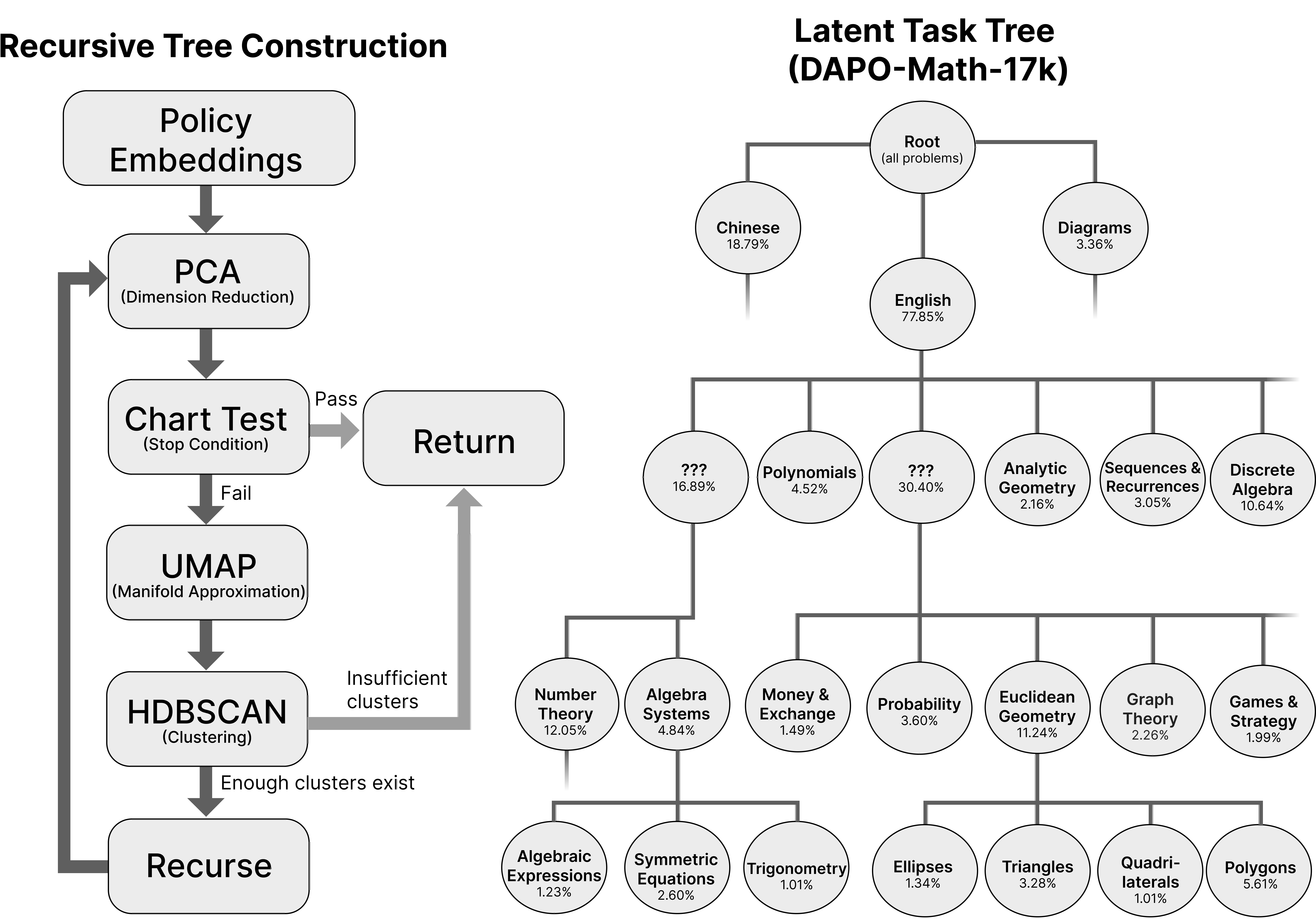}
  \caption{\textbf{Latent Task Tree construction and example structure.}
  \textbf{Left:} We construct the Latent Task Tree recursively from LLM (policy) embeddings. At each node, embeddings are reduced with PCA, tested for approximate chart-like structure, projected with UMAP, and clustered with HDBSCAN; recursion continues only when meaningful substructure remains.
  \textbf{Right:} A (partial) Latent Task Tree induced from \texttt{DAPO-Math-17K}, showing that the policy's latent representation space organizes prompts into fine-grained and imbalanced problem types. Node percentages indicate the fraction of the dataset assigned to each node. Full tree visualization is provided in \hyperref[fig:treebreakdown_dapomath]{Figure~\ref*{fig:treebreakdown_dapomath}}.}
  \label{fig:ltt}
\end{figure}

The construction proceeds by first extracting latent embeddings for each prompt using a forward pass of the policy model. Motivated by prior work suggesting that intermediate layers encode richer semantic and reasoning-related structure~\citep{valeriani2023geometry, zhu2025surveylatentreasoning}, we compute a normalized mean embedding from a selected intermediate layer for each prompt; see \hyperref[app:layer_heterogeneity]{Appendix~\ref*{app:layer_heterogeneity}} for additional discussion of layer heterogeneity. These embeddings serve as a representation of how the policy internally perceives each problem and form the basis for tree construction. Although problems could be represented in other ways, such as through external embedding models, model-generated topic labels, and sparse autoencoders, we use \emph{policy embeddings} to align the tree with the model being trained; see \hyperref[app:external-models]{Appendix~\ref*{app:external-models}} for further discussion. All prompts are initially assigned to the root node.

At each recursive step, the embeddings associated with the current node are standardized and reduced using principal component analysis (PCA). Rather than fixing the number of components, we select the smallest number that explains 95\% of the variance, allowing different regions of the dataset to be represented at different effective dimensionalities. The resulting projections are then mapped using Uniform Manifold Approximation and Projection (UMAP) \citep{mcinnes2018umap} to obtain a low-dimensional approximation of the local manifold structure. Clustering is performed on the resulting representations using HDBSCAN \citep{campello2013hdbscan}, which does not require specifying the number of clusters and is well-suited to data with varying density and non-convex structure. Each detected cluster is added as a child node, and the procedure is recursively applied.

Rather than imposing a fixed depth, recursion is governed by multiple stopping conditions. The most geometry-aware of these is a \emph{Chart Test}, motivated by the intuition that a globally complex manifold should become simple when viewed locally: once a region is small and coherent enough, its points should behave like a single smooth patch rather than several distinct subregions. We refer to such a region as \emph{chart-like}. Subdivision therefore proceeds only while the embeddings exhibit meaningful internal structure beyond what can be captured by one local patch and while HDBSCAN continues to identify meaningful subclusters.

The Chart Test determines whether a node can be treated as approximately chart-like by checking whether its embeddings exhibit low intrinsic dimensionality and form a locally connected region. Low intrinsic dimensionality is assessed via the Two-NN method~\citep{facco2017estimating}, while connectivity is approximated using a $k$-nearest-neighbor graph. The test is performed on the PCA-reduced embeddings, prior to UMAP projection, so that the stopping criterion is evaluated in a stable local representation. Recursion is halted when the Chart Test is passed, when HDBSCAN identifies at most one cluster, or when resulting clusters fall below a minimum size threshold.

\paragraph{General Applicability}
Latent Task Trees require only model embeddings and do not rely on dataset-specific
tuning, external embedding models, or data annotations (e.g., domain labels). As a result,
the same construction pipeline can be applied across model sizes, domains, languages, and
modalities. The procedure is also efficient relative to RL training: after a forward pass
to extract embeddings, tree construction uses standard unsupervised operations whose cost is negligible compared to autoregressive generation. In
\hyperref[app:tree_breakdowns]{Appendix~\ref*{app:tree_breakdowns}}, we support this claim of generality with qualitative
analyses, full tree visualizations, and runtime statistics across math, coding, medical, legal, financial,
multi-domain, multilingual, and multimodal settings.

\section{Bayesian Manifold Curriculum}

We now propose \textbf{Bayesian Manifold Curriculum (BMC)}, a structured decision-making framework for problem selection over a Latent Task Tree. The tree provides a hierarchical approximation to the model's latent task manifold, allowing the scheduler to allocate training effort across task regions and share information between related prompts at multiple levels of abstraction.

BMC aims to select prompts that provide high \emph{learning signal} while maintaining coverage over the latent task manifold. Following prior work in RL for LLMs~\citep{foster2025learning, zhang2025speed, zeng2025cures}, we instantiate learning signal using reward variability across multiple rollouts. Assuming each prompt is allocated $k$ rollouts, the learning signal $y_i$ for prompt $i$ is defined as
\begin{equation}
y_i = \phi\big(\mathrm{Var}([r_{i1}, r_{i2}, \dots, r_{ik}])\big),
\end{equation}
where $\phi: \mathbb{R} \to [0,1]$ is an optional normalization function. For binary rewards, reward variance is maximized at a $50\%$ success rate, so this signal prioritizes prompts near the model's current capability frontier: prompts that are neither already solved nor completely out of reach, and are therefore expected to be most productive for group-relative learning. We provide a more detailed discussion of learning-signal choices and normalization in \hyperref[app:learning_signal]{Appendix~\ref*{app:learning_signal}}.

BMC consists of three recurring steps:
\textbf{(1) select problems top-down} by traversing the Latent Task Tree,
\textbf{(2) update prompt-level beliefs} from the observed rollout rewards, and
\textbf{(3) propagate those updates bottom-up} so the tree reflects the new information.
We describe each step below.

\begin{figure*}[t]
  \centering
  \includegraphics[width=\textwidth]{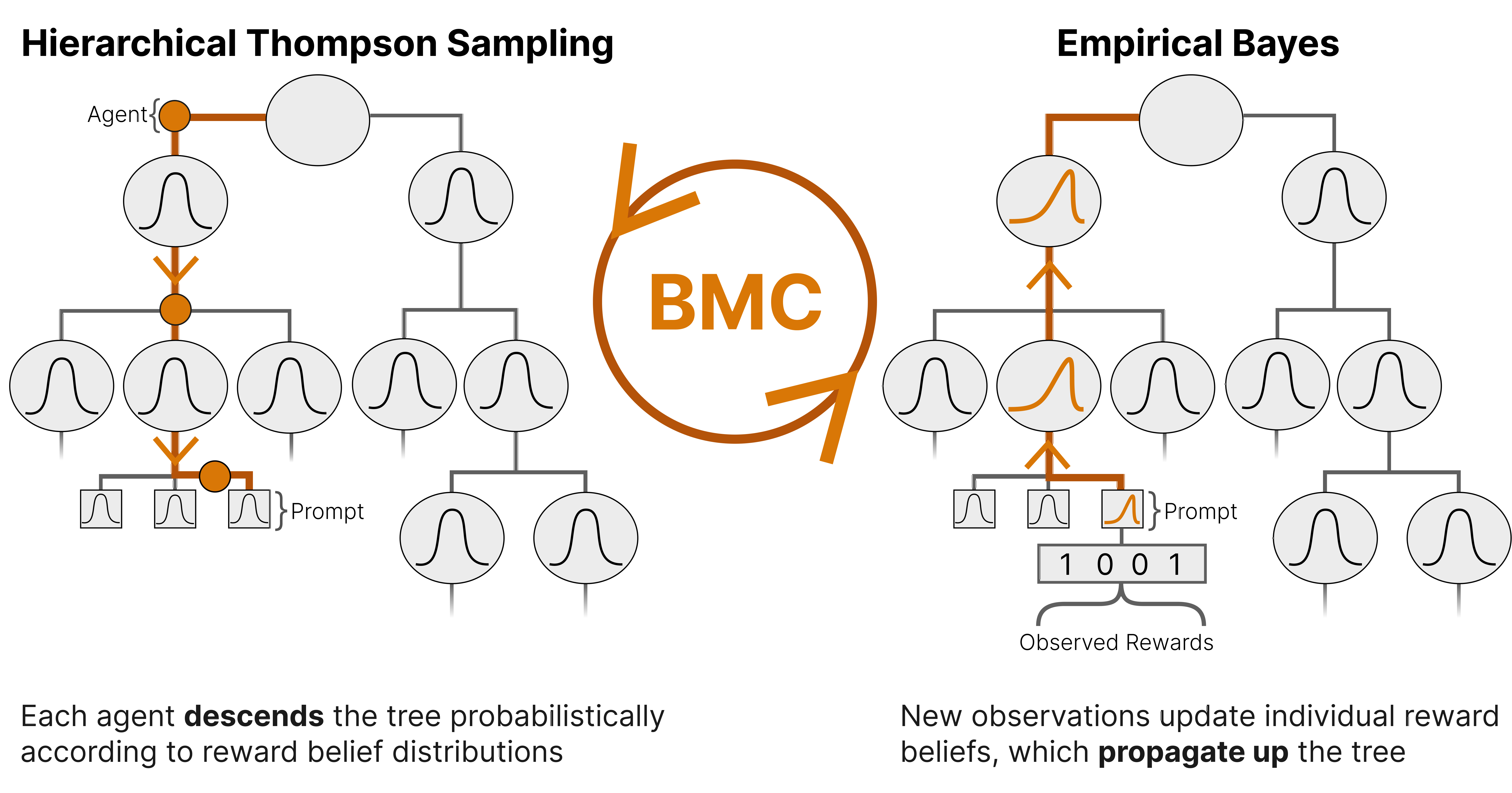}
  %\vspace{-0.1in}
  \caption{\textbf{Bayesian Manifold Curriculum (BMC).}
  \textbf{Left:} During hierarchical Thompson sampling, each batch element independently descends the Latent Task Tree according to sampled reward beliefs, recursively selecting child nodes until reaching a prompt. The figure illustrates the trajectory of a single agent for visualization purposes.
  \textbf{Right:} After rollout rewards are observed, prompt-level beliefs are updated and propagated upward through the tree using empirical Bayes updates, allowing observations from individual problems to influence higher-level regions of the task manifold.}
  \label{fig:bmc}
  %\vspace{-0.1in}
\end{figure*}

\subsection{Top-Down Problem Selection (Hierarchical Thompson Sampling)}

To construct a training batch, we instantiate one independent sampler per batch element. We refer to these samplers as \emph{agents}, since each independently traverses the Latent Task Tree from the root to a leaf. Each agent descends the tree via hierarchical Thompson sampling~\citep{hong2022hierarchicalbayesianbandits}. This creates a natural balance between exploration and exploitation, allowing different elements of the same batch to explore distinct regions of the task space.

At a node with children $c \in \mathcal{C}$, the agent draws a reward estimate for each child:
\begin{equation}
\tilde{y}_c \sim \mathcal{N}(\mu_c, \sigma_c^2), \quad \forall c \in \mathcal{C},
\end{equation}
and selects $c^* = \arg\max_{c \in \mathcal{C}} \tilde{y}_c$. This procedure is applied recursively, guiding the agent from coarse, high-level regions of the tree toward increasingly fine-grained subsets of problems. The descent terminates at a leaf node (an individual prompt), which is then selected for inclusion in the batch.

A key feature of this design is that BMC interleaves \emph{focus} and \emph{dispersion} across levels of abstraction. When one region appears substantially more productive, multiple agents can descend into that region, concentrating training effort where learning signal is high. When several regions appear similarly productive, independent descents spread agents across them, producing batches that cover distinct problem types. In this way, BMC allocates more samples to productive parts of the task manifold while preserving diversity among comparably useful alternatives.

We enforce without-replacement sampling using a batch-aware variant of hierarchical descent, since prompts are selected through the tree rather than by globally ranking prompt-level arms; see \hyperref[app:sampling_without_replacement]{Appendix~\ref*{app:sampling_without_replacement}} for details.

\subsection{Non-Stationary Belief Modeling (Bayesian Filtering)}
\label{sec:non-stationary-belief}

Each leaf of the Latent Task Tree corresponds to an individual prompt and maintains a probabilistic belief over its expected learning signal. For clarity, we present this belief
as a Gaussian model,
\[
y_i \sim \mathcal{N}(\mu_i, \sigma_i^2),
\]
where \(\mu_i\) estimates the expected signal and \(\sigma_i^2\) represents uncertainty.
In practice, we use a Logit-Normal parameterization to respect the bounded range of the
normalized learning signal; see \hyperref[app:logit-normal-sigma-rule]{Appendix~\ref*{app:logit-normal-sigma-rule}} for details.

Since the policy changes during training, a prompt's difficulty, and therefore its learning signal, is not fixed. 
A prompt that was productive earlier may become too easy, while a previously unsolvable prompt may become productive later. 
Thus, BMC needs an online estimation mechanism that updates prompt-level beliefs as new rollout evidence arrives while avoiding overcommitment to stale observations.

We implement this using a lightweight Bayesian filter inspired by the Normal--Inverse-Gamma model, the standard conjugate Bayesian model for a Gaussian with unknown mean and variance. 
Rather than maintaining the full conjugate posterior, we track only the mean \(\mu_i\) (the current belief about the prompt's learning signal), variance \(\sigma_i^2\) (uncertainty in that belief), and an effective prior weight \(\lambda_i\) (the strength of accumulated evidence, which controls how quickly the belief can adapt).

A key component of the belief update is \emph{surprise}, which we define as a standardized prediction error. Let $y_i^{(t)}$ denote the observed learning signal for prompt $i$ at step $t$. We define surprise as
\begin{equation}
s_i^{(t)} = \frac{y_i^{(t)} - \mu_i^{(t-1)}}{\sigma_i^{(t-1)}}.
\end{equation}
This quantity measures how unexpected the new observation is relative to the current belief. The standardization is important: the same absolute error is more informative when the current belief is confident than when it is already highly uncertain.

Surprise modulates the influence of prior evidence through an effective prior weight
\begin{equation}
\label{eq:lambda_update}
\lambda_i^{\mathrm{eff}} = \lambda_i^{(t-1)} \cdot \exp\!\left(- (s_i^{(t)})^2 \right),
\end{equation}
so that highly surprising observations reduce reliance on the prior belief. The posterior mean is updated as
\begin{equation}
\label{eq:belief_update}
\mu_i^{(t)} = \frac{\lambda_i^{\mathrm{eff}} \cdot \mu_i^{(t-1)} + y_i^{(t)}}{\lambda_i^{\mathrm{eff}} + 1}.
\end{equation}

The variance update balances confidence contraction with uncertainty injection:
\begin{equation}
\label{eq:uncertainty_update}
\sigma_i^{2\,(t)} =
\underbrace{
\sigma_i^{2\,(t-1)}
\frac{\lambda_i^{\mathrm{eff}}}{\lambda_i^{\mathrm{eff}} + 1}
}_{\substack{\text{confidence}\\[-1pt]\text{contraction}}}
+
\underbrace{
\frac{\log\!\left(1 + (s_i^{(t)})^2 + \mathrm{staleness}_i\right)}
{\lambda_i^{\mathrm{eff}} + 1}
}_{\substack{\text{uncertainty}\\[-1pt]\text{injection}}},
\end{equation}
where $\mathrm{staleness}_i$ is a nonnegative term that increases as prompt $i$ goes unsampled, reflecting the possibility that its belief has become stale under policy drift.

Intuitively, prompt-level beliefs only change based on direct evidence. The beliefs become more \emph{confident} when observations align with expectations and are recent, and they become more \emph{uncertain} when surprising observations 
are encountered or when a prompt goes a long time without feedback, making its evidence stale. This adaptive mechanism allows the curriculum to remain responsive to model drift, revisit previously explored problems when necessary, and avoid premature convergence to stale regions of the task manifold. Further explanation and intuition for these update rules are provided in \hyperref[app:saturate]{Appendix~\ref*{app:saturate}}.

\subsection{Bottom-Up Belief Propagation (Empirical Bayes)}

After updating individual prompt beliefs, BMC propagates information upward through the Latent Task Tree so that top-down sampling can use new evidence at multiple levels of abstraction. We perform this propagation using an \emph{empirical Bayes} approach: each internal-node belief is estimated directly from the posterior statistics of its children. These aggregated beliefs allow the scheduler to identify productive regions of the task manifold rather than relying only on independent prompt-level estimates.

For each non-leaf node, we aggregate the beliefs of its children using a precision-weighted estimator. Let $\{(\mu_i, \sigma_i^2)\}_{i=1}^K$ denote the posterior mean and variance of the $K$ child nodes. Intuitively, child nodes with more confident beliefs should contribute more to the parent's belief than highly uncertain ones. However, uncertainty alone is not sufficient: child nodes may disagree substantially in their estimated learning signal, reflecting genuine heterogeneity within a region of the task manifold. To account for this, we introduce a subtree-level heterogeneity term inspired by random-effects meta-analysis~\citep{dersimonian1986meta},
\begin{equation}
\tau_p = \max\!\left(0, \mathrm{Var}(\mu_i) - \mathbb{E}[\sigma_i^2]\right),
\end{equation}
which measures the extent to which variation between child means exceeds what would be expected from their internal uncertainty alone.

Using this correction, we define precision weights
\begin{equation}
w_i = \frac{1}{\sigma_i^2 + \tau_p},
\end{equation}
and compute the parent belief as
\begin{equation}
\mu_p = \frac{\sum_i w_i \mu_i}{\sum_i w_i}, \qquad
\sigma_p^2 = \frac{1}{\sum_i w_i}.
\end{equation}

When children agree, $\tau_p$ is near zero and aggregation behaves like ordinary precision weighting. When child beliefs disagree, $\tau_p$ inflates parent uncertainty and reduces overconfident averaging across heterogeneous children. This makes heterogeneous regions less likely to be selected overconfidently during top-down sampling.

Belief propagation proceeds recursively from the leaves to the root after each batch update. 
This creates a soft coupling between related prompts: an observation at one prompt updates 
its leaf belief, changes the beliefs of its ancestors, and thereby affects the sampling 
probabilities of other prompts in the same subtree. In this sense, BMC uses the Latent 
Task Tree to propagate direct evidence into predictions of learning signal for related 
prompts \emph{before} observing them directly. This gives the scheduler a lightweight way to 
exploit endogenous learning dynamics: sampled problems update not only their own beliefs, 
but also the scheduler's expectations over nearby regions of the latent task manifold.

\paragraph{General Applicability}

BMC is compatible with any group-relative RL algorithm that generates multiple rollouts 
per prompt (e.g., GRPO, DAPO, GSPO). Because it is built on Latent Task Trees, BMC inherits the domain, language, and modality generality of the tree construction procedure. Moreover, BMC is not limited to binary-verifier settings: by choosing an appropriate normalized signal over rollout outcomes, the same framework can be extended to continuous reward settings such as RLHF \citep{lambert2026reinforcementlearninghumanfeedback} and rubric-based RL \citep{gunjal2025rubricsrewardsreinforcementlearning}.

\section{Experiments}

We evaluate BMC by asking how different sampling strategies allocate training effort over the problem space, rather than relying only on final benchmark performance. Our analysis is organized around three axes:

\textbf{Training Efficiency (Productivity).} 
How effectively does a sampling strategy allocate computation toward informative training signals, as measured by reward variance and effective sampling ratio? How quickly do these signals translate into learning progress, as measured by pass@1 on the full training set over time?

\textbf{Coverage and Information Sharing (Diversity).} 
How does a method distribute sampling across the latent task manifold, and to what extent does it enable information sharing across related problems?

\textbf{Evaluation Performance (Utility).} 
How do differences in sampling behavior translate to performance on held-out evaluation benchmarks? Do some types of training problems contribute more to evaluation performance than others?

These axes are not necessarily aligned: strategies that improve productivity or coverage do not always yield superior evaluation performance. Our experiments therefore characterize how different samplers navigate the trade-off between learning signal, task coverage, and evaluation relevance.

\subsection{Experimental Setup}
We evaluate BMC in the setting of reinforcement learning for mathematical reasoning tasks, using the \textbf{DAPO-Math-17K} dataset~\citep{yu2025dapo} and base models \texttt{Qwen3-4B-Base} and \texttt{Qwen3-8B-Base}~\citep{qwen3}. We consider both GSPO~\citep{zheng2025group} and GRPO~\citep{shao2024deepseekmath} as backbone policy optimization algorithms. We primarily use GSPO due to its strong empirical performance among recent group-relative policy optimization methods~\citep{zheng2025group,khatri2025art}. All methods are implemented in \texttt{verl}~\citep{sheng2024hybridflow} and share identical model initialization, reward function, optimization hyperparameters, and rollout allocation per prompt. We vary only the \emph{prompt sampling strategy} to isolate its effect on learning dynamics and computational efficiency.

\textbf{Baselines and ablations.}
We compare BMC against commonly used sampling strategies, as well as ablations designed to isolate the contribution of its components:

\colorbullet{uniformgray} \textbf{\emph{Uniform Sampling}.}
Prompts are sampled uniformly at random throughout training, using a fixed number of rollouts per prompt. This serves as the default sampling strategy in many RL pipelines for LLMs.

\colorbullet{dynamicblue} \textbf{\emph{Dynamic Sampling}.}
Prompts are initially sampled uniformly, but prompts whose rollouts exhibit zero reward variance are excluded. Batches are constructed by repeatedly sampling candidate prompts until the final selected batch contains only prompts with non-zero reward variance~\citep{yu2025dapo}. This approach prioritizes informative rollouts at the cost of increased wall-clock time.

\colorbullet{thompsonpurple} \textbf{\emph{Difficulty Only (Thompson Sampling)}.}
Prompts are modeled independently with the same non-stationary beliefs used by BMC, but without the Latent Task Tree. At each step, prompts with the highest posterior samples are selected for training. This is a Thompson-style posterior sampling baseline~\citep{thompson1933likelihood, daniel2018tutorial}, where Thompson Sampling is a standard bandit algorithm for balancing exploration and exploitation through posterior sampling. In our analysis, this baseline serves both as an \emph{ablation}, isolating the effect of removing structure, and as a \emph{representative baseline} for adaptive curriculum methods that operate at the level of individual prompts and implicitly treat prompt selection as an independent-arm bandit problem. See \hyperref[app:standard-bandit-pattern]{Appendix~\ref*{app:standard-bandit-pattern}} for further discussion and comparisons to external baselines.

\colorbullet{treered} \textbf{\emph{Tree Only}.}
Prompts are organized according to the Latent Task Tree, and samples descend the tree probabilistically using fixed beliefs, but these beliefs are not updated with experience. This isolates the effect of structured sampling without adaptive belief updates.

\paragraph{Round-Robin Initialization.}
For BMC and its ablations (Difficulty Only and Tree Only), the first phase of training uses a \emph{round-robin} initialization pass, in which each training prompt is sampled once before adaptive sampling is activated. This phase ends at \textbf{step 136}. This pass is \emph{not required} by the method, but provides a stable and compute-efficient initialization for our experiments by avoiding a purely unobserved cold-start regime. Further details are provided in \hyperref[app:round_robin_initialization]{Appendix~\ref*{app:round_robin_initialization}}.

\begin{figure}[!h]
  \centering
  %\vspace{-4pt}
  \includegraphics[width=0.95\textwidth]{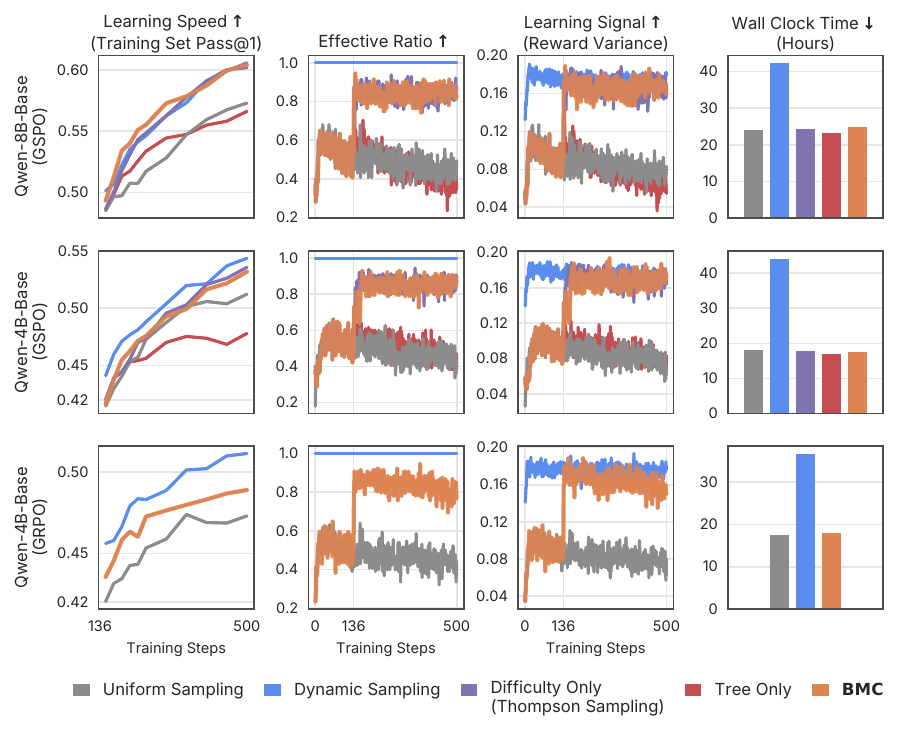}
  \caption{\textbf{Training efficiency across sampling strategies.}
  Dynamic sampling achieves the highest effective ratio and learning signal, but incurs substantially higher wall-clock time due to repeated batch regeneration.
  In contrast, Difficulty Only and BMC achieve comparable learning speed while maintaining training times close to uniform sampling.
  Uniform sampling exhibits lower learning signal and slower training progress.}
  \label{fig:metrics_main}
\end{figure}

\subsection{Training Efficiency (Productivity)}
\label{main:productivity}

Curriculum learning can be viewed as allocating training effort toward data that most effectively improves the model's performance over the training distribution. A more \emph{productive} curriculum is therefore expected to translate into faster learning progress on the training set. This reflects the fact that improvements on one subset of problems can transfer to others within the training distribution, allowing the model to solve problems it has not been explicitly trained on due to shared underlying structure. Thus, we quantify \emph{learning speed} as accuracy (e.g., pass@1) on the full training set over time, providing a direct measure of how quickly each method fits the data. To connect this to the underlying learning signal, we track both the \emph{effective ratio}, the fraction of sampled problems with non-zero reward variance that contribute to the gradient, and the average reward variance itself, which quantifies the strength of the learning signal driving these updates.

From \hyperref[fig:metrics_main]{Figure~\ref*{fig:metrics_main}}, we observe a consistent relationship between learning signal and learning speed. Across both model sizes and group-relative algorithms, higher reward variance and effective ratios correspond to faster improvement in training-set accuracy, with curriculum-based strategies achieving roughly a 40\% effective-ratio improvement over uniform sampling. Dynamic Sampling attains an effective ratio of 1.0 by construction, but does so through repeated resampling, leading to substantially higher wall-clock time. In contrast, BMC and Difficulty Only recover much of Dynamic Sampling's learning-speed benefit while remaining close to uniform sampling in training time, especially for the 8B model.

Importantly, equivalent training-set accuracy does not imply that the same problems are being solved. Different sampling strategies may achieve similar pass@1 values while covering distinct subsets of the training distribution. This distinction highlights that training accuracy alone does not capture how learning is distributed across the task space, and will be important in the following subsections.

\subsection{Coverage and Information Sharing (Diversity)}
\label{main:diversity}

\begin{figure}[!h]
  \centering
  \includegraphics[width=\textwidth]{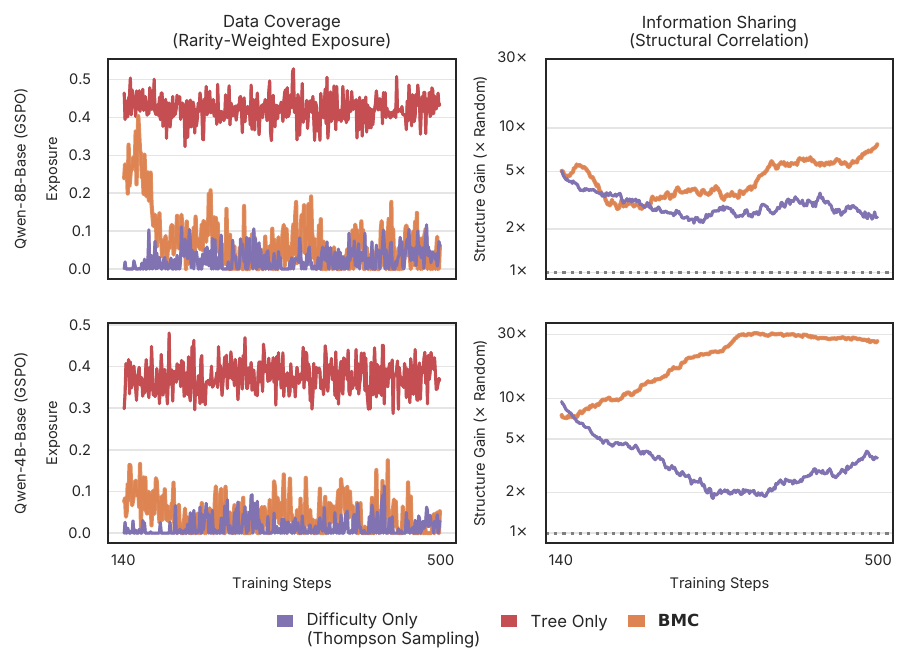}
   \caption{\textbf{Coverage and information sharing across sampling strategies.}
   \textbf{Left}: \emph{Rarity-weighted exposure}, measuring coverage of underrepresented regions of the task manifold.
    BMC balances the diversity-focused behavior of the tree-only ablation with the productivity-focused behavior of Difficulty Only.
    \textbf{Right}: \emph{Structure gain}, measuring how much variation in learning signal is explained by the Latent Task Tree relative to size-matched random partitions; the dotted gray line marks the random-partition baseline ($1\times$).
    Both BMC and Difficulty Only exhibit substantial structure gain, indicating that learning signal already possesses latent structure even when the tree is not explicitly used for sampling.}
\label{fig:diag_main}
\end{figure}

We next analyze how different sampling strategies distribute training effort across the task manifold and how this affects structure in the learned beliefs. Coverage reflects how broadly a method explores distinct regions of the task space, while information sharing is reflected in the extent to which learning signal (and therefore, difficulty) is correlated across related problems. 

To quantify coverage and imbalance correction, we use a diagnostic called \emph{rarity-weighted exposure}, which measures how much of the sampled batch lies in less-represented regions of the data. As shown in \hyperref[fig:diag_main]{Figure~\ref*{fig:diag_main}}, BMC interpolates between Tree Only (which emphasizes diversity) and Difficulty Only (which emphasizes productivity), balancing exploration of the task manifold with prioritization of informative problems.

To study information sharing, we aim to measure whether learning signal is correlated according to the Latent Task Tree. Ideally, this would be done by estimating mean@k with large $k$ over time to directly measure shared difficulty across problems, but this is computationally prohibitive. Instead, we approximate this using the learned beliefs $\mu_i$ for each prompt, which estimate expected learning signal. Using these beliefs, we define a diagnostic called \emph{structure gain}, which measures how well the tree explains variation in learning signal. Specifically, we compute the $R^2$ (variance explained) of the penultimate clusters of the tree and compare it to the average $R^2$ obtained from $n=1000$ random partitions with matching cluster sizes. We report the ratio between these quantities, which captures how much learning signal is explained by the tree relative to random chance. While absolute $R^2$ values are modest, we focus on the relative gain over random partitions.

As shown in \hyperref[fig:diag_main]{Figure~\ref*{fig:diag_main}}, the tree explains more variation in learning signal than random partitions for both BMC and Difficulty Only. Notably, these correlations are present even when the tree is not used to guide sampling, as shown by Difficulty Only's structure gain. This suggests that difficulty and learning signal are already correlated with the latent task structure. BMC then uses the Latent Task Tree as a structured prior over these correlations: observations from one region are propagated through the hierarchy to inform beliefs about related regions. When this prior aligns with real structure in the learning signal, sampling and belief propagation form a feedback loop, allowing the scheduler to build a broader picture of where learning signal lies from fewer direct samples. Intuitively, BMC does not \emph{create} structure or correlations in the learning signal; it exploits structure that \emph{already exists}.

We provide exact formulas and additional discussion of both of these diagnostics in \hyperref[app:treediag]{Appendix~\ref*{app:treediag}}.

\subsection{Evaluation Performance (Utility)}
\label{main:utility}

We now evaluate how differences in sampling behavior translate to performance on held-out benchmarks. While the previous sections focus on learning dynamics within the training distribution (productivity and diversity), evaluation performance reflects how these dynamics transfer to downstream tasks. We refer to evaluation relevance as \emph{utility}: because training problems differ in their relationship to evaluation benchmarks, methods that improve productivity or diversity need not yield superior evaluation gains.

To capture variation in distribution and task type, we evaluate on several benchmark categories. For in-distribution performance, we include standard English math benchmarks (AIME, AMC, MATH500). We also consider less-represented but related domains, such as Chinese math benchmarks (CNMO24, CCEE24). Finally, we include GPQA-Diamond, a challenging science dataset, to assess out-of-domain generalization. Additional dataset and evaluation details are provided in \hyperref[app:datasets]{Appendix~\ref*{app:datasets}}. In the main analysis below, we focus on \texttt{Qwen3-8B-Base} trained with GSPO, which provides the clearest setting for comparing evaluation profiles; additional settings are reported in \hyperref[app:evaluation_curves]{Appendix~\ref*{app:evaluation_curves}}.

\begin{figure*}[t]
  \centering
  \includegraphics[width=\textwidth]{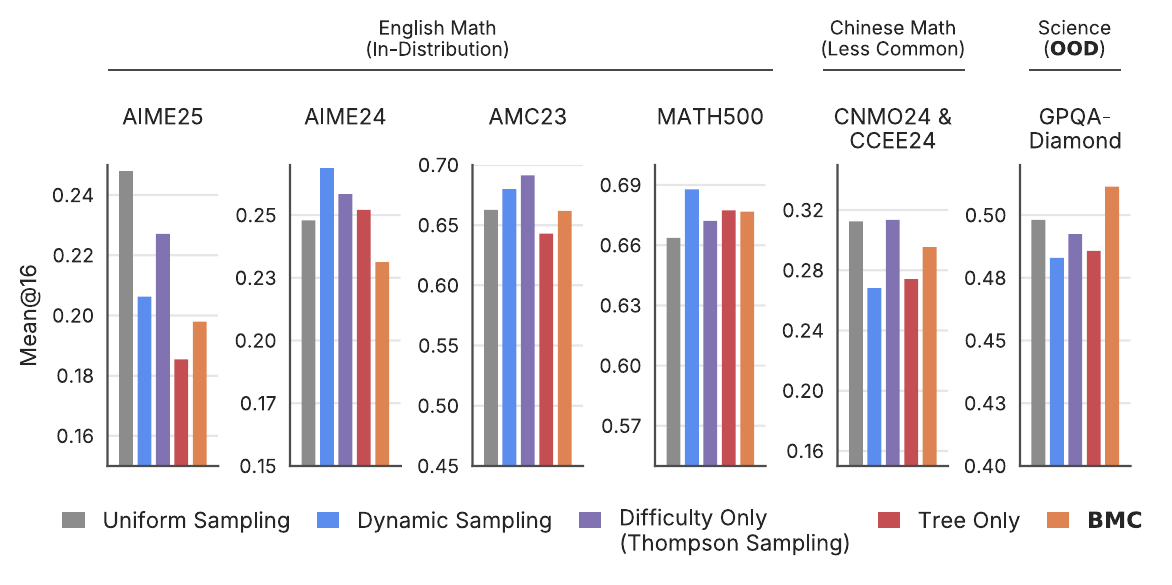}
  \caption{\textbf{Evaluation performance across benchmark categories.}
Results are shown for \texttt{Qwen3-8B-Base} trained with GSPO; additional settings are provided in \hyperref[app:evaluation_curves]{Appendix~\ref*{app:evaluation_curves}}, and per-step trajectories for this run are shown in \hyperref[fig:eval_curve_8b_gspo]{Figure~\ref*{fig:eval_curve_8b_gspo}}.
No single sampling strategy dominates across all evaluations, reflecting tradeoffs between productivity, diversity, and utility. Methods with similar learning signal or training efficiency can produce different evaluation profiles due to differences in problem-type exposure. BMC performs strongly on the OOD GPQA-Diamond benchmark, while exhibiting trade-offs on several in-distribution math evaluations.}

\label{fig:eval_main}
\end{figure*}

The evaluation profiles in \hyperref[fig:eval_main]{Figure~\ref*{fig:eval_main}} show that distinct sampling strategies induce different capability profiles, even when their productivity metrics are similar. To better understand these effects, we analyze the behavior of each method in turn.

\colorbullet{dynamicblue} \textbf{\emph{Dynamic Sampling}.} 
Despite achieving a perfect effective ratio and the highest average learning signal, dynamic sampling performs poorly on less-represented benchmarks. This is consistent with a subtle consequence of the original DAPO implementation, which stores prompts in a first-come-first-served manner and skips excess prompts when batches overflow. In imbalanced datasets, this can disproportionately filter out less-represented problem types, limiting exposure to such problems and potentially degrading performance on Chinese benchmarks and OOD tasks. See \hyperref[fig:problem_skipping]{Figure~\ref*{fig:problem_skipping}} for a visualization of this issue.

\colorbullet{thompsonpurple} \textbf{\emph{Difficulty Only (Thompson Sampling)}.}  
Thompson sampling avoids the prompt-skipping issue of dynamic sampling, leading to improved performance on less-represented benchmarks. However, because selection is performed globally over individual prompts, it still allocates disproportionate effort to more common problem types. This results in improved balance relative to dynamic sampling, but does not fully address sampling imbalance issues; see \hyperref[app:frontier_imbalance]{Appendix~\ref*{app:frontier_imbalance}} for more discussion.

\colorbullet{treered} \textbf{\emph{Tree Only}.}  
The tree-only ablation emphasizes diversity without incorporating learning signal, and does not outperform other methods on any benchmark. Its strongest performance on MATH500 is consistent with the benchmark’s larger size and heterogeneity, where broad coverage can be beneficial. However, the lack of prioritization of informative problems limits its overall effectiveness.

\colorbullet{uniformgray} \textbf{\emph{Uniform Sampling}.}  
Uniform sampling achieves similar learning signal and effective ratio to the tree-only method, but differs in how diversity is realized. While tree-based sampling promotes intra-batch diversity, uniform sampling spreads coverage across batches. This brute-force coverage can occasionally benefit specific benchmarks (e.g., AIME 2025), but lacks systematic prioritization of either informative or underrepresented problem types.

\colorbullet{bmcorange} \textbf{\emph{BMC}.}
BMC achieves learning signal comparable to Difficulty Only while improving coverage across the task manifold. More specifically, it gains increased exposure to rarer, less-represented problem types, as shown in \hyperref[main:diversity]{Section~\ref*{main:diversity}}. In particular, BMC performs strongly on the OOD GPQA-Diamond benchmark, suggesting that jointly optimizing productivity and diversity can promote capabilities not captured by standard English mathematics evaluations. Its trade-offs on several in-distribution benchmarks therefore suggest that evaluation performance depends not only on whether training is productive, but on which \emph{types} of problems receive that productive effort.

We interpret this pattern as evidence of an \textbf{evaluation deadzone}: BMC may allocate training effort to productive types of problems, but the capabilities induced by those problem types are weakly measured by the available evaluation suite. This differs from a \textbf{gradient deadzone}, where prompts are unproductive because they provide little or no policy-gradient signal. Here, the issue is not the absence of learning signal, but a mismatch between the types of problems being trained and the evaluations used to measure progress.

\subsection{Utility-Aware Sampling and Approximate Causality}
\label{sec:bmct}

The evaluation deadzone interpretation raises a natural question: can we intentionally steer BMC toward regions that are more relevant to a target evaluation distribution? The preceding analysis suggests that productivity, diversity, and evaluation utility are not automatically aligned. However, this analysis is still observational: differences in downstream performance show that problem-type exposure matters, but they do not by themselves establish which types of problems in the training distribution support which evaluations. A related limitation applies to the structure gain diagnostic in \hyperref[main:diversity]{Section~\ref*{main:diversity}}. Structure gain shows that learning signal is partially aligned with the Latent Task Tree, supporting the \emph{correlational} premise behind hierarchical selection. It does not establish the stronger \emph{causal} premise motivating bottom-up belief propagation: that learning effects propagate systematically across the latent task manifold, transferring more strongly between nearby or related problems than between distant ones.

To investigate these issues, we introduce a target-aware extension of BMC, denoted \textbf{BMC-T}. BMC-T uses latent proximity between training and target examples as a coarse proxy for evaluation relevance. The guiding assumption is not that we can identify exact causal effects of individual prompts, but that the Latent Task Tree provides an approximate intervention model over tasks\footnote{Throughout this work, we use \emph{prompt}, \emph{problem}, and \emph{task} nearly interchangeably. We use \emph{task} here in this prompt-level sense, rather than to mean an entire dataset or distribution.}: training on one task should, in expectation, have stronger effects on nearby or overlapping tasks than on distant ones. Under this assumption, changing the target distribution used to bias sampling should change which training tasks receive effort and, consequently, shift downstream performance on the corresponding evaluations.

Let $T$ denote an arbitrary target distribution, such as a held-out evaluation set or a collection of problems representing a desired capability. Target examples are \underline{\emph{not}} available for training and cannot be sampled by the bandit scheduler. BMC-T constructs a Latent Task Tree over both training and target examples, then uses the relative overlap between training and target examples within each subtree to assign a utility score to sampleable regions. During top-down sampling, this utility score biases selection toward productive regions of the training distribution that are also relevant to the target distribution. Full implementation details are provided in \hyperref[app:bmct]{Appendix~\ref*{app:bmct}}.

Our goal is not to prescribe whether evaluation-aware sampling is appropriate, but to demonstrate that utility can be treated as an \emph{independent} axis of problem selection. When the target distribution comes from an evaluation benchmark, BMC-T should not be read as a standard held-out evaluation on that same benchmark: although the target examples are not used for training, their distribution is used to steer which training problems are sampled. In strict held-out settings, the target distribution can instead be treated as a validation or development set, with final evaluations reserved separately; see \hyperref[app:target_distributions]{Appendix~\ref*{app:target_distributions}} for further discussion of utility-aware sampling and evaluation protocols.

We evaluate two BMC-T variants on \texttt{DAPO-Math-17K} using \texttt{Qwen3-8B-Base} trained with GSPO: BMC-T with the full evaluation mixture as the target distribution, denoted \textsc{BMC-T} ($T=\mathrm{All}$), and BMC-T using only AIME2024 as the target distribution, denoted \textsc{BMC-T} ($T=\mathrm{AIME2024}$). We compare both variants to standard BMC, tracking the same productivity metrics as in \hyperref[main:productivity]{Section~\ref*{main:productivity}} as well as downstream evaluation performance.

\begin{figure}[!t]
  \centering
  \includegraphics[width=0.8\textwidth]{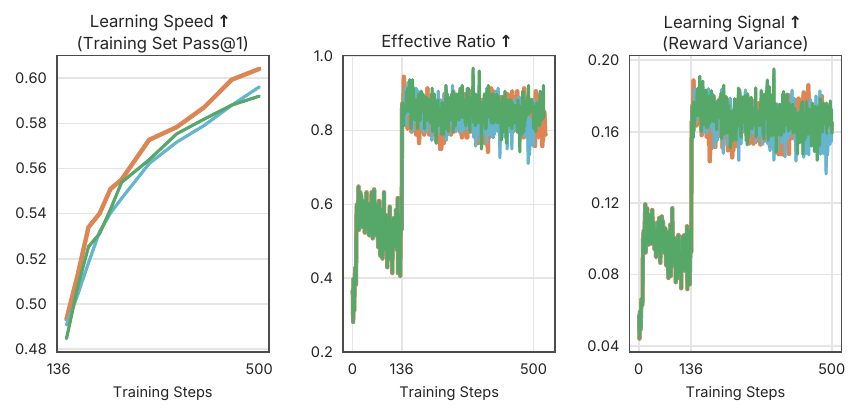}
  \vspace{0.1em}
  \includegraphics[width=0.9\textwidth]{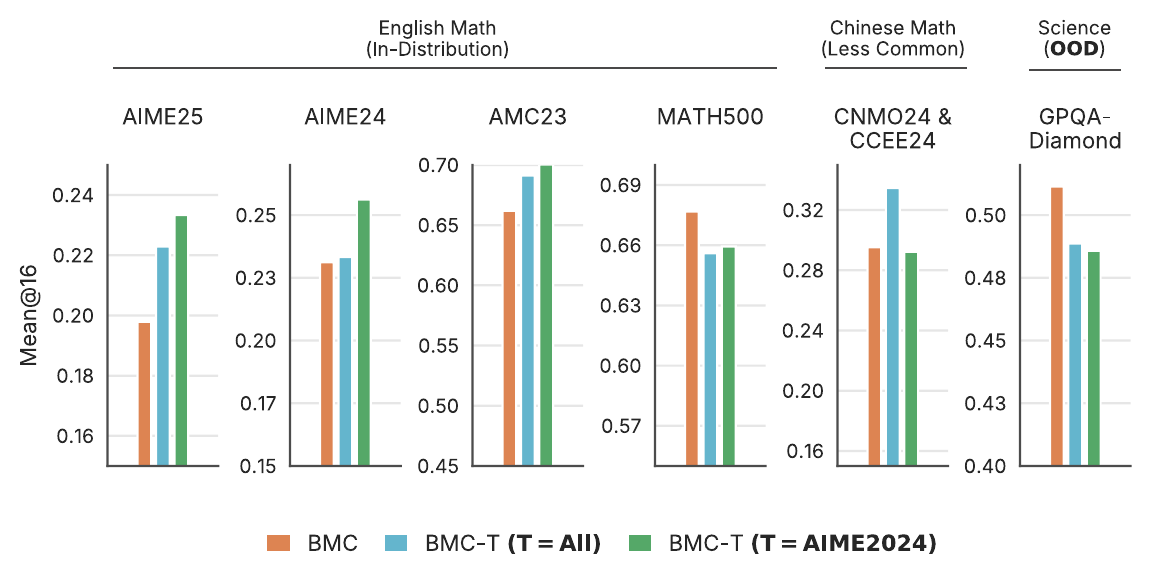}
  \caption{\textbf{Utility-aware sampling changes downstream capability profiles while preserving productivity.}
  \textbf{Top:} BMC and the two BMC-T variants exhibit similar productivity during training, with comparable learning speed, effective ratio, and learning signal.
  \textbf{Bottom:} Despite similar productivity, the variants produce different evaluation outcomes depending on the target distribution used to bias sampling. Targeting AIME2024 improves AIME-style English competition math benchmarks, while targeting the full evaluation mixture shifts performance toward a broader set of evaluations, including stronger Chinese math performance. See \hyperref[fig:eval_curve_bmct]{Figure~\ref*{fig:eval_curve_bmct}} for per-step trajectories.}
  \label{fig:bmct}
\end{figure}

As shown in \hyperref[fig:bmct]{Figure~\ref*{fig:bmct}}, BMC-T preserves BMC's productivity profile: training-set pass@1, effective ratio, and learning signal remain similar across variants. However, the evaluation profiles differ. Targeting AIME2024 improves AIME-style English competition math benchmarks, while targeting the full evaluation mixture shifts performance toward a broader set of evaluations, including stronger Chinese math performance. Thus, \textbf{utility is not reducible to productivity}. If all productive training problems contributed equally to all evaluations, changing the target distribution would have little effect once apparent learning signal remained similar. Instead, BMC-T changes downstream capabilities while preserving productivity, suggesting that different problem types carry different evaluation relevance. These results are consistent with the view that learning effects propagate non-uniformly across the latent task manifold: by changing which target-overlapping training regions receive effort, BMC-T shifts the corresponding evaluation outcomes. This is not an exact identification of prompt-level causal effects, but it suggests that latent proximity provides a useful coarse model of where training effects are likely to transfer.

Finally, BMC-T provides a useful perspective on out-of-domain generalization. In the tree constructed with the full evaluation mixture ($T=\mathrm{All}$), we observed that all GPQA-Diamond examples were isolated to a high-level node below the root, reflecting their distance from the math training distribution. Because this node contained no training examples, it could not be sampled and provided no utility bonus to any sampleable prompt. Thus, if we were to perform BMC-T with $T=\mathrm{GPQA\mbox{-}Diamond}$, the method would effectively reduce to BMC:
\[
\mathrm{BMC\mbox{-}T}\;(T=\mathrm{GPQA\mbox{-}Diamond})\approx \mathrm{BMC}.
\]
This is consistent with BMC's comparatively strong performance on GPQA-Diamond. Intuitively, when no nearby target-aligned training region exists, utility-aware sampling cannot directly steer toward the target. The best available strategy is then broad coverage of diverse, productive problems in the hope of inducing transfer to the distant target. This gives a structured-bandit perspective on LLM generalization: sampled training problems act as levers for inducing model updates whose effects may extend, or ``reach,'' beyond the sampled prompts, shifting reward distributions over nearby and distant regions of the model's latent manifold.

\begin{takeawaybox}

\textbf{Human--Policy Task Mismatch.}
Training datasets are highly heterogeneous and often imbalanced, but their task structure is often either ignored or imposed externally. Treating an entire dataset or domain as a single task assumes away internal structure, while hand-defined taxonomies may carve the problem space differently than the policy model does. In both cases, the resulting task decomposition can obscure fine-grained variation in difficulty, problem type, inter-task dynamics, and downstream relevance.

\vspace{0.45em}

\textbf{Type matters.}
Sampling strategies with similar effective ratios or learning signals can yield substantially different capability profiles and evaluation performance due to differences in problem-type exposure.

\vspace{0.45em}

\textbf{Difficulty alone is insufficient.}
Maximizing learning signal (productivity) does not guarantee better downstream outcomes; methods with lower effective ratios can outperform those with higher ones when they expose the model to more useful types of problems.

\vspace{0.45em}

\textbf{Utility is distinct from productivity and diversity.}
Not all problem types contribute equally to downstream evaluations; even strategies that improve both productivity and diversity can underperform when they allocate training effort to problem types with lower evaluation relevance.
\end{takeawaybox}

\paragraph{Structure Drift and Other Considerations.}
The latent structure underlying the task manifold may evolve over the course of training, meaning that a static tree may not perfectly reflect the model's current representation space. In practice, this introduces a form of model--structure mismatch, which we study in \hyperref[app:structure_drift]{Appendix~\ref*{app:structure_drift}}. Empirically, periodic tree reconstruction produces only modest changes in downstream performance, suggesting that the fixed tree remains a sufficiently useful approximation over the training horizons considered in this work. Additional experiments, discussions, limitations, directions for future work, and implementation details are provided in the \hyperref[app:appendix_overview]{\textbf{Appendix Overview}}.

\section{Related Work}

Following DeepSeek-R1 \citep{deepseek_r1_2025_pdf}, one of the earliest works to demonstrate both the empirical and theoretical importance of reward variance in reinforcement learning was LILO \citep{foster2025learning}, which quantified problem learnability as the variance in model success across rollouts. DAPO \citep{yu2025dapo} introduced dynamic sampling to increase learning signal at the cost of additional wall-clock time, while SRPO \citep{zhang2025srpocrossdomainimplementationlargescale} and POLARIS \citep{Polaris2025} filter prompts whose rollouts become uniformly successful to avoid oversampling saturated problems. SPEED-RL \citep{zhang2025speed} further reduces wasted compute by using partial rollouts to screen prompts for non-trivial reward variance before allocating the full rollout budget. Difficulty-aware approaches such as AdaRFT \citep{shi2026efficientreinforcementfinetuningadaptive} and VCRL \citep{jiang2025vcrl} prioritize problems using difficulty or reward variance signals aligned with the model's current capability.

Additional works attempt to anticipate prompt informativeness rather than relying solely on direct rollout evidence. GRESO \citep{zheng2025actpaysefficientreinforcement} predicts prompt utility using temporal reward dynamics, probabilistically skipping prompts likely to yield negligible reward variance before rollout. CurES \citep{zeng2025cures} instead estimates prompt difficulty via Bayesian modeling of success probability, enabling uncertainty-aware curricula. PCL \citep{gao2025promptcurriculumlearningefficient} learns a value model to predict prompt difficulty (or productivity) online, selecting prompts of intermediate difficulty without requiring explicit rollout-based estimation. DOTS \citep{sun2026improvingdataefficiencyllm} similarly estimates adaptive prompt difficulty with a separate embedding-based predictor that interpolates from a small rollout-labeled reference set, and combines this with rollout replay to reduce per-step generation cost.

Other works take a more principled approach by formulating problem selection as a non-stationary bandit problem, explicitly modeling the exploration--exploitation trade-off. Methods such as DUMP \citep{wang2025dump} and SEC \citep{chen2025self} serve as conceptual inspirations for our work, using bandit schedulers to adaptively prioritize prompt distributions based on estimated learning productivity. Subsequent work has explored combining bandit schedulers with clustering-based structure, where prompts are grouped using flat clustering methods (e.g., k-means) based on precomputed difficulty or semantic embeddings, and bandit algorithms operate over the resulting clusters \citep{do2025sparft, adaptive_curriculum_bandits_2026}. Complementary approaches such as MoPPS \citep{qu2025can}, BOTS \citep{shen2026botsunifiedframeworkbayesian}, and VADE \citep{hu2025vadevarianceawaredynamicsampling} instead treat individual prompts as bandit arms and employ variants of Thompson sampling to select prompts without explicitly modeling distribution-level structure.

In contrast to these works, our method learns structural relationships between prompts directly from the policy's latent representation space and leverages this structure through hierarchical decision-making, performing selection and belief updates over both distributions and individual problems within a unified Bayesian framework. Most notably, we distinguish between problem \emph{difficulty} and problem \emph{type} (derived from the policy's perception), treating them as complementary but distinct factors in forming an effective automated curriculum. See \hyperref[app:extended_work]{Appendix~\ref*{app:extended_work}} for an extended discussion of contemporary work and related work beyond adaptive problem sampling for LLMs.

\section{Conclusion}

This work argues that adaptive problem sampling for RL-based LLM training is governed by three distinct axes: \emph{productivity}, \emph{diversity}, and \emph{utility}. We introduced \emph{Bayesian Manifold Curriculum (BMC)}, a structure-aware framework for navigating these axes through Bayesian belief updates over a Latent Task Tree derived from policy representations. BMC and its utility-aware extension, BMC-T, provide a way to balance productive learning signal, coverage over problem types, and downstream relevance without requiring hand-defined task taxonomies, difficulty labels, external embedding models, or separate critic-style prompt predictors. Because the framework requires only hidden representations from the policy model and rollout rewards, it can be applied across domains, languages, modalities, and reward types.

Conceptually, BMC connects the \emph{manifold hypothesis} with the spirit of the \emph{Bitter Lesson}~\citep{sutton2019bitter}: rather than imposing human-defined notions of difficulty or task structure, it derives curriculum structure from the policy's learned representation space and uses that structure to guide search and learning. As RL is applied to increasingly heterogeneous datasets and reward signals, we believe that problem selection should be framed as a manifold-structured bandit problem over the policy's latent geometry, with endogenous non-stationarity governed by learning dynamics. This view shifts curriculum learning from simply finding problems of the right difficulty toward orchestrating training effort across diverse and interacting problem types, where sampling decisions shape how learning progresses through the policy's task space.

\begin{ack}
We thank Yutao Xie for helpful discussions. We also thank Cale Crowder, whose work in reinforcement learning and curriculum learning greatly inspired the development of this work.

This work was supported by a National Defense Science and Engineering Graduate (NDSEG) Fellowship awarded by the Department of Defense, Office of Naval Research.
%Use unnumbered first level headings for the acknowledgments. All acknowledgments
%go at the end of the paper before the list of references. Moreover, you are required to %declare
%funding (financial activities supporting the submitted work) and competing interests %(related financial activities outside the submitted work).
%More information about this disclosure can be found at: %\url{https://neurips.cc/Conferences/2026/PaperInformation/FundingDisclosure}.

%Do {\bf not} include this section in the anonymized submission, only in the final paper. You %can use the \texttt{ack} environment provided in the style file to automatically hide this %section in the anonymized submission.
\end{ack}

%\section*{References / Acknowledgments}

%References follow the acknowledgments in the camera-ready paper. Use unnumbered first-level %heading for
%the references. Any choice of citation style is acceptable as long as you are
%consistent. It is permissible to reduce the font size to \verb+small+ (9 point)
%when listing the references.
%Note that the Reference section does not count towards the page limit.
%\medskip

{
\small
\bibliographystyle{plainnat}
\bibliography{main}
}

%%%%%%%%%%%%%%%%%%%%%%%%%%%%%%%%%%%%%%%%%%%%%%%%%%%%%%%%%%%%
\clearpage
\appendix

\phantomsection
\section*{Appendix Overview}
\addcontentsline{toc}{section}{Appendix Overview}
\label{app:appendix_overview}

\vspace{0.5em}

\noindent
%{\large\textbf{Webpage:} 
\textbf{Webpage:} \href{https://darrienmckenzie.com/manifold-bandits}{\texttt{darrienmckenzie.com/manifold-bandits}}
%}

\vspace{0.5em}

\noindent
The webpage contains links to code, tree breakdowns (word clouds and prompts-per-cluster), and training logs.

\vspace{0.5em}

\noindent
The appendix is organized into four groups.

\vspace{0.75em}
\noindent\rule{\textwidth}{0.4pt}

\vspace{0.5em}

\noindent
\textbf{Extended Related Work:} Appendix~\ref{app:extended_work}. \\
Additional discussion of related and contemporary work.

\vspace{0.35em}

\noindent
\textbf{Additional Experiments:} Appendices~\ref{app:structure_drift}--\ref{app:tree_breakdowns}. \\
Additional empirical studies, including structure drift investigations, external baseline comparisons and discussions, RL experiments on medical Q\&A data, and tree visualizations across different settings.

\vspace{0.35em}

\noindent
\textbf{Discussions:} Appendices~\ref{app:frontier_imbalance}--\ref{app:future_work}. \\
Broader discussion of frontier imbalance, relationships between utility-awareness and evaluation protocols, inter-batch vs. intra-batch diversity, usage of external models, limitations, and future directions.

\vspace{0.35em}

\noindent
\textbf{Additional Experimental and Algorithmic Details:} Appendices~\ref{app:experimental_details}--\ref{app:algorithmic_details}. \\
Hyperparameters, implementation details, evaluation curves, and algorithmic details.

\vspace{0.5em}
\noindent\rule{\textwidth}{0.4pt}

\vspace{1.0em}

\noindent
\hyperref[app:extended_work]{\textbf{A \quad Extended Related Work}}
\dotfill
\pageref{app:extended_work}

\vspace{0.45em}

\noindent
\hyperref[app:structure_drift]{\textbf{B \quad Structure Drift}}
\dotfill
\pageref{app:structure_drift}

\vspace{0.45em}

\noindent
\hyperref[app:standard-bandit-pattern]{\textbf{C \quad The Standard Bandit Pattern}}
\dotfill
\pageref{app:standard-bandit-pattern}

\vspace{0.45em}

\noindent
\hyperref[app:medical]{\textbf{D \quad Medical Data Experiments}}
\dotfill
\pageref{app:medical}

\vspace{0.45em}

\noindent
\hyperref[app:tree_breakdowns]{\textbf{E \quad Tree Visualizations \& Statistics}}
\dotfill
\pageref{app:tree_breakdowns}

\vspace{0.45em}

\noindent
\hyperref[app:frontier_imbalance]{\textbf{F \quad Frontier Imbalance}}
\dotfill
\pageref{app:frontier_imbalance}

\vspace{0.45em}

\noindent
\hyperref[app:target_distributions]{\textbf{G \quad Utility-Aware Sampling and the Evaluation-Protocol Tension}}
\dotfill
\pageref{app:target_distributions}

\vspace{0.45em}

\noindent
\hyperref[app:external-models]{\textbf{H \quad External Models}}
\dotfill
\pageref{app:external-models}

\vspace{0.45em}

\noindent
\hyperref[app:inter-batch-diversity]{\textbf{I \quad Inter-Batch Diversity (Temporal Awareness)}}
\dotfill
\pageref{app:inter-batch-diversity}

\vspace{0.45em}

\noindent
\hyperref[app:limitations]{\textbf{J \quad Limitations}}
\dotfill
\pageref{app:limitations}

\vspace{0.45em}

\noindent
\hyperref[app:future_work]{\textbf{K \quad Directions for Future Work}}
\dotfill
\pageref{app:future_work}

\vspace{0.45em}

\noindent
\hyperref[app:experimental_details]{\textbf{L \quad Additional Experimental Details}}
\dotfill
\pageref{app:experimental_details}

\vspace{0.45em}

\noindent

\hyperref[app:algorithmic_details]{\textbf{M \quad Additional Algorithmic Details}}
\dotfill
\pageref{app:algorithmic_details}

\vspace{0.25em}

\noindent\hspace*{1.5em}%
\hyperref[app:bmct]{M.7 \quad BMC-T}
\dotfill
\pageref{app:bmct}

\clearpage % end section

\section{Extended Related Work}
\label{app:extended_work}
\subsection{Contemporary Work}

A number of closely related papers appeared during the development and revision of this manuscript. We discuss them here as \emph{contemporary work} because they address overlapping questions in adaptive sampling, curriculum learning, empirical Bayes estimation, task discovery, and structured exploration for LLM training.
Taken together, these works reflect a broader shift from static data curation toward representation-, optimization-, and feedback-aware selection mechanisms that account for the model's evolving state or internal structure. They also highlight several recurring axes of curriculum design: \emph{productivity}, or whether a sample provides useful learning signal; \emph{diversity}, or whether selected batches cover non-redundant regions of the data; and \emph{utility}, or whether the selected data supports a target evaluation distribution. BMC is complementary to these efforts: it treats curriculum selection as a structured bandit problem over a latent task manifold derived from the policy model's own representations.

\textbf{Optimizer-induced Projected Utility Selection (OPUS)} \citep{wang2026opusefficientprincipleddata} is a dynamic data-selection method for LLM pre-training. It defines utility as expected one-step improvement on a proxy distribution, measured in the geometry induced by the optimizer, and scores candidates by the alignment between their optimizer-induced updates and a benchmark-relevant proxy direction. OPUS also includes a redundancy penalty and Boltzmann sampling to preserve diversity during batch construction. This makes OPUS closely related at the level of motivation: both works argue that effective data selection requires balancing informativeness, diversity, and target utility. However, the source of geometry differs. OPUS operates in \emph{optimizer geometry}: it asks which pre-training sequences induce parameter updates aligned with a target gradient direction. BMC operates in \emph{latent geometry}: it asks which RL prompts provide useful learning signal while preserving coverage over the model's representation-induced task manifold. Thus, OPUS and BMC study analogous selection problems in different training regimes (pre-training versus RL post-training), and instantiate geometry, utility, and diversity through different mechanisms.

\textbf{Generalizable Predictive Prompt Selection (GPS)} \citep{qu2026smallgeneralizablepromptpredictive} is an RLVR prompt-selection method that combines predicted prompt difficulty with batch-level diversity. GPS trains a lightweight generalizable prompt predictive model (PPM) over the shared optimization history, allowing difficulty predictions to transfer across prompts. Its acquisition objective combines intermediate-difficulty prioritization with history-anchored diversity, using intra-batch dispersion and inter-step exploration to reduce redundancy and improve coverage. GPS is therefore closely related to BMC in emphasizing that adaptive curricula require more than prioritizing difficulty alone. The main difference is the source and role of structure: GPS uses an external embedding representation and a learned PPM/critic-style predictor to guide max-sum diversity over prompt distances, whereas BMC induces a Latent Task Tree from the policy model's own hidden representations and performs structured Thompson Sampling over this representation-derived task geometry. BMC-T further separates \emph{utility} from productivity and diversity, allowing the curriculum to bias sampling toward a chosen target distribution.

\textbf{GradAlign} \citep{yang2026gradaligngradientaligneddataselection} is a contemporary RL data-selection method that uses gradient alignment to select training problems likely to improve downstream performance. It computes a target direction from the average policy gradient on a small trusted validation set and ranks candidate problems by cosine similarity between their policy gradients and this validation gradient. In this sense, GradAlign addresses the \emph{utility} dimension of curriculum design: it distinguishes prompts that merely provide training signal from prompts whose updates are aligned with a target task. The main difference is the geometry used for selection. GradAlign operates in policy-gradient geometry, while BMC operates in latent geometry. BMC uses the policy model's own hidden representations to induce a task manifold and performs structured sampling over that space, allowing productivity, diversity, and utility to be treated as separate axes. Thus, GradAlign provides a complementary validation-alignment mechanism, but does not explicitly control \emph{diversity} over high-utility problem types.

\textbf{Empirical Bayes Policy Optimization (EBPO)} \citep{han2026ebpoempiricalbayesshrinkage} is a contemporary method that uses empirical Bayes shrinkage to stabilize GRPO-style advantage estimation. It replaces the local group mean baseline with a shrinkage estimator that blends local reward statistics with global policy statistics estimated online, reducing variance and providing non-vanishing signals in saturated failure regimes. This is complementary to BMC: EBPO applies empirical Bayes within the RL objective to regularize advantage baselines, whereas BMC uses Bayesian belief updates and empirical Bayes aggregation over a latent task hierarchy for curriculum scheduling. EBPO also studies clustered sampling, using externally annotated topic clusters or difficulty-based ordering to improve prior estimation. In contrast, BMC induces a Latent Task Tree from the policy model's own prompt representations and uses this hierarchy as the scheduler's action structure, separating problem \emph{type} from problem \emph{difficulty}. This separation is important because prompts with similar difficulty need not represent the same problem type, and prompts of the same type may span a wide range of difficulty.

\textbf{DEPO}~\citep{tang2025highdataefficiencyreinforcement} is a data-efficient RLVR pipeline that combines offline subset construction with online rollout pruning. Its offline stage constructs a representation-based sample graph to select diverse and influential data, then applies difficulty-aware sampling to retain appropriately challenging examples. During training, DEPO filters samples using a sample-level explorability metric and replays under-explored samples to reduce rollout cost while maintaining performance. DEPO is related to BMC in recognizing that sample relationships, diversity, and difficulty matter for efficient RLVR training. However, DEPO uses graph structure primarily for offline data curation, whereas BMC uses a policy-induced Latent Task Tree as the action structure for online curriculum scheduling. These perspectives are complementary: Latent Task Trees could also be used for dataset minimization, while BMC focuses on how latent structure can guide adaptive sampling decisions during RL training.

\textbf{SAERL}~\citep{jing2026guidingllmposttrainingdata} is a recent model-internal data engineering method for RL post-training. It uses sparse autoencoder activations to estimate data diversity, difficulty, and quality, and applies these signals to offline data filtering, batch construction, and curriculum ordering. SAERL is related to BMC in emphasizing that model-internal structure can guide post-training data selection. However, the role of this structure differs: SAERL uses SAE features as an additional representation layer for preprocessing and data engineering, whereas BMC induces a Latent Task Tree directly from the policy model's hidden representations and uses it as the action structure for online curriculum scheduling. BMC updates tree-structured beliefs from observed learning signal during training, while also separating productivity, diversity, and target utility as distinct axes of curriculum design. Thus, SAERL and BMC explore complementary uses of model-internal structure: SAE-based feature analysis for data engineering, and lightweight policy-induced latent geometry for adaptive RL sampling.

\textbf{FineRouter} \citep{zhang2026scalablepromptroutingfinegrained} is a prompt routing method that automatically discovers fine-grained latent task types. It builds a prompt graph from LLM-generated task descriptions, external text embeddings, and model-preference rankings, then clusters this graph to train a task-aware router with specialized prediction heads. FineRouter is related to BMC in arguing that manually defined task taxonomies are too coarse to capture fine-grained model behavior. However, it addresses inference-time model routing rather than RL curriculum scheduling, and its task discovery relies on external task descriptions, embedding models, reward-model scores, and model-preference patterns. BMC instead induces a Latent Task Tree from the policy model's own hidden representations and uses the resulting hierarchy as the scheduler's action structure, enabling belief sharing, diversity-aware sampling, and multiple levels of task abstraction.

\paragraph{Geometries of adaptive selection.}
More broadly, OPUS, GradAlign, and BMC highlight that adaptive selection can be defined over different geometries. OPUS uses optimizer-induced update geometry, GradAlign uses policy-gradient geometry, and BMC uses latent geometry induced from the model's own representations. Each perspective captures a different signal about the training process, and understanding their tradeoffs or combinations remains an interesting open direction.

\subsection{Latent Geometry of LLMs}
\subsubsection{Layer Heterogeneity}
\label{app:layer_heterogeneity}

A substantial body of work demonstrates that transformer layers are functionally and geometrically heterogeneous. Analyses of hidden-state geometry show that representations evolve non-uniformly across depth, often exhibiting changes in intrinsic dimensionality and neighborhood structure that peak in intermediate layers \citep{valeriani2023geometry}. Layer-specific probing studies indicate that lexical semantics and structured knowledge are often most accessible in lower-to-intermediate layers, whereas later layers increasingly specialize toward next-token prediction and output reconstruction \citep{liu2024fantasticsemanticstheminvestigating, zhang2024investigatinglayerimportancelarge, lei2025layerwiserecallgeometryinterwoven}. As a result, further studies report that intermediate layer representations frequently outperform final-layer embeddings on downstream tasks, suggesting that semantically rich features are concentrated prior to the final, generation-focused layers \citep{skean2024doesrepresentationmatterexploring, skean2025layerlayeruncoveringhidden}. Complementary perspectives connect these phenomena to compression dynamics, arguing that intermediate layers behave like information bottlenecks where salient structure is compactly organized \citep{yin2024entropylawstorydata}. 

Motivated by this literature, we construct Latent Task Trees using intermediate-layer embeddings rather than final-layer representations. Our choice reflects prior evidence that these layers capture semantically meaningful structure while remaining less entangled with token-generation objectives.

\subsubsection{Manifold Representations \& Regularization}
\label{app:manifold_representations}
Beyond layer heterogeneity, another ever-growing body of literature studies the geometric structure of representations in large language models. Several works provide empirical evidence that LLM embeddings exhibit manifold-like organization, including low intrinsic dimensionality, locally linear structure, and coherent semantic clustering \citep{modell2025originsrepresentationmanifoldslarge, doimo2024representationlandscapefewshotlearning, saglam2026largelanguagemodelsencode, lee2025geometricsignaturescompositionalitylanguage}. These studies suggest that semantic relationships are not uniformly distributed in representation space but instead occupy structured, lower-dimensional regions. Most directly related to our hierarchical construction, \citet{park2025geometrycategoricalhierarchicalconcepts} show that categorical concepts can be represented as polytopes and that semantic hierarchies are reflected geometrically through orthogonality between representation subspaces. Recent work further characterizes activation geometry using local decomposition techniques, arguing that model representations are better understood as structured regions rather than isolated directions in feature space \citep{shafran2026directionsregionsdecomposingactivations}.

At the same time, the manifold hypothesis has been critically examined. One concern is methodological: UMAP can be over-interpreted, since their low-dimensional projections may preserve local neighborhoods while distorting global distances, densities, and inter-cluster relationships~\citep{jeon2025stopmisusingtsneumap}. We therefore use UMAP only as an intermediate neighborhood-preserving transformation for clustering, not as visual evidence of global geometry. A second concern is structural: \citet{robinson2025tokenembeddingsviolatemanifold} show that token input embeddings can contain singularities and non-constant local dimension, violating the requirements of a smooth global manifold. We view this as a critique of a strong \emph{global} manifold assumption, whereas BMC adopts a weaker operational view: prompt representations need only exhibit enough \emph{local} geometric regularity to support neighborhood-based partitioning. This perspective is consistent with subsequent work suggesting that LLM representations may be better described as stratified or piecewise manifolds rather than globally smooth ones~\citep{li2025unravelinglocalizedlatentslearning}.

Aside from representation analysis, several methods attempt to explicitly regularize or align manifold structure during training. These include mixture-of-experts routing based on localized manifold structure \citep{li2025routing}, and architectural constraints designed to preserve geometric coherence across layers \citep{xie2025mhc}. 

Taken together, these works suggest that while strict manifold assumptions may not hold globally, latent representations in large language models exhibit structured geometric organization that can be approximated and exploited. Our Latent Task Tree adopts this empirical viewpoint: rather than assuming a smooth manifold, we recursively approximate locally coherent regions in intermediate-layer latent space and leverage this structure for curriculum learning.

\subsubsection{Mechanistic Interpretability \& Sparse Autoencoders}

In parallel to geometric analyses of representation space, mechanistic interpretability studies often use sparse representation learning to expose latent features in neural networks. This perspective is rooted in classical sparse coding, where sparsity constraints were shown to produce localized and interpretable visual features \citep{olshausen1996emergence, olshausen1997sparse}. Recent work applies related ideas to language models through sparse autoencoders (SAEs), which learn sparse decompositions of LLM activations and can reveal interpretable latent factors \citep{cunningham2023sparseautoencodershighlyinterpretable}. Related approaches use sparsity-inducing probes, dictionary learning, and factor-analysis-style methods to identify disentangled subspaces associated with lexical, syntactic, semantic, or world-knowledge features \citep{kantamneni2025sparseautoencodersusefulcase, kim2026disentangledsparserepresentationsconceptseparated}. These methods support human interpretation and modular intervention, including causal steering \citep{Arad_2025, he2025saifsparseautoencoderframework}.

Our objective is different. We do not aim to recover human-aligned latent factors or decompose activations into individually interpretable features. Instead, we use raw intermediate representations to induce a task-level geometry over prompts, which then serves as a structure for adaptive curriculum learning. Although the resulting Latent Task Trees are often human-interpretable, this interpretability is incidental rather than required: the central goal is to allocate training according to the model's implicit organization of the problem space. Further discussion of SAEs in relation to our approach is provided in \hyperref[app:em_saes]{Section~\ref*{app:em_saes}}.

\subsection{Hierarchical Retrieval and Structured Aggregation}

A parallel line of work explores hierarchical organization for long-document summarization. Methods such as HIBRIDS \citep{cao2022hibridsattentionhierarchicalbiases} inject hierarchical document-structure biases into attention score calculation, improving the modeling of long-range document structure. RAPTOR \citep{sarthi2024raptorrecursiveabstractiveprocessing} proposes recursive abstractive summarization to build tree-structured retrieval indices, enabling coarse-to-fine information access. LeanRAG \citep{zhang2025leanragknowledgegraphbasedgenerationsemantic} combines hierarchical retrieval with knowledge graph aggregation to improve structured reasoning over long contexts. Automatic Data Curation for Self-Supervised Learning \citep{vo2024automaticdatacurationselfsupervised} applies recursive clustering to organize data prior to generation, which later inspired HERCULES \citep{petnehazi2025herculeshierarchicalembeddingbasedrecursive}, a hierarchical embedding-based recursive clustering framework for efficient summarization and retrieval.

These works leverage hierarchical structure over embedding spaces to improve inference-time processing. In contrast, our method constructs hierarchical structure over prompt representations to guide adaptive sampling during reinforcement learning. Rather than organizing knowledge for retrieval over static corpora, we use latent geometry to allocate training effort during policy optimization.

\subsection{Curriculum Learning \& Active Learning}

Outside of using RL to train LLMs, curriculum learning has long been studied as a strategy for improving optimization and generalization in neural networks. The foundational work of \citet{bengio2009curriculum} proposed presenting training examples in a meaningful order, typically progressing from easy to difficult, to stabilize and accelerate learning. Subsequent work on self-paced learning \citep{kumar2010self} formalized this intuition by introducing adaptive sample weighting based on model-dependent difficulty, allowing the curriculum to evolve as training progresses. Extensions such as teacher-student curriculum learning and automated curriculum selection \citep{matiisen2017teacherstudentcurriculumlearning, graves2017automated} further introduced adaptive mechanisms in which a policy selects training examples based on estimated learning progress.

In reinforcement learning, related ideas appear in unsupervised environment design (UED), where tasks or environments are adaptively generated to maximize learning signal \citep{dennis2021emergentcomplexityzeroshottransfer}. Approaches such as CLUTR \citep{azad2023clutrcurriculumlearningunsupervised} further explore organizing tasks within a learned latent representation space, treating tasks as residing on a structured manifold. More generally, Prioritized Experience Replay \citep{schaul2016prioritizedexperiencereplay} allocates updates based on estimated informativeness, sampling transitions with high learning signal more frequently.

Related notions of informative batch construction also appear in active learning. Bayesian Active Learning by Disagreement (BALD) selects examples that maximize the mutual information between predictions and model parameters, prioritizing points expected to provide high epistemic information~\citep{houlsby2011bayesianactivelearningclassification}. However, greedily selecting the highest-scoring examples can produce redundant batches, since each point is evaluated independently. BatchBALD addresses this by maximizing joint mutual information over a batch, penalizing redundant selections and encouraging complementary examples~\citep{kirsch2019batchbaldefficientdiversebatch}. This is closely aligned with our motivation: curriculum design should not only identify individually productive prompts, but should also account for diversity and structure when forming training batches.

\subsection{Structured \& Causal Multi-Arm Bandits}
The general notion that bandit arms can share correlations~\citep{boda2019correlatedbanditsorminimize,Gupta_2021}, reside on a shared structure~\citep{ginebra1995response,pandey2007bandits,mersereau2009structured}, or support learnable causal interventions~\citep{lattimore2016causalbanditslearninggood} served as a main catalyst for this work.

Additionally, several variants of Thompson Sampling are especially relevant to our setting. Hierarchical Thompson Sampling (HierTS) was introduced for hierarchical Bayesian bandits, where related bandit tasks share information through a common hyper-posterior~\citep{hong2022hierarchicalbayesianbandits}. This is conceptually aligned with BMC, where prompts can be viewed as related tasks whose learning-signal estimates should not be treated as independent. Subsequent work extends this idea to tree-structured action spaces, where actions correspond to leaves and internal latent variables induce correlations among action rewards~\citep{hong2022deephierarchybandits}. More recently, empirical Bayesian multi-bandit Thompson Sampling (ebmTS) estimates shared prior covariance across bandit instances from data and uses the resulting empirical posterior for exploration~\citep{jiang2025empiricalbayesianmultibanditlearning}. BMC differs from these works by inducing a latent task tree from the model's own prompt representations, rather than relying on a given task relation, action taxonomy, or covariance structure. The resulting hierarchy supports curriculum scheduling in RL by combining structured exploration, diversity-aware sampling, and belief sharing over prompts.

\clearpage % end section
%%%%%%%%%%%%%%%%%%%%%%%%%%%%%%%%%%%%%%%%%%%%%%%%%%%%%%%%%%%%%%%%%%%%%%%%%%%%%%%%%%%%%%%%%%%%%%%%%%%%%%%%%%%%%%%%%%%%%

\section{Structure Drift}
\label{app:structure_drift}

BMC constructs the Latent Task Tree at the start of training and keeps its topology fixed throughout RL. The beliefs associated with each node are updated over time, but the tree structure itself remains static. This accounts for changes in the model's observed reward landscape, but not for changes in the model's latent representation of prompts during training.

We keep the tree fixed in the main experiments for two reasons. First, RLVR typically induces relatively small parameter updates compared with large-scale supervised fine-tuning \citep{mukherjee2025reinforcementlearningfinetunessmall, zhu2025pathtakenrlvrprovably}. Second, the tree is constructed from prompt representations, and prompt tokens are masked from the policy-gradient objective. As a result, the representations used to build the tree may be less directly perturbed by RL updates than the response tokens being optimized. Nonetheless, it is unreasonable to assume that the model's perception of prompts remains completely fixed. We therefore investigate structural drift in two ways: (1) by comparing Latent Task Trees before and after RL, and (2) by running BMC with periodic tree reconstruction.

\subsection{Tree Structure Before and After RL}

We first compare Latent Task Trees constructed from pre-RL and post-RL model checkpoints. This analysis is not intended to isolate the causal effect of a particular RL algorithm, since the compared checkpoints may differ in training data, optimization details, and model lineage. Instead, the goal is diagnostic: to test whether the latent task geometry used by BMC can change after RL-style post-training, and whether those changes follow a consistent pattern. To reduce variation from the tree-construction procedure itself, we fix all random seeds and construct the compared trees on the same machine. Under this setting, tree construction is reproducible across repeated runs, though we do not assume exact reproducibility across different hardware or software environments.

\begin{table}[!h]
\centering
\caption{\textbf{Latent Task Tree structure before and after RL for Qwen3-8B.}
We compare the tree induced by \texttt{Qwen3-8B-Base} before RL training and the corresponding tree after RL, trained using BMC.}
\label{tab:pre_post_rl_tree_structure}
\begin{tabular}{lccc}
\toprule
\textbf{Metric} & \textbf{Pre-RL} & \textbf{Post-RL} & \textbf{Change} \\
\midrule
Total nodes & 50 & 48 & $-2$ \\
Internal nodes & 7 & 7 & $0$ \\
Terminal nodes & 43 & 41 & $-2$ \\
Maximum width & 16 & 20 & $+4$ \\
Maximum depth & 4 & 4 & $0$ \\
Maximum branching factor & 13 & 17 & $+4$ \\
Mean internal branching factor & 7.00 & 6.71 & $-0.29$ \\
\bottomrule
\end{tabular}
\end{table}

\hyperref[tab:pre_post_rl_tree_structure]{Table~\ref{tab:pre_post_rl_tree_structure}} summarizes the structural changes in the Latent Task Tree before and after RL training for \texttt{Qwen3-8B}. The post-RL tree does not become deeper, but it becomes wider at its largest level and exhibits a larger maximum branching factor. This suggests that RL can change the latent organization of task regions by altering the breadth of certain partitions, even when the overall hierarchy depth remains stable.

\begin{table}[!h]
\centering
\caption{\textbf{Latent Task Tree structure before and after RL for coding models.}
We compare the tree induced by \texttt{DeepSeek-R1-Distilled-Qwen-14B} and the corresponding post-RL model, \texttt{DeepCoder-14B-Preview}.}
\label{tab:deepcoder_tree_structure}
\begin{tabular}{lccc}
\toprule
\textbf{Metric} 
& \makecell{\textbf{DeepSeek-R1-Distilled}\\\textbf{(Pre-RL)}} 
& \makecell{\textbf{DeepCoder}\\\textbf{(Post-RL)}} 
& \textbf{Change} \\
\midrule
Total nodes & 21 & 19 & $-2$ \\
Internal nodes & 2 & 2 & $0$ \\
Terminal nodes & 19 & 17 & $-2$ \\
Maximum width & 17 & 15 & $-2$ \\
Maximum depth & 2 & 2 & $0$ \\
Maximum branching factor & 17 & 15 & $-2$ \\
Mean internal branching factor & 10.00 & 9.00 & $-1.00$ \\
\bottomrule
\end{tabular}
\end{table}

We repeat the same analysis for a pre-RL and post-RL model pair that did not use our sampling methods. Specifically, we compare the trees generated for the \texttt{DeepCoder-Preview} dataset using \texttt{DeepSeek-R1-Distilled-Qwen-14B}~\citep{deepseek_r1_2025_pdf} as the pre-RL model and \texttt{DeepCoder-14B-Preview}~\citep{deepcoder2025} as the post-RL model that was trained on the dataset. \hyperref[tab:deepcoder_tree_structure]{Table~\ref*{tab:deepcoder_tree_structure}} summarizes the structural differences between the pre-RL and post-RL coding trees. In contrast to the \texttt{Qwen3-8B} comparison, the post-RL coding tree does not become wider; instead, it is slightly smaller, with fewer terminal nodes and a lower maximum branching factor. Both trees remain shallow, suggesting a modest contraction of the induced latent task partition rather than an increase in hierarchical complexity.

\begin{table}[h]
\centering
\caption{\textbf{Latent Task Tree for  before and after RL for Guru-7B.}
We compare the tree induced by \texttt{Qwen2.5-7B} before RL and the corresponding post-RL model, \texttt{Guru-7B}.}
\label{tab:guru_tree_structure}
\begin{tabular}{lccc}
\toprule
\textbf{Metric} 
& \makecell{\textbf{Qwen2.5-7B}\\\textbf{(Pre-RL)}} 
& \makecell{\textbf{Guru-7B}\\\textbf{(Post-RL)}} 
& \textbf{Change} \\
\midrule
Total nodes & 50 & 56 & $+6$ \\
Internal nodes & 8 & 12 & $+4$ \\
Terminal nodes & 42 & 44 & $+2$ \\
Maximum width & 18 & 13 & $-5$ \\
Maximum depth & 4 & 7 & $+3$ \\
Maximum branching factor & 14 & 13 & $-1$ \\
Mean internal branching factor & 6.13 & 4.58 & $-1.54$ \\
\bottomrule
\end{tabular}
\end{table}

Finally, we compare the tree generated for the \texttt{GURU-92k} dataset using \texttt{Qwen2.5-7B} as the  pre-RL model and \texttt{Guru-7B} as the corresponding post-RL model that was trained on the dataset \citep{cheng2026revisiting}. \hyperref[tab:guru_tree_structure]{Table~\ref*{tab:guru_tree_structure}} shows a third pattern. Unlike the \texttt{Qwen3-8B-Base} comparison, where the largest partition becomes wider, and the \texttt{DeepCoder} comparison, where the tree slightly contracts, the \texttt{Guru-7B} tree becomes deeper and contains more internal nodes. At the same time, its maximum width and maximum branching factor decrease. This suggests a shift from a broader partition toward a more hierarchical decomposition of latent task regions.

Taken together, these comparisons suggest that RL can alter latent task geometry, but not in a single universal direction. Across the three settings, post-RL trees differ in whether they become wider, smaller, or more hierarchical. We therefore treat these measurements as a diagnostic of structure drift, rather than as evidence for a fixed pattern of representational change. Whether these structural changes connect to broader observations about how RLVR alters response diversity and representation geometry is an interesting direction for future work.

For BMC, the practical question is not only whether the latent tree drifts, but whether rebuilding it during training improves curriculum decisions. We therefore treat periodic reconstruction as an empirical design choice rather than an obvious requirement.

\subsection{Periodic Tree Construction Experiment}

To test whether reconstruction is useful in practice, we compare standard BMC, which keeps the Latent Task Tree fixed throughout training, against a variant that periodically reconstructs the tree during RL. If the model's latent representation of prompts drifts in a way that materially affects sampling decisions, then updating the tree should provide a better approximation to the current task geometry and improve downstream performance. Conversely, if the static tree remains sufficiently aligned with the relevant curriculum structure, periodic reconstruction should provide little benefit.

\begin{figure*}[!h]
    \centering
    \includegraphics[width=\textwidth]{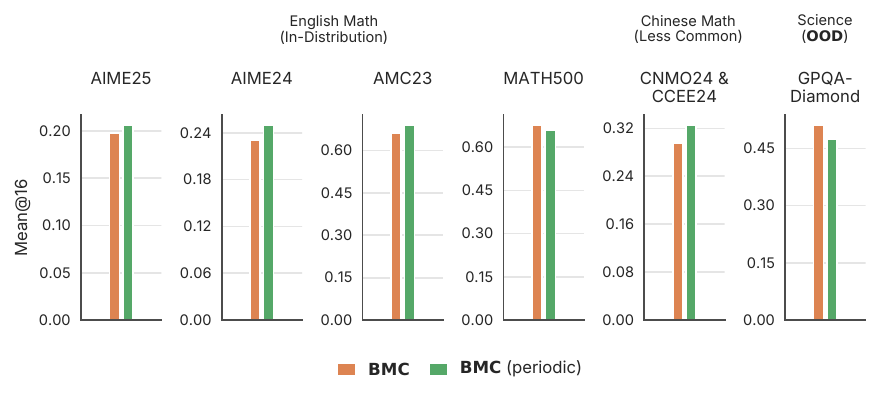}
    \caption{\textbf{Periodic tree reconstruction.}
    We compare standard BMC, which keeps the Latent Task Tree fixed throughout training, against a variant that reconstructs the tree every 100 steps after round-robin initialization.}
    \label{fig:bmc_periodic}
\end{figure*}

\hyperref[fig:bmc_periodic]{Figure~\ref{fig:bmc_periodic}} shows that periodic reconstruction produces only modest differences across evaluations. Reconstruction slightly improves several concentrated math benchmarks, including AIME-style benchmarks, AMC, and the Chinese math aggregate, while slightly reducing performance on MATH500 and GPQA-Diamond. One possible explanation is that reconstruction changes which latent regions are emphasized as the model's representation evolves, benefiting some evaluation distributions while reducing coverage or transfer to others. We treat this interpretation as speculative, but the small magnitude of the changes suggests that the static tree remains a reasonable approximation over the training horizon. We therefore keep the tree fixed in the main experiments for simplicity and stability.
\clearpage % end section
%%%%%%%%%%%%%%%%%%%%%%%%%%%%%%%%%%%%%%%%%%%%%%%%%%%%%%%%%%%%%%%%%%%%%%%%%%%%%%%%%%%%%%%%%%%%%%%%%%%%%%%%%%%%%%%%%%%%%

\section{The Standard Bandit Pattern}
\label{app:standard-bandit-pattern}

The purpose of this section is to clarify why we treat the \textbf{Difficulty Only} ablation as a representative adaptive curriculum learning baseline, rather than attempting to exhaustively compare against every recent adaptive prompt-selection method. We view our work as part of a growing line of adaptive curriculum learning methods for RL, many of which differ in motivation and implementation but share a common productivity-oriented abstraction: \emph{estimate which individual prompts are likely to be informative and sample those prompts more often}. Our goal is not to claim that all instantiations are identical, but to identify the common mechanism they share and evaluate a strong implementation of that mechanism under the same training stack.

This abstraction arises from a common bottleneck in group-relative policy optimization: prompts whose sampled rollouts receive identical rewards provide little or no advantage signal. Dynamic sampling mitigates this issue by resampling until informative groups are found, but can incur substantial wall-clock overhead. Adaptive curriculum learning methods typically try to increase the \emph{effective ratio} (the fraction of sampled prompts that contribute to the gradient) \emph{without} the increase in wall-clock time by predicting useful prompts before sampling.

Across this literature, we refer to the recurring abstraction as the \emph{standard bandit pattern}. We use the term ``bandit pattern'' descriptively rather than literally: a method need not implement a formal UCB or Thompson-sampling algorithm to follow this pattern, as long as it repeatedly assigns scalar values to prompt-level arms and adapts sampling according to those values.

In this pattern, each prompt is treated as an arm whose value estimates how useful the prompt is expected to be for learning. In this work, we refer to this value as the \emph{learning signal}. As discussed in \hyperref[app:learning-signal-quantification]{Appendix~\ref*{app:learning-signal-quantification}}, several common choices are closely related in binary-reward RLVR: reward variance is maximized near a $50\%$ success rate, while expected absolute advantage and related information measures also prioritize prompts with nontrivial rollout-level contrast. Methods may differ in their statistical representation and sampling rule, for example by using Beta or Gaussian distributions, UCB or Thompson Sampling, and value estimates derived from running statistics or critic-style predictors. A scheduler then repeatedly samples prompts according to these values and updates the corresponding estimates from observed rewards.

More concretely, the pattern can be summarized as:
\begin{enumerate}
    \item assign each prompt $x_i$ an estimated learning value $v_i$;
    \item initialize $\{v_i\}_{i=1}^N$ uniformly or from a prior model;
    \item for each training step $t$:
    \begin{enumerate}
        \item sample a batch of individual prompts according to their estimated values, often without replacement;
        \item generate rollouts and observe rewards for the sampled prompts;
        \item update the values of the sampled prompts using the observed rollout rewards;
    \end{enumerate}
    \item repeat as the policy evolves.
\end{enumerate}

The standard bandit pattern is therefore a strong mechanism for improving sampling productivity, but it does not explicitly control several dimensions that can be important when training LLMs on broad, heterogeneous mixtures. In particular, it does not directly control \emph{diversity} over problem types or \emph{utility} relative to a target distribution. If prompts are prioritized only by expected productivity (usually defined in terms of difficulty), then coverage over problem types becomes a byproduct of the productivity objective rather than a controlled property of the scheduler. As a result, a productivity-only sampler can repeatedly allocate effort to a narrow set of high-signal regions even when other, less-represented regions also provide useful learning signal. This becomes especially subtle when the policy-dependent frontier is imbalanced differently from the raw dataset; see \hyperref[app:frontier_imbalance]{Appendix~\ref*{app:frontier_imbalance}}. Similarly, prompts that are highly productive for training may not be the prompts most aligned with a particular target distribution. 

This does not mean that difficulty-only methods are unimportant or solved: better predictors of problem difficulty can still improve the productivity axis. Rather, our point is that productivity remains only one axis of curriculum design, and improving it does not by itself guarantee diversity over problem types or benchmark superiority.

We design Difficulty Only to serve as a strong internal instantiation of the standard bandit pattern under our training stack. Rather than constructing an artificially weak baseline, Difficulty Only preserves the same policy optimization setup as BMC, but removes the latent-task structure and associated structure-aware sampling components. To validate that its behavior is representative of the broader pattern, we compare it against \textbf{MoPPS}~\citep{qu2025can}, an external adaptive curriculum method for RLVR. Like Difficulty Only, MoPPS uses Thompson-style posterior sampling, but with a different belief model and selection rule. Specifically, MoPPS models each prompt's success rate using a Beta posterior,
\[
p_i \sim \mathrm{Beta}(\alpha_i,\beta_i),
\]
samples from each posterior, and selects the top-$B$ prompts whose sampled success probabilities are closest to $0.5$:
\[
\mathcal{B}_t
=
\operatorname{TopB}_{i}
\left(
-\left|p_i - 0.5\right|
\right).
\]
This corresponds to prioritizing prompts near maximum reward variance under binary rewards. After sampling, the posterior parameters are updated from observed rewards. Unlike our Difficulty Only ablation, MoPPS decays posterior statistics only for sampled prompts; unsampled prompts do not become more uncertain over time. Despite these differences, MoPPS follows the same broad standard bandit pattern: it maintains prompt-level value estimates and adaptively samples prompts expected to be informative.

\begin{figure}[t]
  \centering
  \includegraphics[width=0.85\textwidth]{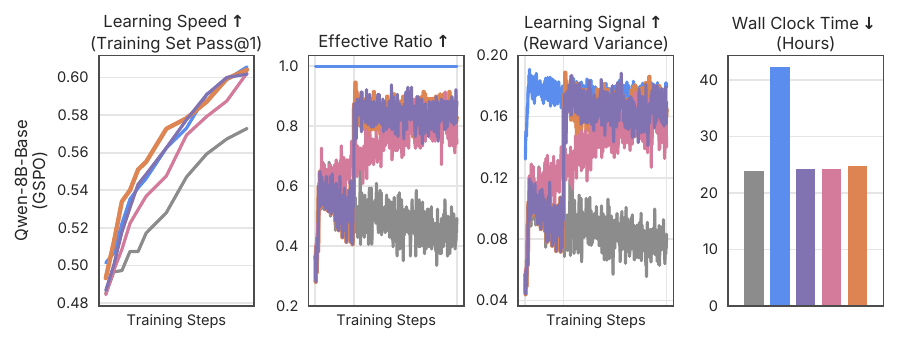}
  \vspace{0.1em}
  \includegraphics[width=\textwidth]{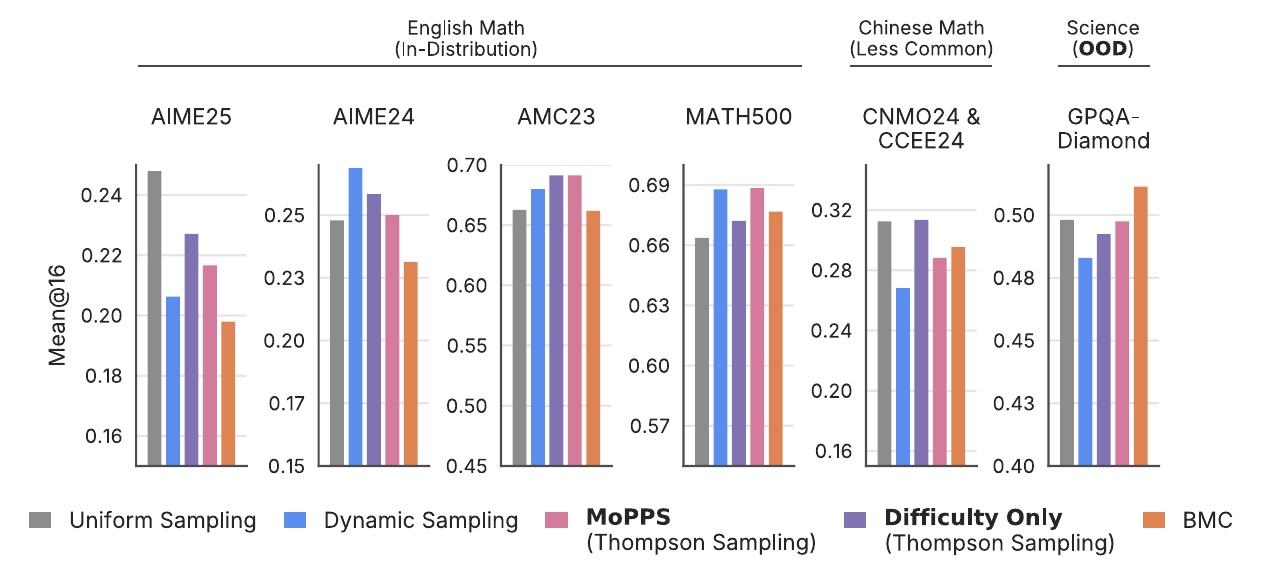}
  \caption{\textbf{Productivity and evaluation behavior under the standard bandit pattern.}
\textbf{Top:} Training productivity metrics on \texttt{DAPO-Math-17K}. MoPPS and the Difficulty Only ablation both improve learning speed and effective ratio relative to uniform sampling while avoiding the wall-clock overhead of dynamic sampling.
\textbf{Bottom:} Evaluation performance across English mathematics, Chinese mathematics, and out-of-distribution science benchmarks. MoPPS and the Difficulty Only ablation exhibit similar qualitative behavior, supporting their use as representative instantiations of the standard bandit pattern. See \hyperref[fig:eval_curve_sbp]{Figure~\ref*{fig:eval_curve_sbp}} for per-step trajectories.}
  \label{fig:sbp}
\end{figure}

We evaluate MoPPS on \texttt{DAPO-Math-17K} and compare it with uniform sampling, dynamic sampling, our Difficulty Only ablation, and BMC. Since MoPPS does not use round-robin initialization by default, we preserve its original sampling behavior rather than modifying it to match our ablations.

\begin{figure*}[h]
  \centering
  \includegraphics[width=\textwidth]{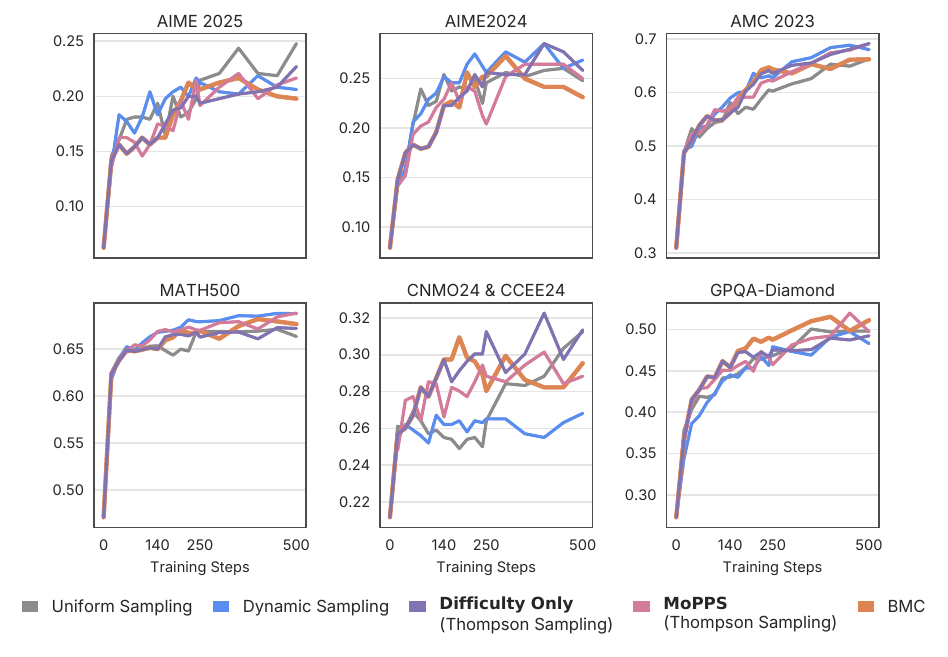}
     \caption{\textbf{Evaluation curves.} Performance across evaluations for \texttt{Qwen3-8B-base} while training on \texttt{DAPO-Math-17k} using \textbf{GSPO}. These evaluation trajectories correspond to the bar graphs visualized in \hyperref[fig:sbp]{Figure~\ref*{fig:sbp}}.}  
\label{fig:eval_curve_sbp}
\end{figure*}

\hyperref[fig:sbp]{Figure~\ref*{fig:sbp}} supports the role of Difficulty Only as a representative baseline for the standard bandit pattern. Although MoPPS and Difficulty Only differ in belief models, update rules, and initialization, both improve training-set learning speed and effective ratio relative to uniform sampling without incurring the wall-clock cost of dynamic sampling. Both also achieve their strongest relative performance on in-distribution English mathematics evaluations, where productivity-oriented sampling is well aligned with the evaluation distribution.

At the same time, both methods illustrate the limitations of productivity-only curricula. They perform better than dynamic sampling on the less-represented Chinese mathematics evaluations, consistent with the concern that generate-then-discard strategies can skip problem regions where the current policy receives little immediate learning signal. However, both standard-bandit methods perform lower on average than BMC on the out-of-distribution GPQA-Diamond benchmark\footnote{MoPPS briefly exceeds BMC on one intermediate GPQA-Diamond evaluation. We emphasize the overall trajectory, rather than isolated pointwise crossings, when interpreting this result.}. Moreover, despite achieving higher effective ratios and absorbing more of the training data, they fall behind uniform sampling on AIME2025, suggesting that brute-force coverage can sometimes expose the model to high-utility regions that productivity-oriented sampling underweights.

We therefore view the standard bandit pattern as an important and effective baseline for adaptive curriculum learning in RLVR. It captures the dominant productivity-oriented mechanism in this line of work, and different instantiations can produce similar qualitative behavior. However, these results reinforce the central claim of this work: productivity, or ``difficulty awareness,'' is not a complete model of curriculum design. When most of the training distribution has high utility for the target evaluations, methods that follow the standard bandit pattern may be sufficient. However, as the training mixture becomes broader and more heterogeneous, and as target evaluations emphasize different regions of the task manifold, productivity-oriented curricula become increasingly vulnerable to train-test mismatch unless diversity and utility are explicitly controlled.

BMC generalizes the standard bandit pattern by changing the action space over which curriculum decisions are made: from independent prompt-level arms to structured regions of a latent task manifold. This allows the scheduler to exploit structural correlations between prompts, share evidence across related regions, and reason about sampling decisions at multiple levels of abstraction. As a result, \emph{diversity} over latent problem types becomes an explicit part of the sampling process rather than an incidental byproduct of prompt-level exploration. BMC-T further extends this view by incorporating \emph{utility} (evaluation relevance), using latent proximity to bias sampling toward regions expected to support a target evaluation distribution. Together, these extensions move beyond asking only how much learning signal a prompt provides, and instead ask which regions of the task manifold should receive training effort, how evidence should propagate across related tasks, and which downstream evaluations that effort is likely to support.

\clearpage % end section
%%%%%%%%%%%%%%%%%%%%%%%%%%%%%%%%%%%%%%%%%%%%%%%%%%%%%%%%%%%%%%%%%%%%%%%%%%%%%%%%%%%%%%%%%%%%%%%%%%%%%%%%%%%%%%%%%%%%%

\section{Medical Data Experiments}
\label{app:medical}

To further evaluate the domain generality of Latent Task Trees and BMC, we run an additional experiment on the medical question-answering dataset \texttt{AlphaMed19K}~\citep{liu2025distillationpushinglimitsmedical}. We use \texttt{Qwen3-8B} with the same training hyperparameters as in the main experiments; see \hyperref[app:hyperparameters]{Appendix~\ref*{app:hyperparameters}}. This experiment is intended as a domain-transfer stress test rather than a fully tuned medical-training recipe.

\begin{figure*}[!h]
    \centering
    \includegraphics[width=\textwidth]{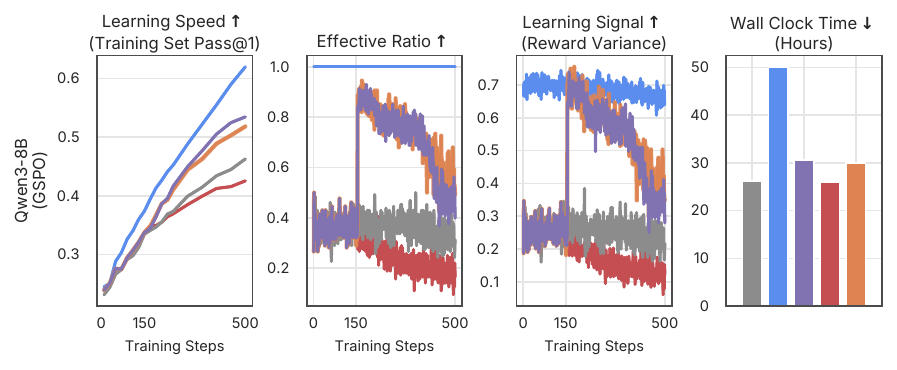}
    \includegraphics[width=0.8\textwidth]{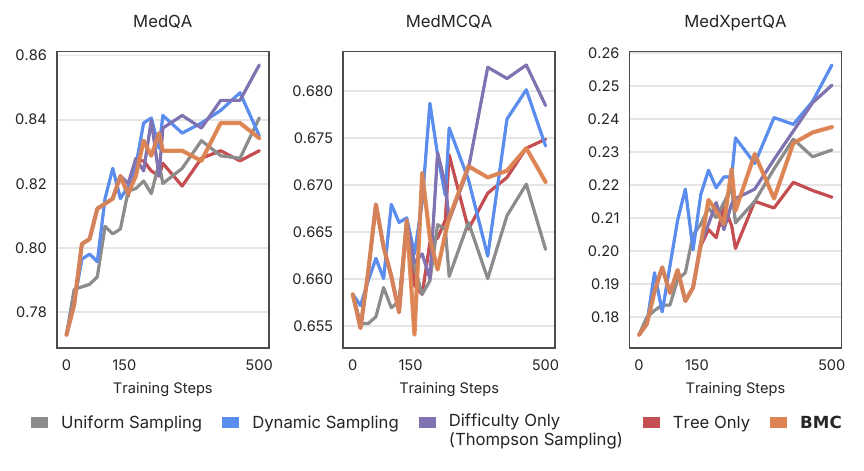}
   \caption{\textbf{Training and evaluation dynamics on AlphaMed19K.}
\textbf{Top:} Training productivity metrics for Uniform Sampling, Dynamic Sampling, Difficulty Only, Tree Only, and BMC. Dynamic Sampling maintains the highest effective ratio and reward variance, but incurs substantially higher wall-clock time due to repeated filtering. BMC and Difficulty Only improve learning speed (training-set pass@1) relative to Uniform Sampling, even though their effective ratio and observed learning signal decline after adaptive sampling begins.
\textbf{Bottom:} Medical evaluation performance on MedQA, MedMCQA, and MedXpertQA. Evaluation performance is strongest for Dynamic Sampling and Difficulty Only, while BMC's structure-aware sampling does not translate into the strongest downstream accuracy in this setting. This reinforces the tradeoffs between productivity, diversity, and utility.}
    \label{fig:alphamed_metrics}
\end{figure*}

\paragraph{Productivity.}
\hyperref[fig:alphamed_metrics]{Figure~\ref*{fig:alphamed_metrics}} shows that \texttt{AlphaMed19K} exhibits different training dynamics from the main \texttt{DAPO-Math-17K} experiments. After the round-robin initialization phase, the effective ratio and observed learning signal decline over training for both BMC and Difficulty Only. Nevertheless, both methods improve training-set pass@1 relative to Uniform Sampling, indicating that adaptive prompt selection still identifies useful regions of the dataset early in training.

Dynamic Sampling maintains the highest effective ratio and reward variance by construction, since it filters candidate prompts until it forms an effective composite batch. However, this comes at a substantial wall-clock cost. In the main mathematics experiments, Dynamic Sampling required roughly two filtering iterations on average to construct an effective batch; on \texttt{AlphaMed19K}, it required roughly three. This suggests that, for this model--dataset pair, productive prompts are sparser or become depleted more quickly, forcing generate-then-discard sampling to spend more computation searching for informative examples.

\paragraph{Utility.}
For evaluation, we partially follow the original medical question-answering setup from \citet{liu2025distillationpushinglimitsmedical} and evaluate on MedQA~\citep{jin2020diseasedoespatienthave}, MedMCQA~\citep{pal2022medmcqalargescalemultisubject}, and MedXpertQA~\citep{zuo2025medxpertqabenchmarkingexpertlevelmedical}. Due to the larger size of these evaluation sets, we report pass@1 rather than mean@16.

The evaluation results show that adaptive sampling remains effective in the medical domain and exhibits a pattern consistent with the main \texttt{DAPO-Math-17K} experiments. Dynamic Sampling and Difficulty Only are strongest on several medical evaluations, while BMC remains competitive but is not uniformly best. This mirrors the broader theme of the paper: productivity, diversity, and utility are distinct axes of curriculum design, and optimizing for one does not guarantee optimality along the others. As in the main experiments, this suggests that downstream evaluation performance may depend on how well structure-aware coverage aligns with the particular capabilities measured by the evaluation suite.

One possible explanation is that the Latent Task Tree derived from \texttt{AlphaMed19K} exhibits a more fine-grained structure than the tree derived from \texttt{DAPO-Math-17K} (see \hyperref[fig:treebreakdown_alphamed]{Figure~\ref*{fig:treebreakdown_alphamed}}). In such settings, structure-aware sampling may allocate effort across semantically distinct regions whose relevance to a fixed evaluation suite is uneven or difficult to infer from productivity alone. This interpretation is suggestive rather than causal, but it reinforces the motivation for target-aware variants such as BMC-T, where latent structure can be combined with an explicit utility signal.

Overall, these experiments further support the domain generality of Latent Task Tree construction and BMC, while reinforcing the central point that curriculum design should consider more than difficulty alone. On \texttt{AlphaMed19K}, BMC improves over Uniform Sampling in training productivity and approaches Dynamic Sampling without incurring the same wall-clock overhead, but Difficulty Only and Dynamic Sampling achieve stronger performance on several medical evaluations. This should not be interpreted as evidence that structure-aware sampling is domain-specific or ineffective in medical question answering. Rather, it provides another example in which productivity, diversity, and evaluation utility need not align automatically. This makes \texttt{AlphaMed19K} a useful case for target-aware variants such as BMC-T, where structure-aware coverage can be biased toward regions more directly connected to a target evaluation distribution or desired capability.

\paragraph{Base Model Attempt.}
We also attempted to run the same experiment with the base \texttt{Qwen3-8B-Base} model. However, despite testing multiple prompt templates, we found that the base model often converged to producing only the multiple-choice answer without the reasoning format required by our RL setup. The instruction-tuned model did not exhibit this issue. As a result, we use the instruction-tuned model for the reported \texttt{AlphaMed19K} experiment. This leaves open whether the observed dynamics are primarily due to the medical domain, the strength of the instruction-tuned initialization, or the interaction between the model and dataset.

%%%%%%%%%%%%%%%%%%%%%%%%%%%%%%%%%%%%%%%%%%%%%%%%%%%%%%%%%%%%%%%%%%%%%%%%%%%%%%%%%%%%%%%%%%%%%%%%%%%%%%%%%%%%%%%%%%%%%
\clearpage
\section{Tree Visualizations \& Statistics}
\label{app:tree_breakdowns}

\begin{figure*}[p]
    \centering
    \makebox[\textwidth][c]{%
        \includegraphics[width=1.30\textwidth]{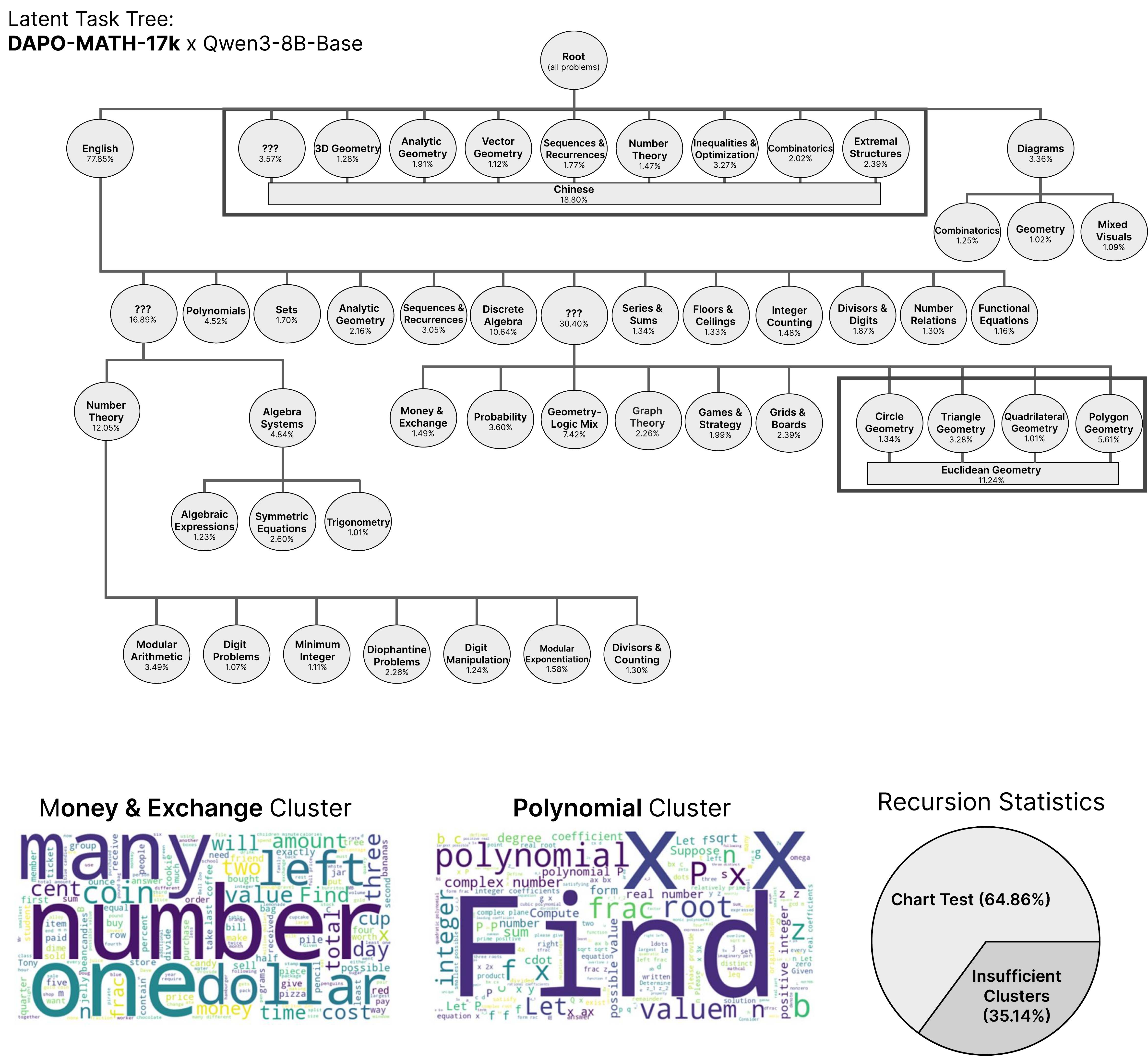}
    }
    \vspace{-0.5em}
    \caption{
    \textbf{Latent Task Tree for \texttt{DAPO-Math-17k} using Qwen3-8B-Base's latent space.}
    The dataset contains mathematical reasoning problems in both Chinese and English.
    Themes such as \emph{Chinese} and \emph{Euclidean Geometry} are not tree nodes, but descriptive annotations used to summarize the condensed view of the full tree shown in \hyperref[fig:ltt]{Figure~\ref*{fig:ltt}}.
    The induced tree contains $50$ nodes.
    On one node with $8\times$H100 GPUs, loading the latent representations took 5:35 minutes and recursive tree construction took 6:29 minutes, for a total construction time of 12:04 minutes.
    Since the corresponding training run took 25.06 hours (\emph{including} the tree construction time), the tree construction accounted for only $0.802\%$ of the total training time.
    }
    \label{fig:treebreakdown_dapomath}
\end{figure*}

\begin{figure*}[p]
    \centering
    \makebox[\textwidth][c]{%
        \includegraphics[width=1.30\textwidth]{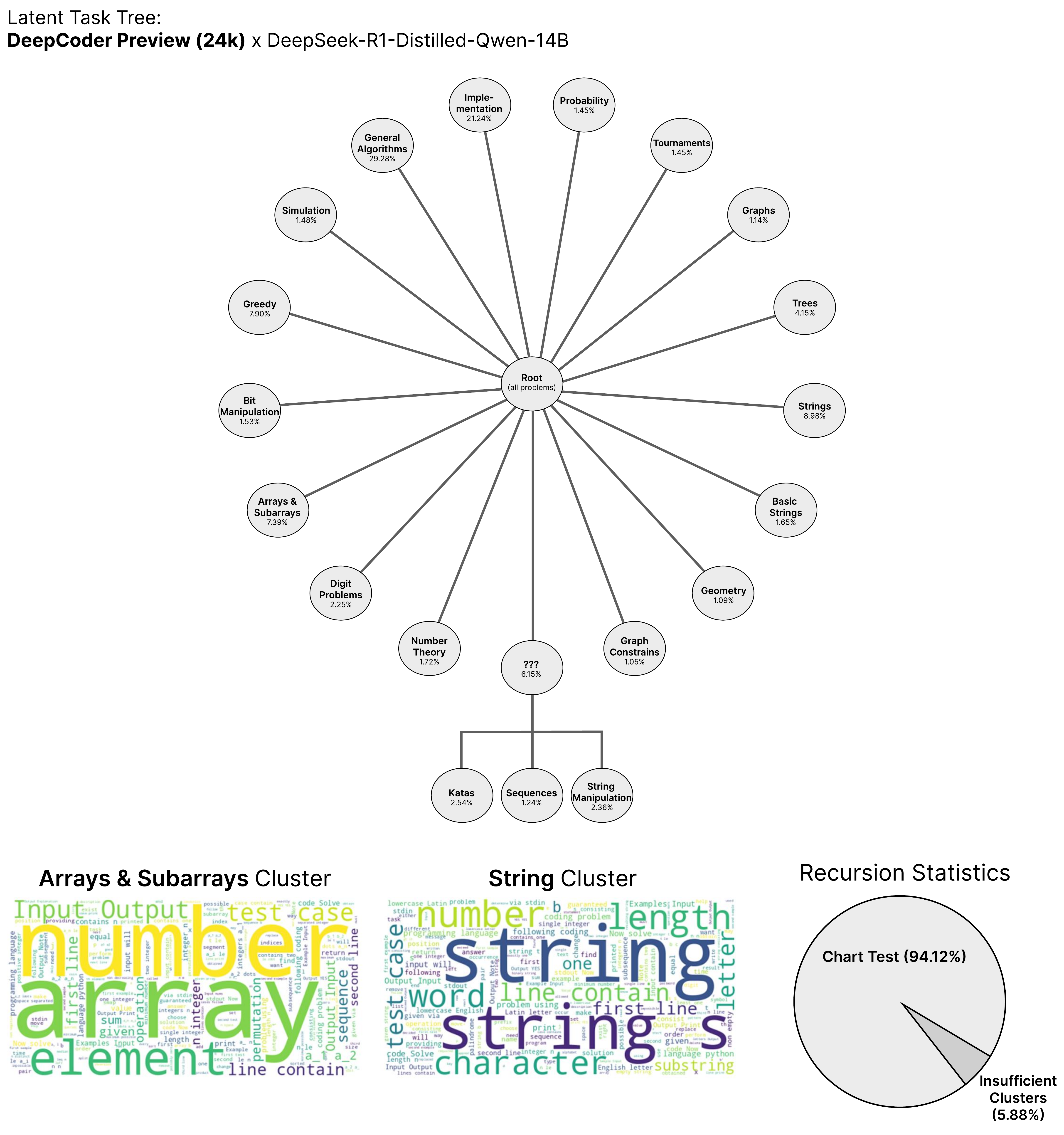}
    }
    \vspace{-0.5em}
    \caption{
    \textbf{Latent Task Tree for \emph{DeepCoder-Preview-Dataset} \citep{deepcoder2025} using DeepSeek-R1-Distilled-Qwen-14B's latent space.}
    The dataset consists of 24K coding problems. The induced tree contains $49$ nodes. On one node with $8\times$H100 GPUs, loading the latent representations took 12:56 minutes and recursive tree construction took 7:22 minutes, for a total construction time of 20:18 minutes.
    }
    \label{fig:treebreakdown_deepcoder}
\end{figure*}

\begin{figure*}[p]
    \centering
    \makebox[\textwidth][c]{%
        \includegraphics[width=1.30\textwidth]{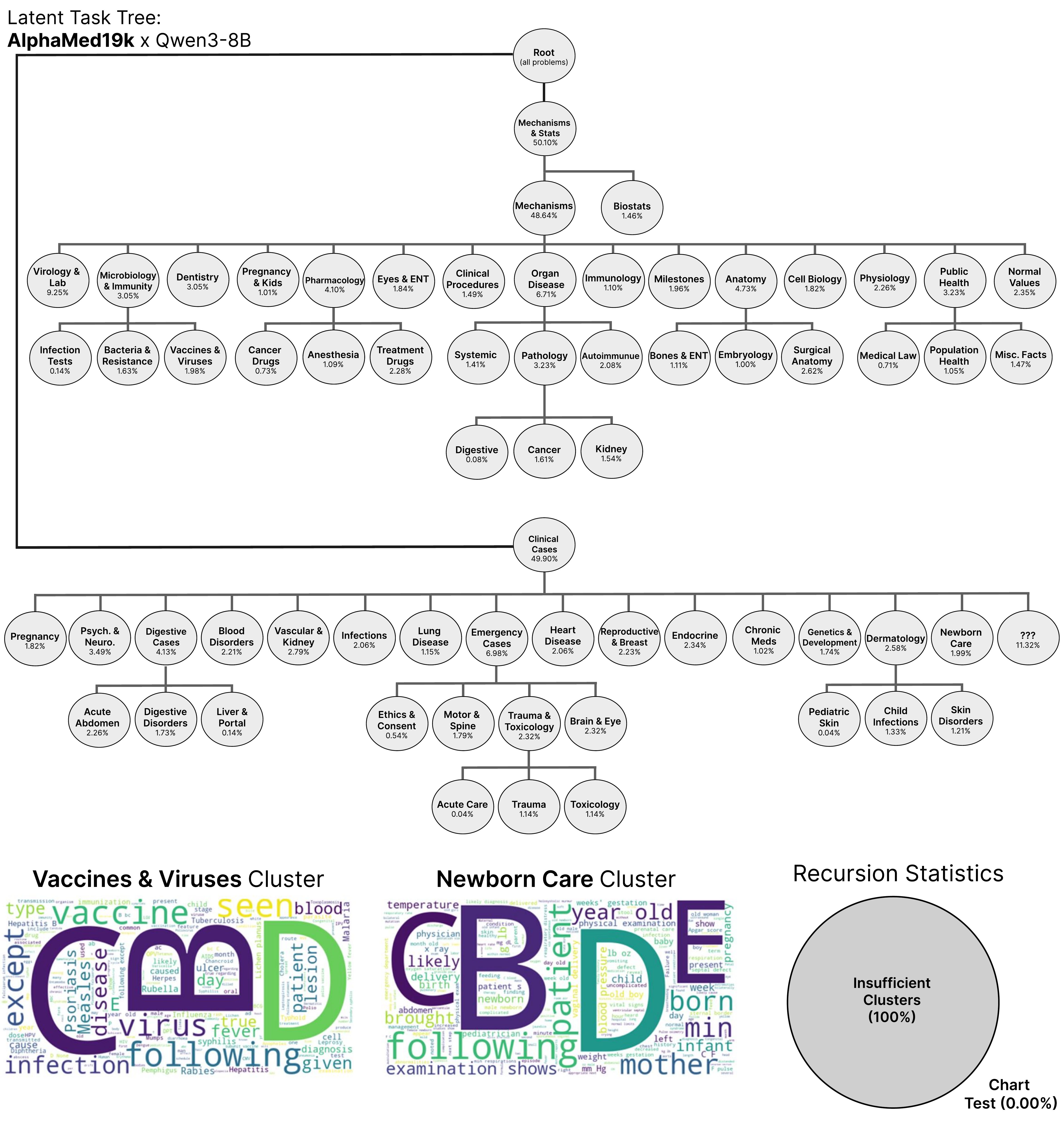}
    }
    \vspace{-0.5em}
    \caption{
    \textbf{Latent Task Tree for \emph{AlphaMed19k} \citep{liu2025distillationpushinglimitsmedical} using Qwen3-8B's latent space.}
    The dataset contains medical Q\&A multiple-choice problems across varying topics.
    The induced tree contains $67$ nodes.
    On one node with $8\times$H100 GPUs, loading the latent representations took 10:07 minutes and recursive tree construction took 10:46 minutes, for a total construction time of 20:53 minutes.
    Since the corresponding training run (see \hyperref[app:medical]{Appendix~\ref*{app:medical}}) took 30.30 hours (\emph{including} tree construction time), tree construction accounted for only $1.15\%$ of the total training time.
    }
    \label{fig:treebreakdown_alphamed}
\end{figure*}

\begin{figure*}[p]
    \centering
    \makebox[\textwidth][c]{%
        \includegraphics[width=1.30\textwidth]{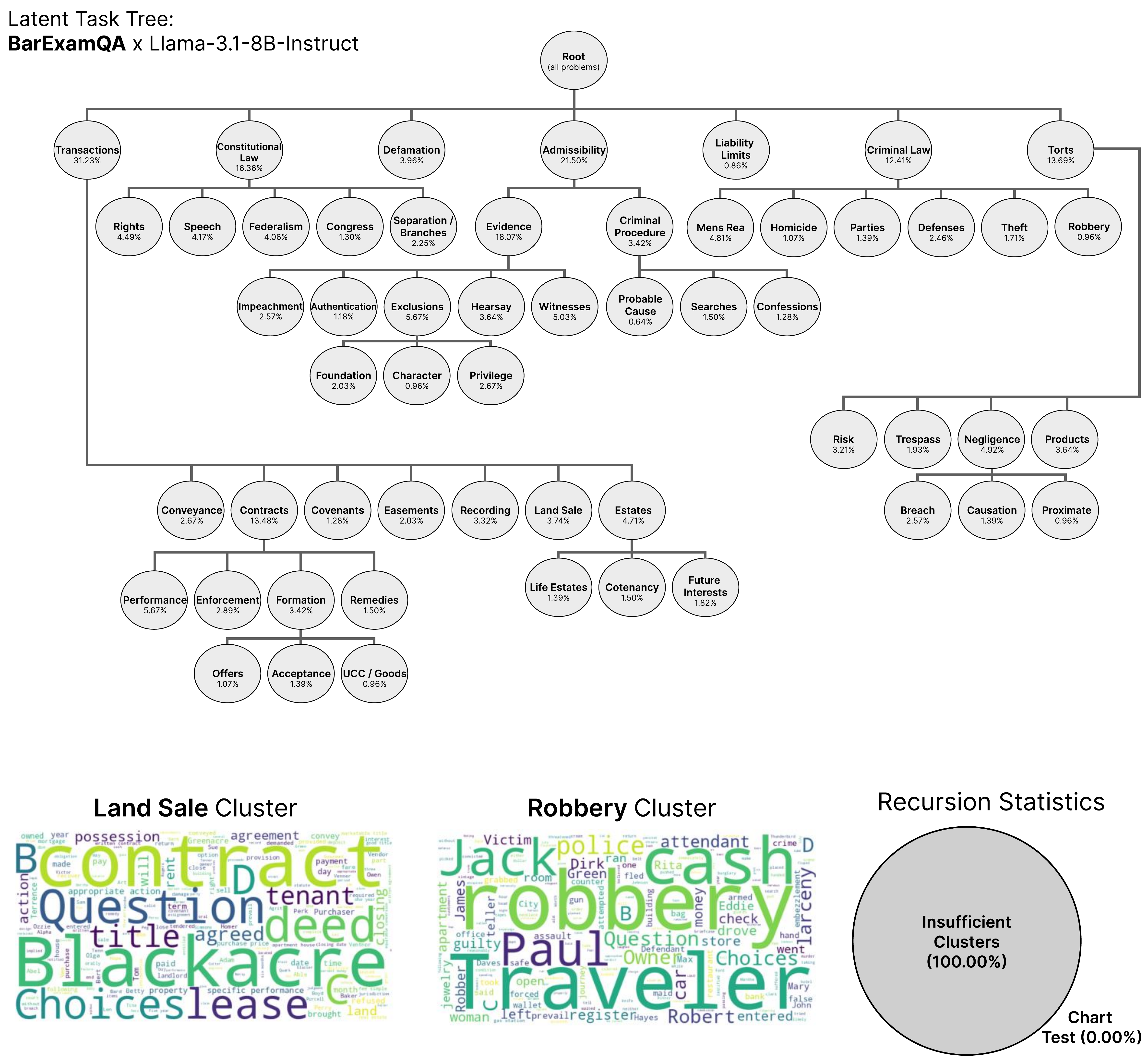}
    }
    \vspace{-0.5em}
    \caption{
    \textbf{Latent Task Tree for \emph{BarExamQA} \citep{Zheng_2025} using the latent space of Llama-3.1-8B-Instruct \citep{grattafiori2024llama3herdmodels}.}
    The dataset consists of ~1K multistate bar exam (MBE) questions. The induced tree contains $56$ nodes. On one node with $8\times$H100 GPUs, loading the latent representations took 52 seconds and recursive tree construction took 51.5 seconds, for a total construction time of 1:44 minutes.
    }
    \label{fig:treebreakdown_barexam}
\end{figure*}

\begin{figure*}[p]
    \centering
    \makebox[\textwidth][c]{%
        \includegraphics[width=1.30\textwidth]{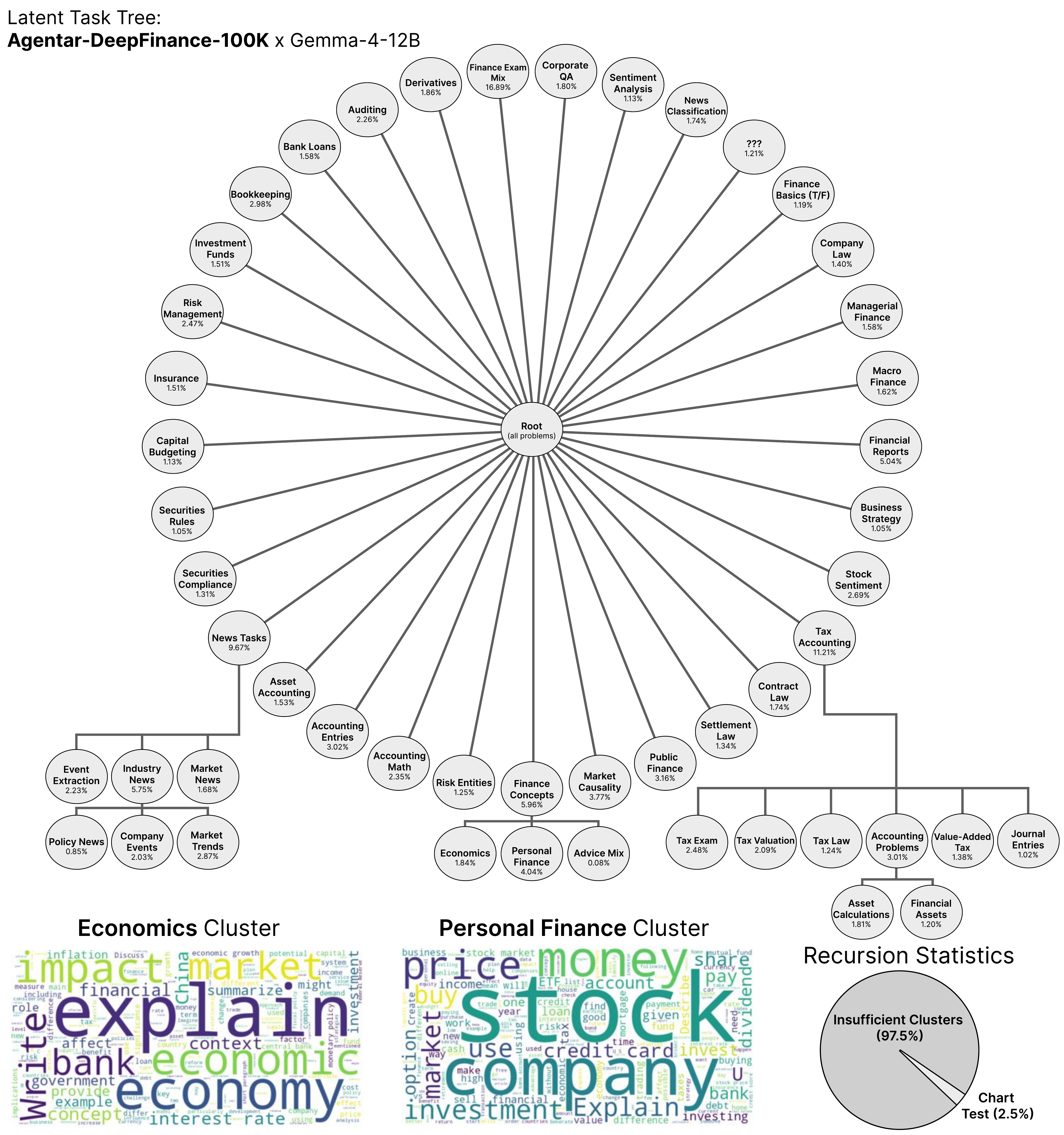}
    }
    \vspace{-0.5em}
    \caption{
    \textbf{Latent Task Tree for \emph{Agentar-DeepFinance-100K} \citep{zhao2025agentardeepfinance100klargescalefinancialdataset} using the latent space of Gemma-4-12B \citep{gemmateam2024gemmaopenmodelsbased}}
    The dataset consists of 100K financial reasoning problems. The dataset is predominantly Chinese. A very small minority of prompts appear to be intended for SFT rather than RLVR. The induced tree contains $51$ nodes. On one node with $8\times$H100 GPUs, loading the latent representations took 58:45 minutes and recursive tree construction took 48:19 minutes, for a total construction time of 1:47:04 hours.
    }
    \label{fig:treebreakdown_deepfinance}
\end{figure*}

\begin{figure*}[p]
    \centering
    \makebox[\textwidth][c]{%
        \includegraphics[width=1.30\textwidth]{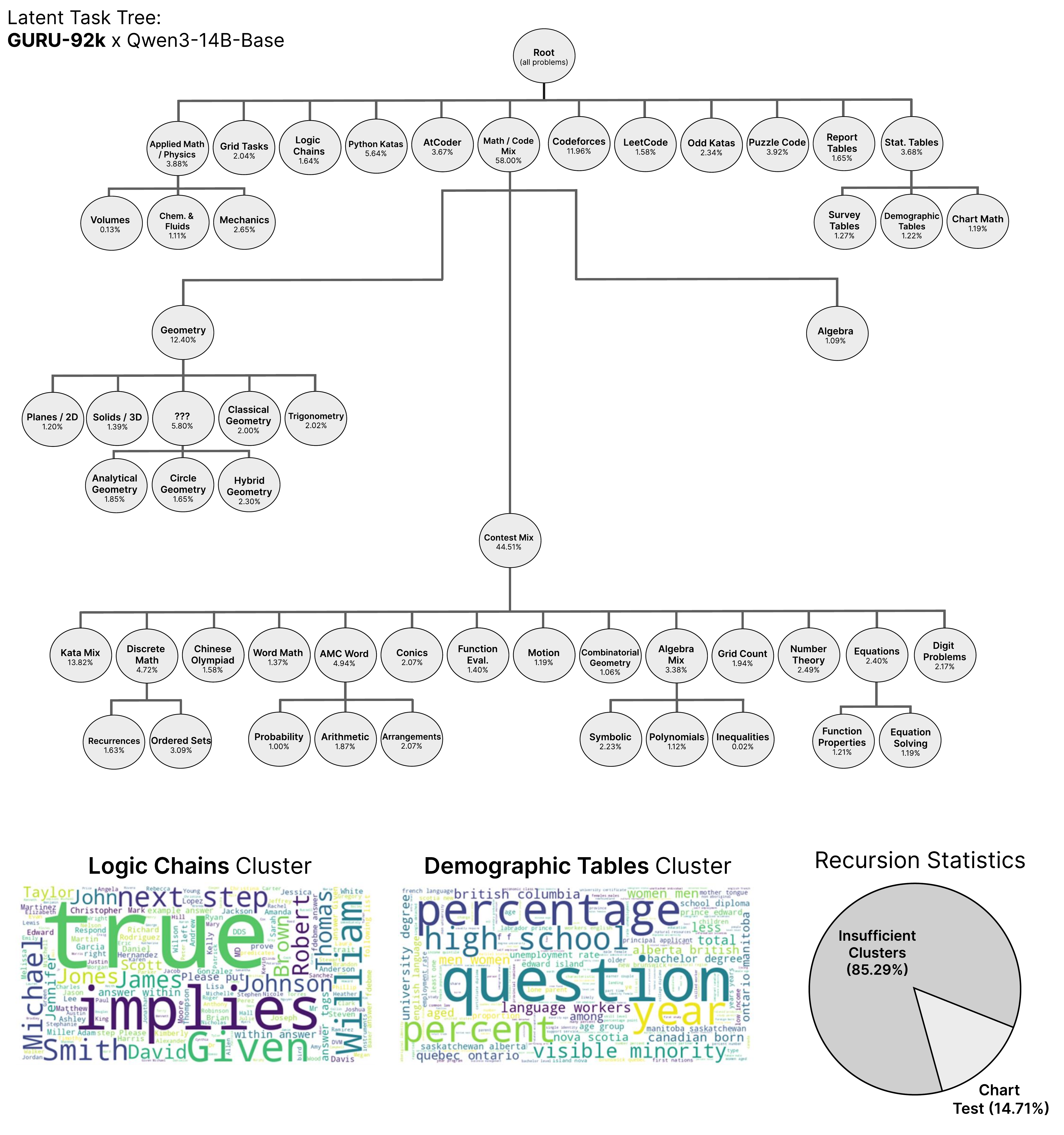}
    }
    \vspace{-0.5em}
    \caption{
    \textbf{Latent Task Tree for \emph{GURU-92k} \citep{cheng2026revisiting} using Qwen3-14B-Base's latent space.}
    \emph{GURU-92k} is a multi-domain reasoning dataset spanning math, code, science, logic, simulation, and tabular reasoning problems, allowing us to test whether latent task trees can organize heterogeneous training data beyond a single domain. The induced tree contains $54$ nodes.
    On one node with $8\times$H100 GPUs, loading the latent representations took 1:25:16 hours and recursive tree construction took 52:31 minutes, for a total construction time of 2:17:53 hours.
    For scale, the original work reports a training wall-clock time of three days; thus, although this tree is more expensive to construct than those for smaller datasets, its construction time is still on the order of hours rather than days.
}
    \label{fig:treebreakdown_guru}
\end{figure*}

\begin{figure*}[p]
    \centering
    \makebox[\textwidth][c]{%
        \includegraphics[width=1.30\textwidth]{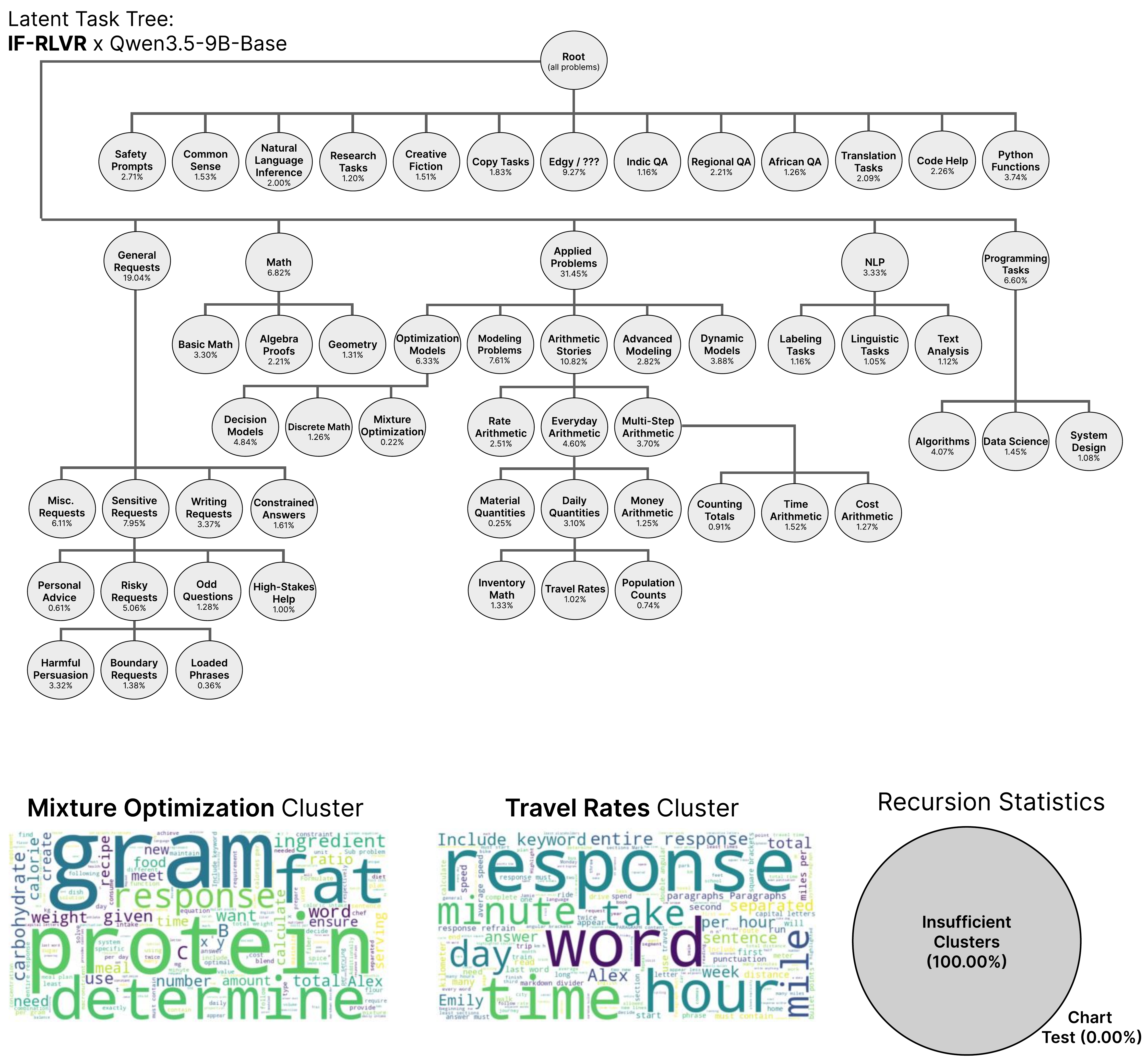}
    }
    \vspace{-0.5em}
    \caption{
    \textbf{Latent Task Tree for \emph{IF-RLVR} \citep{pyatkin2025generalizingverifiableinstructionfollowing} using the latent space of Qwen3.5-9B-Base \citep{qwen3.5}}
    The dataset consists of 95K instruction-following problems spanning mathematics, coding, NLP tasks, multilingual QA, writing, safety-sensitive requests, and other domains. Each prompt pairs a base task with automatically verifiable output constraints, so the induced hierarchy reflects both semantic topic and instruction-following form. The induced tree contains 59 nodes. On one node with (8$\times$) H100 GPUs, loading the latent representations took 48:37 minutes and recursive tree construction took 48:32 minutes, for a total construction time of 1:37:09 hours.
    }
    \label{fig:treebreakdown_if}
\end{figure*}

\begin{figure*}[p]
    \centering
    \makebox[\textwidth][c]{%
        \includegraphics[width=1.30\textwidth]{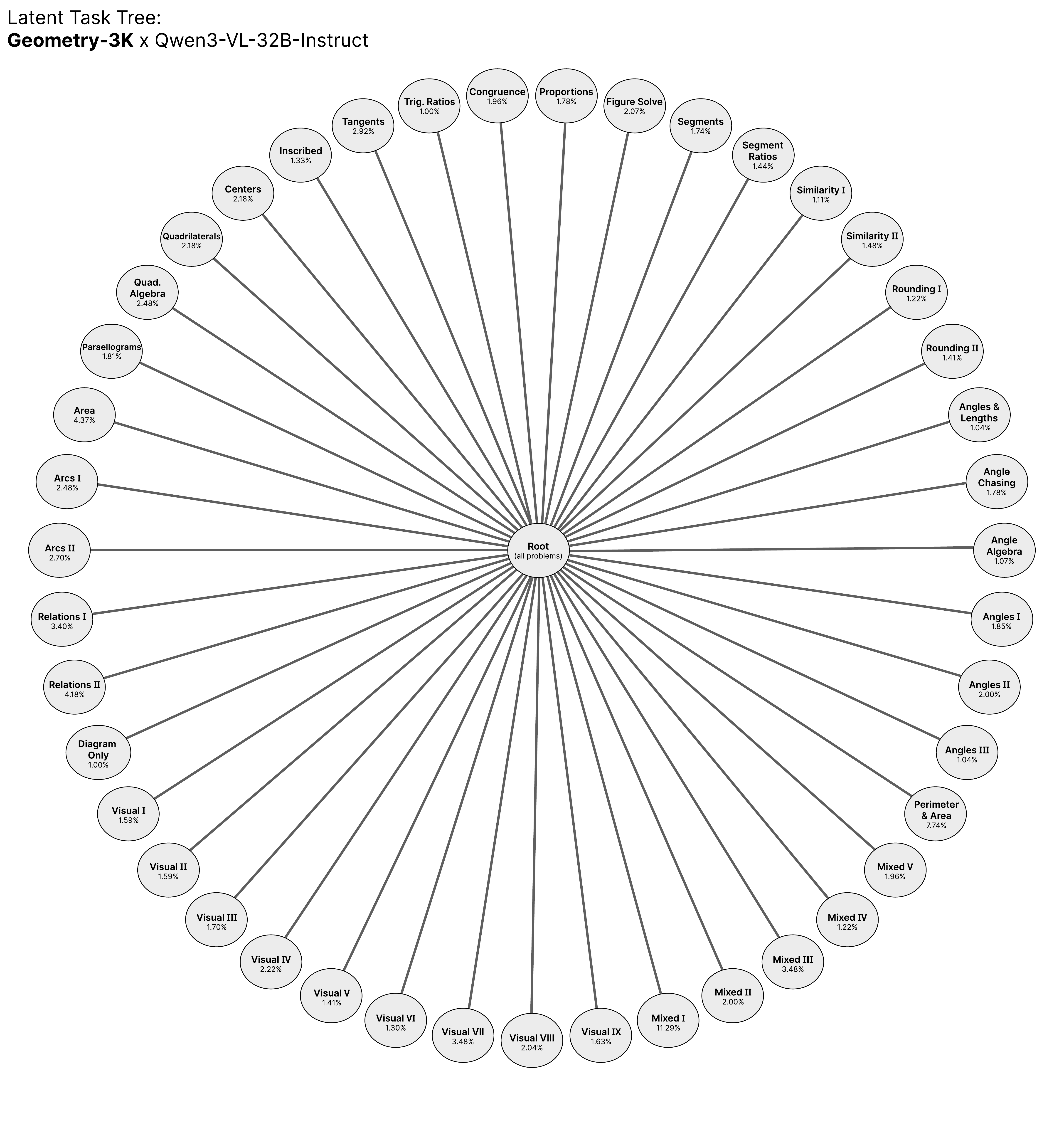}
    }
    \vspace{-0.5em}
    \caption{
    \textbf{Latent Task Tree for \emph{Geometry3K} \cite{lu2021inter} using Qwen3-VL-32B-Instruct's \citep{bai2025qwen3vltechnicalreport} latent space.}
    The dataset consists of multimodal geometry problems, and the induced tree contains $54$ nodes.
    On one node with $8\times$H100 GPUs, loading the latent representations took 3:26 minutes, and recursive tree construction took 2:21 minutes, for a total construction time of 5:47 minutes.
    Because our cluster labels are assigned from example prompts and word clouds, several clusters with minimal or ambiguous text could not be cleanly differentiated without inspecting the associated diagrams.
    These cases suggest that the induced structure is not driven by language alone, but also reflects visual and multimodal properties of the problems.
    A finer-grained visual audit could further refine these labels, but the prompt-level analysis already shows that the latent tree captures structure beyond surface text.
    }
    \label{fig:treebreakdown_geo}
\end{figure*}

To demonstrate the generality of Latent Task Tree construction, we provide full tree visualizations and construction statistics for the following datasets:
\begin{itemize}
    \item \hyperref[fig:treebreakdown_dapomath]{\textbf{DAPO-Math-17K}} \citep{yu2025dapo}: a multilingual math dataset used in our main experiments.
    \item \hyperref[fig:treebreakdown_deepcoder]{\textbf{DeepCoder-Preview}} \citep{deepcoder2025}: a coding dataset.
    \item \hyperref[fig:treebreakdown_alphamed]{\textbf{AlphaMed19K}} \citep{liu2025distillationpushinglimitsmedical}: a medical reasoning dataset.
    \item \hyperref[fig:treebreakdown_barexam]{\textbf{BarExamQA}} \citep{Zheng_2025}: a legal reasoning dataset.
    \item \hyperref[fig:treebreakdown_deepfinance]{\textbf{Agentar-DeepFinance-100k}} \citep{zhao2025agentardeepfinance100klargescalefinancialdataset}: a financial reasoning dataset.
    \item \hyperref[fig:treebreakdown_guru]{\textbf{GURU-92k}} \citep{cheng2026revisiting}: a multi-domain dataset spanning math, code, science, logic, simulation, and tabular reasoning.
    \item \hyperref[fig:treebreakdown_if]{\textbf{IF-RLVR}} \citep{pyatkin2025generalizingverifiableinstructionfollowing}: a verifiable instruction-following dataset spanning mathematics, coding, NLP tasks, multilingual QA, writing, safety-sensitive requests, and other open-ended user requests, with prompts combining base tasks and automatically checkable output constraints. 
    \item \hyperref[fig:treebreakdown_geo]{\textbf{Geometry3K}} \citep{lu2021inter}: a multimodal vision-language math dataset.
\end{itemize}

\textbf{All trees were constructed using the same hyperparameters reported in \hyperref[app:hyperparameters]{Appendix~\ref*{app:hyperparameters}}}.

The tree structures are determined algorithmically, but the displayed node labels are interpretive annotations. To assign labels, we inspected $10$ representative prompts and word clouds for each cluster, using ChatGPT to assist with summarization. These labels are necessarily coarse due to space constraints, and clusters without a clear dominant theme are denoted by ``???''. Each visualization is paired with example word clouds and recursion statistics, which summarize why recursive calls terminated: either because a node passed the chart test or because it failed to meet the required clustering criteria (``insufficient clusters''). 

Although latent task trees can be useful for interpreting datasets, node labeling is not required to use BMC. The algorithm operates directly on the unlabeled tree structure induced from the model's latent representations. The manual labels are included only to make the induced semantic organization visible to the reader, demonstrating that coherent task groupings emerge from raw LLM embeddings without requiring external embedding models or hand-designed taxonomies.

Finally, while fixing the random seed makes tree construction reproducible on a given machine, exact trees may vary across machines due to implementation- and hardware-level nondeterminism. We expect the broad semantic organization to remain similar, but the exact hierarchy may not be identical.

\vspace*{1.5cm}
\textbf{The tree visualizations are shown on the following pages.}

\clearpage

\clearpage

\begin{figure*}[!h]
    \centering
    \includegraphics[width=1.20\textwidth]{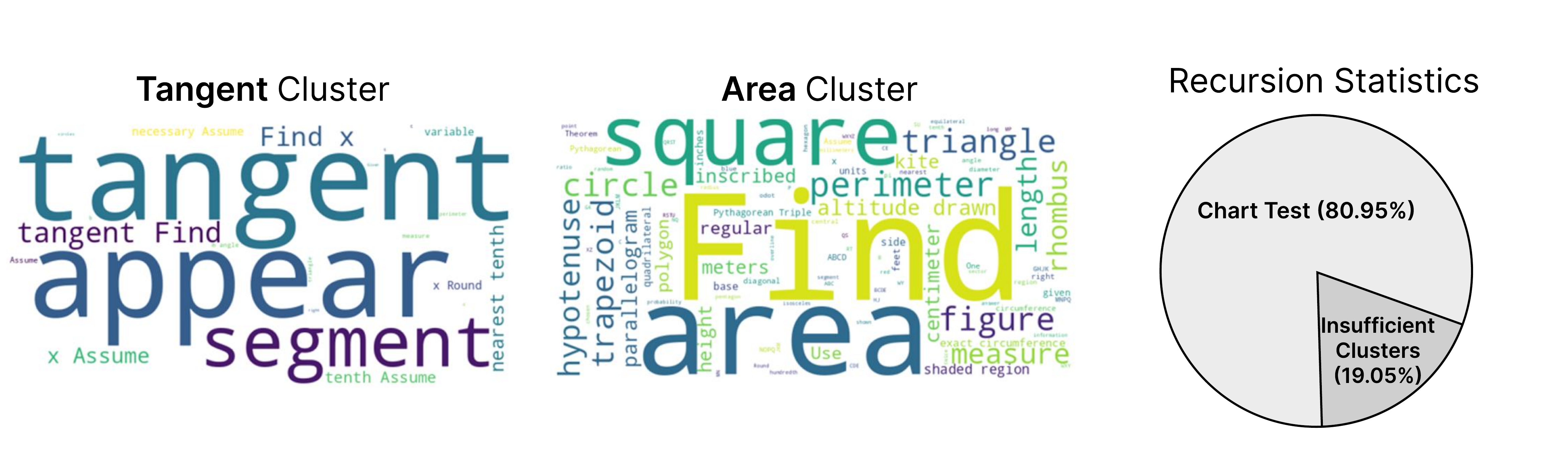}
    \caption{\textbf{Additional information for \hyperref[fig:treebreakdown_geo]{Figure~\ref*{fig:treebreakdown_geo}}}.
    Example word clouds and recursion statistics shown separately from the original figure to due space constraints.}
    \label{fig:treebreakdown_geo_extra}
\end{figure*}

\clearpage % end section
%%%%%%%%%%%%%%%%%%%%%%%%%%%%%%%%%%%%%%%%%%%%%%%%%%%%%%%%%%%%%%%%%%%%%%%%%%%%%%%%%%%%%%%%%%%%%%%%%%%%%%%%%%%%%%%%%%%%%

\clearpage
\section{Frontier Imbalance}
\label{app:frontier_imbalance}
\begin{figure*}[!h]
    \centering
    \includegraphics[width=\textwidth]{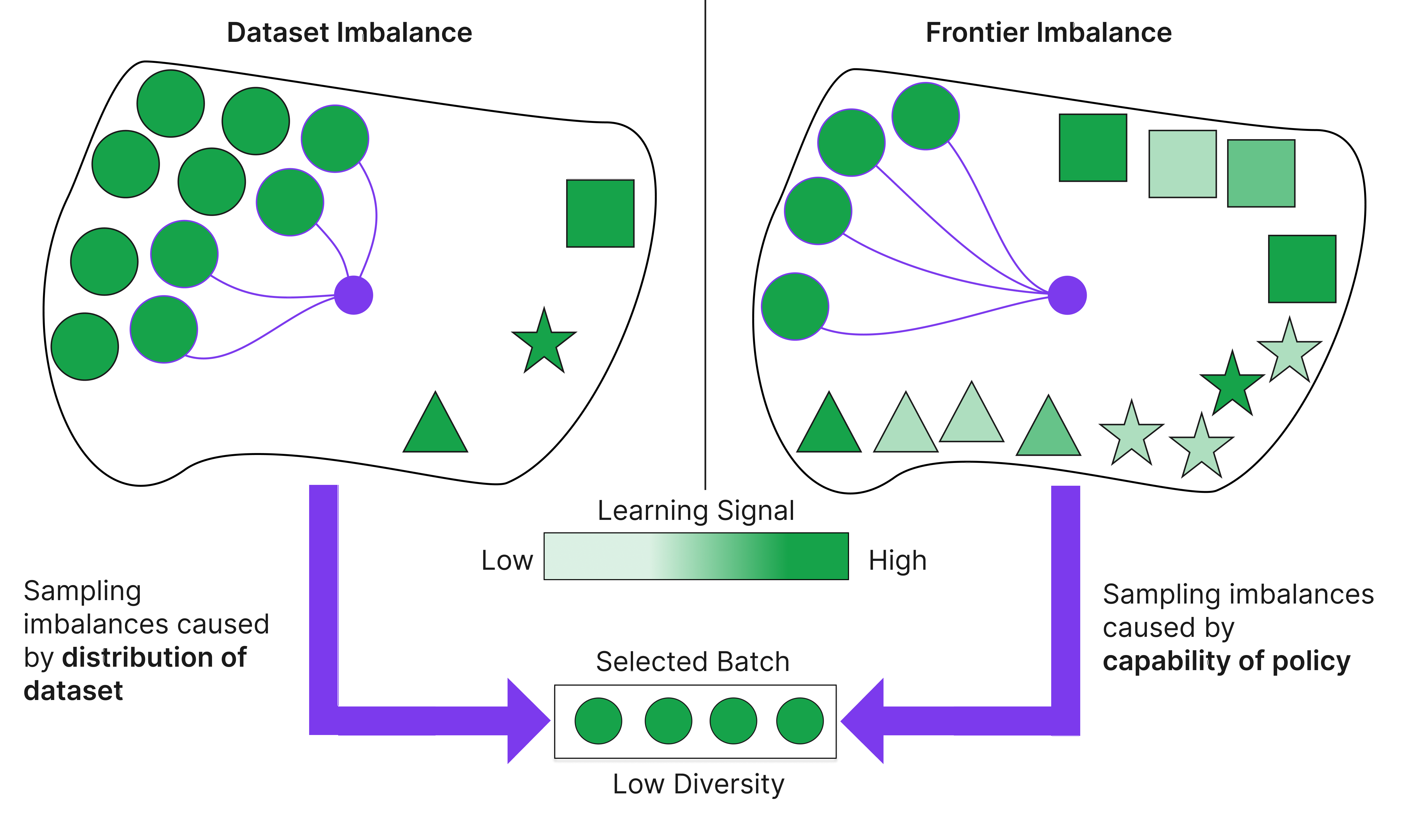}
    \caption{\textbf{Dataset imbalance and frontier imbalance.}
    Dataset imbalance arises from the static composition of the training set, while frontier imbalance arises from the policy-dependent distribution of problems that currently provide learning signal. A productivity-oriented sampler may therefore concentrate on a subset of problem types either because they are common in the dataset, overrepresented on the current frontier, or both.}
    \label{fig:frontier_imbalance}
\end{figure*}

Throughout this work, we emphasize that RL training datasets are heterogeneous and often conceptually imbalanced. If a dataset contains many more examples of one problem type than another, then a sampler that is not type-aware may allocate more training effort to the majority type simply because it appears more often. We refer to this as \emph{dataset imbalance}.

However, adaptive sampling methods introduce a second and more subtle source of imbalance. At any point in training, only a subset of problems provides substantial learning signal under the current policy. Following the terminology of \citet{foster2025learning}, we refer to this subset informally as the model's \emph{frontier of learnability}, or simply the \emph{frontier}. Frontier imbalance occurs when this productive subset is itself unevenly distributed across problem types. Importantly, frontier imbalance is policy-dependent: as the model improves, the set of problems that are too easy, too hard, or productively learnable can change.

To illustrate the distinction, consider a dataset $D$ consisting of two problem types, $P_A$ and $P_B$. Suppose $P_A$ makes up $80\%$ of the dataset, while $P_B$ makes up the remaining $20\%$. This is a dataset imbalance. Now consider two possible policies. For a first policy, the frontier may be balanced: among the problems that currently provide high learning signal, $P_A$ and $P_B$ may be represented equally. In this case, a productivity-oriented sampler may select useful examples from both types despite the underlying dataset imbalance.

For a second policy, the frontier may be imbalanced in the opposite direction: perhaps $80\%$ of currently productive problems are of type $P_B$, even though $P_B$ is only $20\%$ of the dataset. In this case, a productivity-oriented sampler may concentrate training on $P_B$, not because $P_B$ is more common, but because it is overrepresented on the current frontier. This is not necessarily undesirable: those problems may genuinely provide stronger learning signal at that stage of training. The issue is that productivity alone does not specify how training effort should be distributed across problem types, especially when the goal is broad improvement across a heterogeneous evaluation suite.

The most concerning case is when dataset imbalance and frontier imbalance reinforce each other. If $P_A$ is both the majority type in the dataset and the majority type on the current frontier, then a sampler that prioritizes learning signal may further concentrate training on $P_A$. Conversely, minority types may receive little exposure either because they are rare in the dataset, because they are rarely productive under the current policy, or because they are skipped by the mechanics of the sampling procedure. This can create a feedback loop in which some regions of the task space receive substantial training effort while others remain undertrained.

\begin{figure*}[t]
    \centering
    \includegraphics[width=\textwidth]{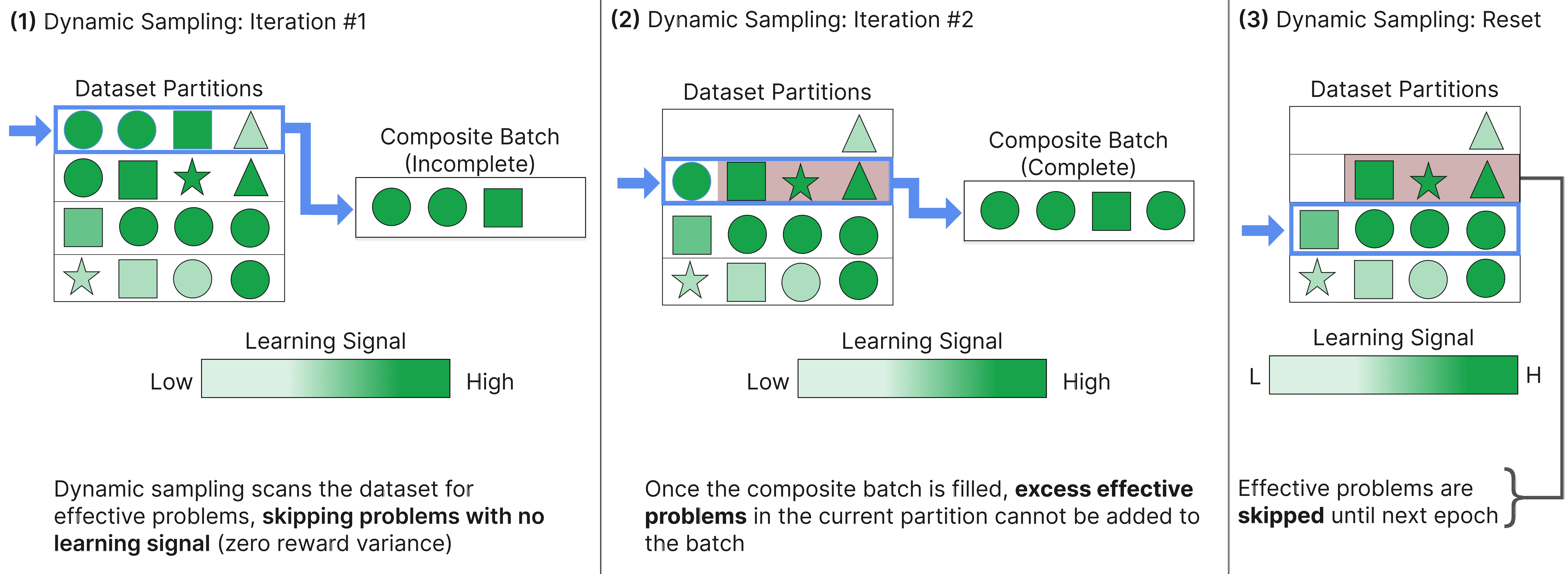}
    \caption{\textbf{Problem skipping in Dynamic Sampling.}
    Dynamic Sampling scans candidate prompts and adds prompts with nonzero reward variance to a composite batch. Once the composite batch is filled, later effective prompts in the current partition may be deferred until a future pass. When effective prompts are unevenly distributed across problem types or dataset partitions, this mechanism can interact with dataset and frontier imbalance. See \hyperref[fig:problem_skipping_eval]{Figure~\ref*{fig:problem_skipping_eval}} for a consequence of this mechanism.}
    \label{fig:problem_skipping}
\end{figure*}

This distinction helps clarify the behavior of generate-then-discard methods such as Dynamic Sampling. Dynamic Sampling constructs effective batches by generating rollouts for candidate prompts and discarding prompts whose rollouts have zero reward variance. This ensures that the final composite batch contains prompts with nonzero learning signal. However, depending on implementation details, prompts that are effective but appear after the composite batch has already been filled may not be used until a later pass through the data. \hyperref[fig:problem_skipping]{Figure~\ref*{fig:problem_skipping}} illustrates this mechanism. When productive prompts are unevenly distributed across dataset partitions or problem types, this kind of skipping can interact with dataset and frontier imbalance.

\begin{figure*}[!h]
    \centering
    \includegraphics[width=\textwidth]{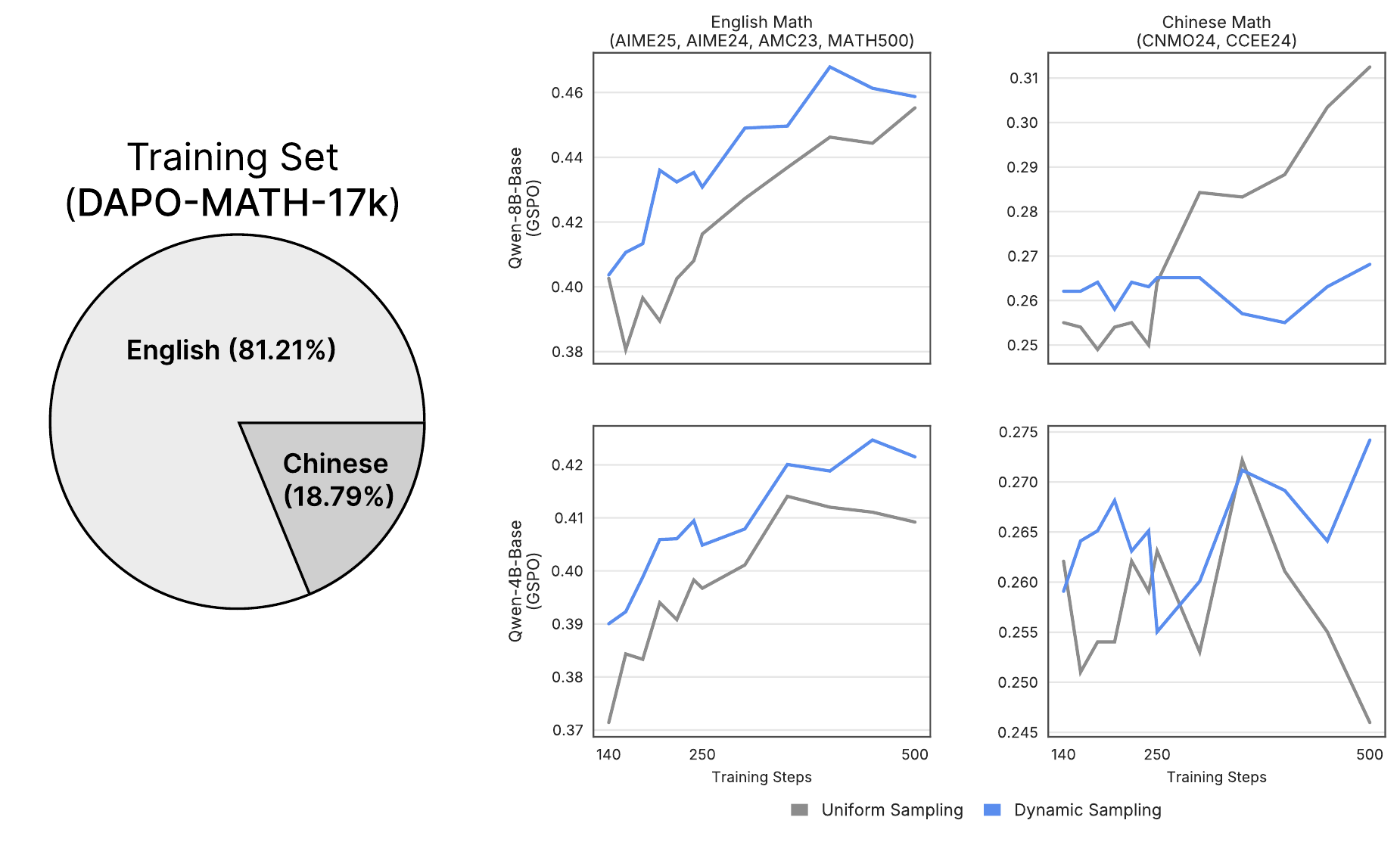}
    \caption{\textbf{Dynamic Sampling and minority-type exposure.}
    The training data is dominated by English mathematics problems, while Chinese mathematics problems form a smaller subset. For the 8B model, Dynamic Sampling improves strongly on English mathematics evaluations but plateaus on Chinese mathematics evaluations relative to uniform sampling. This behavior is consistent with an interaction between dataset imbalance, policy-dependent frontier imbalance, and problem skipping, though the experiment does not isolate these factors causally.}
    \label{fig:problem_skipping_eval}
\end{figure*}

In our experiments, this issue is most visible in the English--Chinese mathematics split. As shown in \hyperref[fig:problem_skipping_eval]{Figure~\ref*{fig:problem_skipping_eval}}, the dataset is dominated by English mathematics problems, while Chinese mathematics problems form a smaller fraction of the training set. For the 8B model, Dynamic Sampling improves strongly on English mathematics evaluations but plateaus or underperforms on the Chinese mathematics evaluations relative to uniform sampling, despite constructing batches with high effective ratio. This pattern is consistent with the hypothesis that productivity-oriented filtering and problem skipping can underexpose minority regions of the dataset. However, the experiment does not isolate whether the effect is due to dataset imbalance, frontier imbalance, the skipping mechanism, or their interaction.

The behavior also differs across model sizes. For the 4B model, the same plateau is not observed. One possible explanation is that the smaller model's frontier contains a different mixture of productive prompts, giving Chinese mathematics problems more exposure under Dynamic Sampling. We treat this interpretation as suggestive rather than causal: the key point is that the frontier of learnability depends on the policy, and therefore sampling imbalances can change with model size and training stage.

Notably, this discussion is not meant to argue that Dynamic Sampling is inherently flawed. Alternative implementations could reduce or eliminate the skipping issue, for example by returning unused effective prompts to the active pool rather than deferring them to a later epoch. More broadly, the point is that adaptive sampling methods should account for the heterogeneous, multi-task nature of LLM training. A sampler can be highly effective at maximizing immediate learning signal while still allocating training effort unevenly across problem types. This motivates type-aware or structure-aware sampling methods such as BMC, which explicitly organize sampling over the latent task geometry rather than relying on productivity (difficulty) alone.

Finally, frontier imbalance is not necessarily a failure mode. In some stages of training, disproportionately sampling a subset of problem regions may be desirable if those regions genuinely provide stronger learning signal. BMC is therefore not designed to eliminate imbalance, but to make such imbalance structured and \emph{explicitly controllable}. The concern is not that adaptive samplers focus training effort unevenly, but that this unevenness can arise implicitly, without direct awareness of whether it reflects dataset composition, policy-dependent frontier structure, sampling mechanics, or their interaction.

\clearpage
\section{Utility-Aware Sampling and the Evaluation-Protocol Tension}
\label{app:target_distributions}

BMC-T uses a target distribution to estimate which parts of a heterogeneous training set are most relevant to a desired capability or evaluation. In our experiments, we instantiate this target distribution using held-out evaluation benchmarks in order to test whether different training problem types carry different evaluation relevance. Target examples are never used as training examples and cannot be sampled by the scheduler; they only influence the utility scores assigned to sampleable regions of the training distribution.

The broader motivation is that, in large RL training datasets, it is often unclear which problems are most important for inducing a desired capability. A dataset may contain many superficially relevant examples, but only some problem types may actually support transfer to the target behavior. For example, if the goal is to improve performance on a specific, extremely difficult class of competition mathematics, such as International Mathematical Olympiad (IMO)-style reasoning, one could construct a large heterogeneous math training set without training directly on IMO problems. A utility-aware scheduler could then use a representative target set to bias sampling toward training problems that the policy model views as nearby or relevant, allowing training effort to concentrate on problems that better support the desired capability.

This results-oriented use case is distinct from the role of target distributions in algorithmic evaluation. In algorithmic research, held-out benchmarks are typically intended to measure generalization independently of the training process, but each benchmark also emphasizes a particular distribution of capabilities. At the same time, benchmark performance is often treated as the primary evidence for a sampling method's effectiveness, even though heterogeneous training data may not contribute equally to those benchmarks. This creates an apparent paradox: the scheduler is judged by downstream evaluation performance, but evaluation relevance is not part of its objective unless target information is explicitly provided. Ignoring evaluation relevance can therefore obscure an important axis of curriculum design, while using an evaluation benchmark to guide sampling changes the role of that benchmark.

This motivates a protocol distinction. If a benchmark is used to define the target distribution for BMC-T, then it should be understood as part of the curriculum design process rather than as an untouched final test set. This is analogous to the role of a \textbf{validation set}: it may guide model or training decisions, but should not also serve as the final measure of generalization. Under a strict held-out protocol, one can instead designate a separate validation or development set as the target distribution, use it to construct the utility signal, and reserve separate benchmarks for final evaluation.

Constructing such a validation or target set is itself nontrivial. What appears relevant to humans may not be what is most relevant to the policy model. Human-defined data taxonomies, topic labels, or perceived similarity between problems may fail to capture how the model organizes the task space internally. As a result, a validation set that appears representative under human semantics may provide a weaker utility signal than expected if it is poorly aligned with the regions of the model's latent task manifold that contain useful training problems. Latent Task Trees can help diagnose this mismatch by showing how candidate target examples are distributed relative to the training data under the policy's own representation space. In this way, the same representation used for utility-aware sampling can also help design or audit validation sets intended to represent a desired capability.

We do not aim to resolve broader questions about benchmark awareness or train--test separation. These protocol choices depend on the goals of a given study: whether the objective is to build a target-aware training system, compare sampling algorithms under strict held-out evaluation, or diagnose which training problems support a particular benchmark. Our use of BMC-T is primarily diagnostic and methodological: it tests whether productivity and diversity fully explain downstream evaluation performance, or whether some problem types have greater influence on specific evaluation benchmarks. The results in \hyperref[sec:bmct]{Section~\ref{sec:bmct}} support the latter interpretation, showing that changing the target distribution can shift evaluation outcomes even when apparent training productivity remains similar.

\clearpage
\section{External Models}
\label{app:external-models}

BMC constructs Latent Task Trees from the policy model's own hidden representations. This choice is somewhat different from many standard approaches to dataset organization, which often rely on semantic embedding models, frontier-model annotations, or manually defined task labels. We use the policy's internal representation space because our goal is not simply to recover human-interpretable task categories, but to approximate the structure of the problem space as perceived by the model being trained. We discuss this design choice below.

\subsection{Semantic Embedding Models}
A natural alternative is to cluster prompts using a pretrained embedding model. Such models are typically optimized to capture \emph{human-facing} semantic similarity, for example by placing paraphrases, related documents, or retrieval-relevant texts close together in embedding space. This is useful for retrieval, ranking, and semantic search, but it is not necessarily the same notion of similarity needed for curriculum learning.

In BMC, the relevant geometry is the geometry of the training model's own behavior. For example, two prompts that are paraphrases, translations, or surface variants of the same problem may be semantically similar to a human or to a retrieval model, while still inducing different hidden representations, rollout behavior, reward variance, or learning dynamics in the policy model. Conversely, prompts that look different to humans may be treated similarly by the model if they activate similar internal reasoning patterns. Since the scheduler must decide which prompts are useful for training this particular policy, we use the policy model's hidden states rather than an external embedding model as the basis for task structure. 

One possible concern is that generic embedding models may produce cleaner semantic clusters than raw hidden states from a non-embedding LLM. However, our Latent Task Tree visualizations suggest that the policy model's internal representations are sufficiently structured for curriculum construction: the induced clusters are often coherent and fine-grained, even though human-readable labels are added only post hoc for analysis.

\subsection{Frontier Model Annotations \& Prompt-based Clustering.}
Another reasonable option is to ask a frontier model to label, summarize, or categorize each prompt, and then cluster prompts using those generated descriptions. This can produce highly interpretable taxonomies, and we use such annotations for analysis and visualization. However, we do not use them to construct the Latent Task Tree itself.

The main limitation is that a frontier model supplies an external semantic perspective. Its labels may reflect \emph{human-aligned} categories or its own internal organization, rather than the representation geometry of the policy model being trained. For curriculum scheduling, this distinction matters: if the training model represents two problem types similarly, confuses them in similar ways, or generalizes between them, that relationship should be reflected in the scheduler even if an external annotator would label them differently.

There are also practical considerations. Prompt-based labeling can become expensive or cumbersome at large scale, especially for multimodal datasets where each example may require image, audio, or structured-context understanding. In contrast, if the policy model's hidden states are accessible, the same representation-extraction pipeline can be applied across modalities supported by the model.

A related alternative is to prompt the same model being trained to describe or cluster the dataset. This is closer in spirit to our goal, but it still replaces internal representation geometry with verbal self-description. Model-generated explanations or labels may be useful, but they need not faithfully expose the model's internal organization of prompts \citep{parcalabescu2024measuringfaithfulnessselfconsistencynatural}. We therefore treat prompt-based labeling as an analysis tool rather than as the source of task structure.

\subsection{Sparse Autoencoders.}
\label{app:em_saes}
Sparse autoencoders (SAEs) provide another possible route for analyzing model representations. SAEs learn sparse dictionaries over activations, often decomposing hidden states into more interpretable features \citep{cunningham2023sparseautoencodershighlyinterpretable, kim2026disentangledsparserepresentationsconceptseparated}. This makes them valuable for mechanistic interpretability and for identifying \emph{human-interpretable} components of model computation.

However, human interpretability is not the primary goal of BMC. The Latent Task Tree is used as a curriculum structure, not as a mechanistic explanation of individual features. While SAE features could potentially provide a useful basis for task discovery, training and validating SAEs introduce substantial additional complexity, including choices of layer, dictionary size, sparsity penalty, feature aggregation, and feature-to-prompt representation. For this work, we instead use normalized hidden-state representations directly, which keeps the tree construction pipeline simpler while remaining tied to the policy model's own latent geometry.

\subsection{Critic-Style Prompt Predictors}
Another alternative is to train an additional model to estimate prompt difficulty, learning potential, or policy-improvement utility for curriculum selection~\citep{gao2025promptcurriculumlearningefficient, mahrooghi2026goldilocksrltuningtask, gu2026actorcuratorcoadaptivecurriculumlearning}. BMC takes a different approach to the same underlying problem of estimating which prompts should receive training effort. Rather than training a separate predictor to directly regress a policy-dependent target, BMC uses the policy model's own hidden representations to define a latent geometry over prompts, then maintains online Bayesian beliefs over learning signal within that geometry.

This shifts the modeling burden from learning a high-dimensional productivity function to maintaining local, uncertainty-aware estimates and sharing evidence between geometrically related prompts. Since prompt productivity changes as the policy evolves, we view this as a useful inductive bias: the tree supplies structure, while surprise-based forgetting and uncertainty growth allow beliefs to adapt to non-stationarity. This does not rule out learned productivity predictors, which may be valuable when sufficiently accurate and responsive. Rather, BMC provides a complementary alternative that avoids adding a separate critic or teacher model as the central mechanism. A secondary benefit is that once curriculum selection is organized over latent geometry, the same structure can also support explicit control over type coverage (diversity) and utility, as in BMC-T.

\paragraph{Summary.}
Overall, external models can be useful for labeling, validation, and analysis, but BMC deliberately avoids making them the source of curriculum structure. The Latent Task Tree is derived from the policy model's own hidden representations because, for curriculum scheduling, the relevant semantics are the semantics of \emph{the model being trained} rather than an external model's taxonomy. This also preserves one of the practical advantages of group-relative RLVR: curriculum selection can be added without introducing a separate critic, teacher, annotator, or embedding model as a required component. BMC therefore remains a lightweight extension of the training loop while still providing structured control over productivity, diversity, and utility.

\clearpage
\section{Inter-Batch Diversity (Temporal Awareness)}
\label{app:inter-batch-diversity}

BMC primarily addresses diversity as coverage over latent problem types within a sampled batch (\emph{intra}-batch diversity). A separate notion is whether the scheduler exposes the model to different combinations of problem types across nearby batches over time (\emph{inter}-batch diversity). In other words, even if each individual batch is diverse, consecutive batches may still repeatedly draw from the same regions of the task space.

From this perspective, uniform sampling and dynamic sampling can provide strong inter-batch diversity by repeatedly sweeping through the dataset. However, this form of diversity is not necessarily type-aware: it may expose the model to many different prompts over time while still reflecting dataset imbalance, frontier imbalance, or ordering effects from the sampling procedure. BMC instead makes type coverage explicit within each batch, but does not currently include a separate mechanism for discouraging repeated selection of the same subtrees across consecutive batches.

We did not explicitly penalize repeated visitation in this work. This was partly motivated by evidence suggesting that RLVR can generalize even from extremely small training sets, including settings where a single example is sampled repeatedly and still induces broad downstream improvements~\citep{wang2025reinforcementlearningreasoninglarge}. Thus, repeated exposure to a small region of the prompt space is not necessarily harmful in RLVR. Nevertheless, inter-batch diversity may still matter in heterogeneous datasets. Even if repeated sampling can generalize, a productivity-oriented sampler may over-concentrate on a small set of currently informative regions, delaying exposure to other useful regions of the latent task manifold.

A natural extension is therefore to add a temporally aware visitation penalty at the leaf level. Let $\nu_i^{(t)}$ denote the visitation memory for prompt $i$ at scheduler step $t$. For a sampled batch $\mathcal{B}_t$, define
\[
\nu_i^{(t)}
=
\alpha \nu_i^{(t-1)}
+
\mathbf{1}[i \in \mathcal{B}_t],
\]
where $\alpha \in [0,1)$ controls how quickly past visitation decays. If $\tilde S_i^{(t)}$ denotes the original BMC leaf score for prompt $i$, a visitation-aware score could be written as
\[
S_i^{(t)}
=
\tilde S_i^{(t)}
-
g(\nu_i^{(t)}),
\]
where $g$ is a nonnegative increasing penalty function, such as $g(\nu)=\log(1+\nu)$ or $g(\nu)=\sqrt{\nu}$. The overall strength and saturation behavior of the visitation penalty can be absorbed into the choice of $g$.

The adjusted leaf scores could then propagate upward through the same empirical-Bayes aggregation used by BMC. If many leaves within a subtree have recently been sampled, their visitation-penalized scores lower the aggregated score of the corresponding internal node. Conversely, if only a small number of leaves have been visited, the penalty remains localized. This makes visitation a dynamic analogue of the static utility bias used in BMC-T: utility can bias the tree toward target-relevant regions, while visitation can temporarily repel the scheduler away from recently sampled regions. Such a mechanism would encourage inter-batch diversity while preserving BMC's intra-batch diversity mechanism.

Overall, this extension would make BMC's diversity control more temporally aware and could help encourage exploration of the latent task manifold. However, to our knowledge, the empirical importance of inter-batch diversity remains less established than difficulty- or task-type-related curriculum mechanisms. Recent related work such as GPS~\citep{qu2026smallgeneralizablepromptpredictive} includes inter-batch diversity mechanisms and reports ablations suggesting that they can be beneficial. We therefore view visitation-aware sampling as a promising but relatively unvalidated extension rather than a necessary component of BMC.

\clearpage
\section{Limitations}
\label{app:limitations}

\subsection{Tree Construction Heuristics}

The recursive construction of the Latent Task Tree is heuristic. Although we find that the current pipeline produces coherent, fine-grained trees across different models, datasets, languages, domains, and modalities, there may be better choices for the latent representation, layer selection, dimensionality reduction, manifold approximation, stopping criterion, or clustering algorithm.

These choices also introduce hyperparameters whose effects are difficult to evaluate in isolation. Constructing trees is relatively inexpensive compared to RL training, but determining whether one tree is better than another for \emph{curriculum scheduling} ultimately requires running downstream RL experiments. This makes comprehensive tree-construction ablations substantially more expensive than tree generation itself. We therefore view the current construction pipeline as a practical instantiation rather than an optimized solution, and leave systematic exploration of alternative tree-building procedures to future work.

\subsection{Computational Requirements}

Latent Task Trees are broadly applicable across model sizes, domains, languages, and modalities, but they require extracting and processing hidden representations for the training prompts. If the combination of model size and dataset size is sufficiently large, the default tree-construction pipeline may become computationally or memory intensive.

Our current implementation primarily uses CPU-based preprocessing for tree construction. This is not a fundamental requirement: GPU-accelerated variants of PCA, UMAP, and HDBSCAN could reduce construction time substantially (e.g., \citep{nolet2021bringingumapcloserspeed}). However, such a GPU-based implementation is not included at the time of writing. Since LLM training is already heavily GPU-dependent, an optimized GPU pipeline would be a useful engineering improvement. For very large datasets or models, additional memory-efficient techniques, such as streaming PCA, approximate nearest-neighbor construction, batching, or subsampling-based initialization, may also be necessary.

\subsection{Structure Drift}

The Latent Task Tree is constructed once before training and then held fixed throughout RL. We analyze this issue in \hyperref[app:structure_drift]{Appendix~\ref*{app:structure_drift}}. Our results suggest that a static tree can remain useful during training, but this remains a limitation: as the policy changes, the model's hidden representations may drift, and the latent geometry used to construct the original tree may become less aligned with the current policy.

One natural mitigation is to periodically reconstruct the tree from updated model representations. We explored this direction empirically, but online reconstruction introduces practical overhead. In our implementation, rebuilding the tree required interrupting and resuming training, since latent extraction and RL training could not be performed conveniently with all required models loaded simultaneously. This is primarily an engineering limitation rather than a conceptual one: prompt-level beliefs can be reassigned to the corresponding leaves after reconstruction using prompt identities, and the scheduler's action space is simply defined by the newly constructed tree.

The remaining question is whether the benefit of periodic reconstruction justifies its cost. In our experiments, periodic reconstruction did not substantially change the main conclusions, suggesting that the initial latent geometry is sufficiently stable for the training horizons considered here. However, representation drift may become more important for longer training runs, larger policy updates, or settings where the model's capabilities change more dramatically. We therefore leave more efficient online tree reconstruction and adaptive update schedules to future work.

\subsection{Belief Bounds and Uncertainty Saturation}
\label{app:limitations_4}

BMC was designed to avoid introducing many additional tuning coefficients. Aside from tree-construction choices, the main scalar hyperparameters in the belief updates are the bounds on $\mu_i$, $\lambda_i$, and $\sigma_i^2$, together with the sigma-rule constant used to determine the maximum uncertainty bound (\hyperref[app:logit-normal-sigma-rule]{Appendix~\ref{app:logit-normal-sigma-rule}}). Other quantities, such as the coverage coefficient used for staleness accumulation, are determined by the data and batch size rather than manually tuned.

Nevertheless, the resulting uncertainty dynamics may still require further study. In particular, we observed that many prompt-level beliefs reached the maximum uncertainty bound by the end of training, with  $80\%$ of prompts reaching maximum uncertainty by the end of the main training runs. This may indicate that the current uncertainty-growth rule re-inflates stale prompts more aggressively than necessary, or that the maximum variance bound determined by the sigma rule is not well matched to the observed learning-signal dynamics. If too many prompts saturate at maximum uncertainty, the posterior may become less informative for distinguishing genuinely uncertain prompts from merely stale ones.

This suggests several possible refinements. For example, staleness could accumulate sublinearly rather than through the current coverage-adjusted counter, or the uncertainty injection term could use a different saturating transform. Similarly, the sigma-rule bound could be adjusted to better match the empirical range of transformed learning signals. We leave these variants to future work.

Despite these limitations, the current bounded update rule was sufficient to improve learning signal and effective ratio in the main experiments. The behavior was less consistent in the medical-domain experiments in \hyperref[app:medical]{Appendix~\ref*{app:medical}}, where learning signal decreased over time for both BMC and the Difficulty Only ablation. This may reflect the relative difficulty or reward structure of the dataset, overly aggressive uncertainty growth, or a more fundamental mismatch between the learning-signal statistic and that domain.

\clearpage
\section{Directions for Future Work}
\label{app:future_work}

\subsection{Rollout Allocation}
In this work, we deliberately keep the rollout allocation per prompt fixed. This isolates the effect of changing \emph{which} problems are sampled, rather than simultaneously varying how much compute is assigned to each sampled problem. However, several recent works study adaptive rollout allocation, where harder or more promising problems receive more rollouts in order to improve exploration, solution discovery, or compute efficiency \citep{zeng2025cures,li2025knapsackrlunlockingexploration,yao2026cobarlcapabilityorientedbudgetallocation,cheng2026isocomputeplaybookoptimallyscaling}.

A natural direction is to combine adaptive rollout allocation with the axes studied in this work. If rollout budgets are allocated only by difficulty or immediate productivity, they may encounter the same coverage and target-mismatch issues discussed throughout the paper. Future methods could therefore make rollout allocation \emph{type-aware}, allocating compute not only according to prompt difficulty, but also according to diversity and utility.

\subsection{Prioritization Mechanisms}

BMC prioritizes sampleable problems using uncertainty-aware estimates of learning signal, with evidence shared through the Latent Task Tree. This is only one possible acquisition rule. As inter-task learning dynamics become better understood, more informed prioritization mechanisms may become possible.

For example, BMC currently favors problems whose local tree neighborhoods have high aggregated productivity. However, a productive problem may be especially valuable if it lies nearby other hard problems in the policy's latent space. If learnability propagates locally through the latent task manifold, then such problems may act as \emph{frontier expanders}: training on them could help unlock nearby problems that are not yet productive. This would make them more valuable than equally productive problems surrounded mostly by already-learnable or easy problems.

This suggests a broader direction for future work: acquisition functions that reason not only about the current learning signal of a problem, but also about how training on that problem may change the learnability of nearby problems. Such mechanisms could extend BMC from sampling currently productive problems to actively expanding the frontier of learnability.

\subsection{Connections to the Bandit Literature}

BMC was inspired by bandit settings in which the arms are not independent, but related through structure, hierarchy, interventions, constraints, or shared context. We believe this broader literature offers many tools that remain underexplored in LLM training. Standard prompt-level curricula already resemble bandit algorithms in spirit, but most current methods use only a small subset of the available ideas: scalar prompt values, simple uncertainty estimates, and difficulty-oriented sampling rules.

Future work could draw more directly from structured bandits, causal bandits, combinatorial bandits, pure-exploration bandits, and budgeted bandits. \textbf{Structured bandits} provide tools for sharing information across related arms. \textbf{Causal bandits} \citep{lattimore2016causalbanditslearninggood} and intervention-based formulations may be useful for asking not only which prompts currently provide learning signal, but which training interventions change the learnability of other regions. \textbf{Combinatorial bandits} \citep{pmlr-v28-chen13a} are relevant because each training batch is a constrained subset of prompts rather than a single arm. \textbf{Pure-exploration bandits} \citep{bubeck2010pureexplorationmultiarmedbandit} suggest mechanisms for identifying frontier regions, high-utility regions, or underexplored task types under limited sampling budgets. \textbf{Budgeted bandits} \citep{tranthanh2012knapsackbasedoptimalpolicies} connect naturally to rollout allocation and compute-aware training.

More broadly, this perspective suggests that curriculum learning for LLMs need not rely only on manually designed heuristics or fixed notions of difficulty. Bandit-style formulations offer a way to make sampling decisions adaptive, uncertainty-aware, and compute-conscious while remaining lightweight enough to fit into existing training pipelines.

\clearpage
\section{Additional Experimental Details}
\label{app:experimental_details}

\subsection{Hyperparameters}
\label{app:hyperparameters}

\begin{table}[h]
\centering
\caption{\textbf{Hyperparameters.} We detail hyperparameters central to our experimental setup below.}
\label{tab:appendix-hparams}

\footnotesize
\setlength{\tabcolsep}{4pt}
\renewcommand{\arraystretch}{0.88}

\begin{tabular}{@{}p{0.62\textwidth}p{0.28\textwidth}@{}}
\toprule
\textbf{Hyperparameter} & \textbf{Value} \\
\midrule

\multicolumn{2}{@{}l}{\textbf{\underline{Model Generation}}} \\
Train Temperature & $1.0$ \\
Train Top-P & $1.0$ \\
Evaluation Temperature & $0.6$ \\
Evaluation Top-P & $0.95$ \\
Rollouts-per-prompt ($k$) & $8$ \\
Maximum Input Length & $3000$ \\
Maximum Response Length & $4096$ \\
Inference Engine & vLLM \\

\addlinespace[0.35em]
\multicolumn{2}{@{}l}{\textbf{\underline{Optimization}}} \\
Optimizer & AdamW \\
Learning Rate & $0.000001$ \\
Learning Rate Warmup Steps & $0$ \\
Grad Clip & $1.0$ \\
Weight Decay & $0.0$ \\
Batch Size & $128$ \\
Mini-Batch Size & $32$ \\
Training Engine & FSDP2 \\

\addlinespace[0.35em]
\multicolumn{2}{@{}l}{\textbf{\underline{GSPO}}} \\
Clip Ratio (High) & $0.0004$ \\
Clip Ratio (Low) & $0.0003$ \\
Use KL & No \\
Entropy Coef & $0.0$ \\

\addlinespace[0.35em]
\multicolumn{2}{@{}l}{\textbf{\underline{GRPO}}} \\
Clip Ratio (High) & $0.2$ \\
Clip Ratio (Low) & $0.2$ \\
Use KL & Yes \\
KL Coefficient & $0.001$ \\
Entropy Coef & $0.0$ \\

\addlinespace[0.35em]
\multicolumn{2}{@{}l}{\textbf{\underline{Tree Construction}}} \\
Seed & $0$ \\
Layer Depth & $0.75$ \\
Variance Explained Ratio (PCA) & $0.95$ \\
KNN Min Points for Connectivity (Chart Test) & $500$ \\
KNN Connectivity Threshold (Chart Test) & $10$ \\
TwoNN Dimension Threshold (Chart Test) & $1.5$ \\
Min. Distance (UMAP) & $0.0$ \\
Max Projection Dimension (UMAP) & $50$ \\
N Neighbors (UMAP) & $30$ \\
Min. Cluster Percentage (HDBSCAN) & $0.01$ \\
Min. Global Cluster Size (HDBSCAN) & $5$ \\

\addlinespace[0.35em]
\multicolumn{2}{@{}l}{\textbf{\underline{BMC}}} \\
Initial $\mu$ & $0.0$ \\
Initial $\lambda$ & $1.0$ \\
Minimum $\lambda$ & $0.2$ \\
Maximum $\lambda$ & $10$ \\
Minimum $\sigma^2$ & $0.05$ \\
$\sigma^2$ Rule (determines maximum $\sigma^2$) & $3$ \\
$\gamma$ (BMC-T only) & $0.25$ \\
$\rho$ (BMC-T only) & $0.05$ \\

\bottomrule
\end{tabular}
\end{table}

\subsection{Compute}
Across all experiments, training was performed on a single H100 GPU node. Due to machine availability, the node interconnect varied across runs, with both Ethernet- and InfiniBand-connected machines used. Latent task tree construction was performed on CPU, using machines with approximately 32 CPU cores. All methods were run under the same hardware constraints, and comparisons are made within matched experimental settings.

\subsection{Round-Robin Initialization}
\label{app:round_robin_initialization}

At the start of training, before adaptive sampling is activated, we perform a round-robin initialization pass for BMC and its ablations. During this phase, each training prompt is sampled once using the same rollout allocation used throughout training, yielding an initial estimate of its learning signal. This avoids a cold-start regime in which the scheduler must compare observed prompts against prompts that have not yet been sampled, and provides initial empirical statistics for information sharing across the Latent Task Tree. We use this initialization for experimental stability and compute efficiency; it is not a requirement of the method, and alternative cold-start strategies could be used in settings where a full round-robin pass is impractical.

BMC and Difficulty Only are initialized from the same post-round-robin policy checkpoint and the same initial prompt-level belief statistics. For consistency in model exposure, Tree Only also starts from the same post-round-robin policy checkpoint. However, because Tree Only is designed to isolate the effect of the Latent Task Tree without adaptive belief learning, it uses fixed uniform beliefs rather than the round-robin learning-signal estimates. Thus, BMC, Difficulty Only, and Tree Only share the same policy state when adaptive sampling begins, while differing only in how sampling decisions are made thereafter.

\subsection{Evaluation Datasets}
\label{app:datasets}

We evaluate on a mixture of competition mathematics and scientific reasoning benchmarks. The mathematics benchmarks test exact-answer problem solving across varying levels of difficulty, while GPQA evaluates out-of-domain scientific reasoning beyond the mathematical training distribution.

\paragraph{AIME.}
AIME consists of problems from the American Invitational Mathematics Examination, a challenging high-school mathematics competition. We evaluate on AIME 2024 and AIME 2025 using public dataset releases~\citep{aime2024_hf,aime2025_hf}. These benchmarks test olympiad-style mathematical reasoning across topics such as algebra, geometry, number theory, and combinatorics, with answers given as integers between 000 and 999.

\paragraph{AMC 2023.}
AMC 2023 consists of problems from the American Mathematics Competition, drawn from the AI-MO validation set~\citep{aimo_amc_hf}. Compared with AIME, AMC problems are generally shorter and more accessible, but still require nontrivial mathematical reasoning across algebra, geometry, counting, probability, and number theory.

\paragraph{MATH500.}
MATH500 is a 500-problem subset of the MATH benchmark~\citep{hendrycks2021measuringmathematicalproblemsolving}, commonly used to evaluate mathematical reasoning in large language models. The dataset covers a broad range of competition-style mathematics topics, including algebra, geometry, number theory, counting and probability, intermediate algebra, prealgebra, and precalculus.

\paragraph{CNMO 2024.}
CNMO 2024 consists of problems from the China Mathematical Olympiad, drawn from LiveMathBench~\citep{livemathbench_hf}. The benchmark evaluates olympiad-style mathematical reasoning in Chinese, emphasizing non-routine problem solving, multi-step derivations, and advanced mathematical concepts.

\paragraph{CCEE 2024.}
CCEE 2024 consists of mathematics problems from China's College Entrance Examination, also drawn from LiveMathBench~\citep{livemathbench_hf}. Compared with olympiad-style benchmarks such as AIME and CNMO, CCEE reflects a standardized national examination setting and tests mathematical reasoning across the Chinese high-school curriculum.

\paragraph{GPQA.}
GPQA~\citep{rein2023gpqagraduatelevelgoogleproofqa} is a graduate-level multiple-choice question-answering benchmark designed to evaluate scientific reasoning in domains such as biology, physics, and chemistry. Questions are written by domain experts and are intended to be difficult for non-experts even with access to search. We report results on GPQA-Diamond, a more selective subset containing especially high-quality and challenging questions.

\subsection{Dense-to-Sparse Evaluation Schedule}
\label{app:dense-to-sparse-eval}

Evaluation can be a substantial source of cost during RL training, especially when tracking both training-set absorption and downstream benchmark performance across long runs. In this work, we evaluate two complementary quantities: training-set pass@1, which measures how quickly the policy absorbs the training distribution, and downstream mean@16 evaluation performance, which measures generalization to held-out benchmarks. Running both evaluations densely throughout training would substantially increase the financial cost of the experiments.

We therefore use a simple dense-to-sparse evaluation schedule. During the early phase of training, when policy performance often changes rapidly, we evaluate every $10$ optimizer steps. To reduce cost, these early evaluations are interleaved: one checkpoint evaluates training-set pass@1, while the next evaluates downstream mean@16 performance. After step $250$, we switch to a sparser schedule, evaluating every $50$ optimizer steps. At these later checkpoints, we evaluate both training-set pass@1 and downstream mean@16 performance together.

This schedule can be viewed as a simple form of \emph{annealed evaluation}: evaluation frequency is high when the model is changing quickly and lower once training dynamics have slowed. This allows us to measure early learning dynamics while still tracking longer-term behavior at substantially lower cost than fully dense evaluation. The trade-off is that the later part of each curve is sparser, so it may miss short-lived fluctuations or brief regressions. We therefore interpret late-stage evaluation results primarily as coarse trends rather than as fine-grained measurements of every training transition.

\subsection{Evaluation Curves}
\label{app:evaluation_curves}

\begin{figure*}[h]
  \centering
  \includegraphics[width=\textwidth]{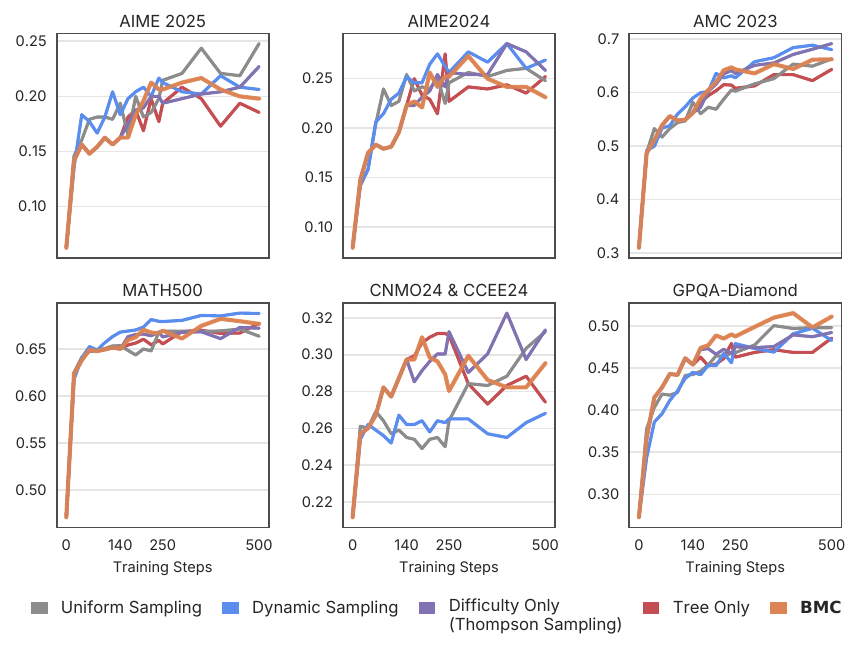}
    \caption{\textbf{Evaluation curves.} Performance across evaluations for \texttt{Qwen3-8B-base} while training on \texttt{DAPO-Math-17k} using \textbf{GSPO}. This corresponds to the bar graphs visualized in \hyperref[fig:eval_main]{Figure~\ref*{fig:eval_main}}. We refer to \hyperref[app:frontier_imbalance]{Appendix \ref*{app:frontier_imbalance}} for additional intuition and discussion on why the performances between the 8B and 4B models can vary on the same dataset.} 
\label{fig:eval_curve_8b_gspo}
\end{figure*}

\begin{figure*}[h]
  \centering
  \includegraphics[width=\textwidth]{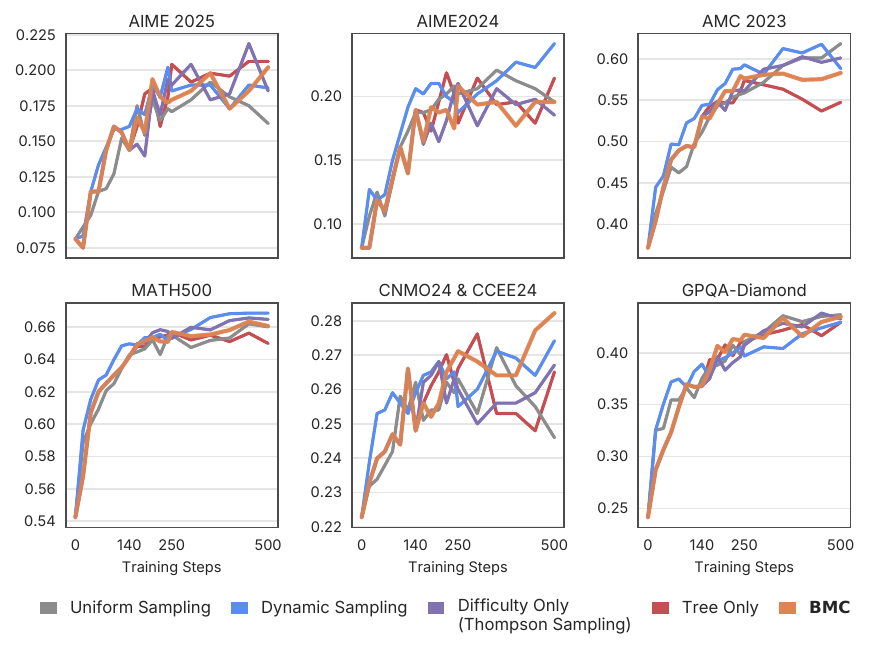}
  \caption{\textbf{Evaluation curves.} Performance across evaluations for \texttt{Qwen3-4B-base} while training on \texttt{DAPO-Math-17k} using \textbf{GSPO}. We refer to \hyperref[app:frontier_imbalance]{Appendix \ref*{app:frontier_imbalance}} for additional intuition and discussion on why the performances between the 8B and 4B models can vary on the same dataset.} 
\label{fig:eval_curve_4b_gspo}
\end{figure*}

\begin{figure*}[h]
  \centering
  \includegraphics[width=\textwidth]{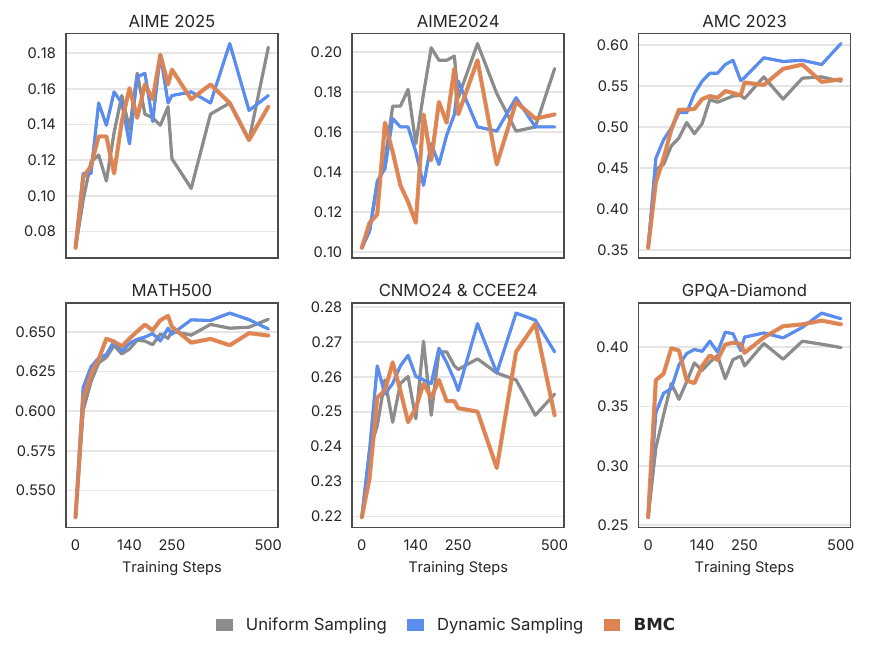}
     \caption{\textbf{Evaluation curves.} Performance across evaluations for \texttt{Qwen3-4B-base} while training on \texttt{DAPO-Math-17k} using \textbf{GRPO}. We refer to \hyperref[app:frontier_imbalance]{Appendix \ref*{app:frontier_imbalance}} for additional intuition and discussion on why the performances between the algorithms can vary on the same model-dataset combination.} 
\label{fig:eval_curve_4b_grpo}
\end{figure*}

\begin{figure*}[h]
  \centering
  \includegraphics[width=\textwidth]{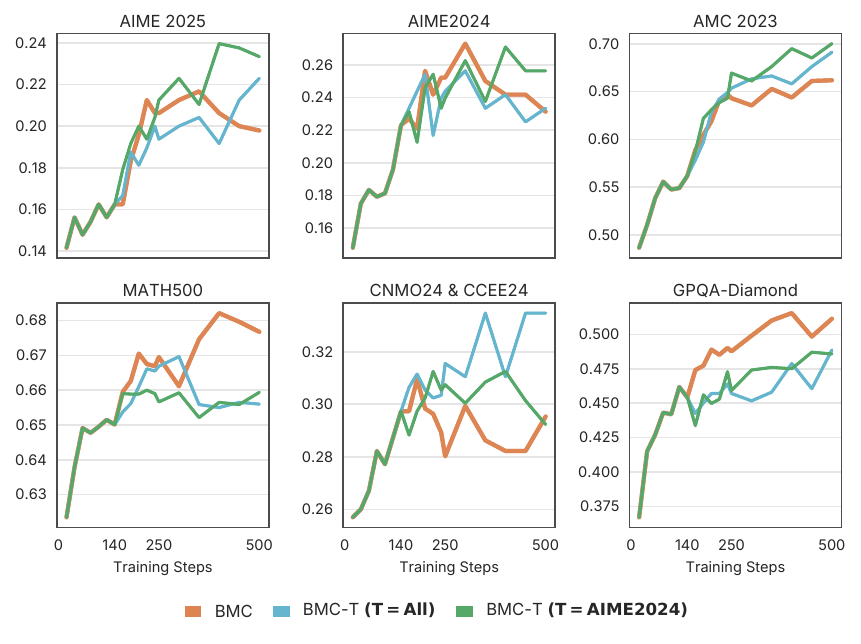}
     \caption{\textbf{Evaluation curves.} Performance across evaluations for \texttt{Qwen3-8B-base} while training on \texttt{DAPO-Math-17k} using variations of BMC-T. These evaluation trajectories correspond to the bar graphs visualized in \hyperref[fig:bmct]{Figure~\ref*{fig:bmct}}.}  
\label{fig:eval_curve_bmct}
\end{figure*}

\clearpage
\section{Additional Algorithmic Details}
\label{app:algorithmic_details}

\subsection{Learning Signal}
\label{app:learning_signal}

In our framework, the reward maximized by the multi-arm bandit scheduler is referred to as the \emph{learning signal}. It is not the same reward that is maximized by the policy. Here we provide more details regarding this signal not given in the main body of work.

\subsubsection{Quantifying the Learning Signal}
\label{app:learning-signal-quantification}

Applying a bandit scheduler to RL training requires a scalar feedback signal for each sampled prompt. We refer to this quantity as the \emph{learning signal}: a prompt-level statistic that should be inexpensive to compute and correlated with expected policy improvement. In RLVR, a natural choice is the variance of rewards across multiple rollouts from the same prompt. Reward variance appears frequently in adaptive curriculum learning methods for RLVR because it captures whether a prompt induces meaningful within-group contrast: if all sampled rollouts receive the same reward, then the prompt provides little or no relative signal for assigning credit among sampled responses.

We can formalize this intuition under a standard group-relative policy optimization setting, such as GRPO, DAPO, or GSPO. Consider a prompt $x_i$ with $k$ sampled responses $\{y_{i,j}\}_{j=1}^k$ and corresponding rewards $\{r_{i,j}\}_{j=1}^k$. Let
\[
\bar r_i = \frac{1}{k}\sum_{j=1}^k r_{i,j}
\]
denote the within-prompt mean reward. A typical group-relative advantage estimator takes the form
\[
A_{i,j} \propto r_{i,j} - \bar r_i,
\]
or, in normalized variants,
\[
A_{i,j} = \frac{r_{i,j} - \bar r_i}{s_i + \epsilon},
\]
where $s_i$ is the within-prompt reward standard deviation. The policy gradient contribution for a sampled response is weighted by this advantage:
\[
\nabla_\theta J(\theta)
\approx
\mathbb{E}_{x_i, y_{i,j}}
\left[
A_{i,j}\nabla_\theta \log \pi_\theta(y_{i,j}\mid x_i)
\right].
\]
Thus, when all rollouts for a prompt receive the same reward, the unnormalized group-relative advantages are zero:
\[
r_{i,1}=\cdots=r_{i,k}
\quad \Longrightarrow \quad
A_{i,j}=0 \;\; \forall j.
\]
Such prompts may be either too easy, with all rollouts succeeding, or too difficult, with all rollouts failing. In both cases, the prompt provides little relative information for improving the policy.

For binary rewards, this connects directly to the common intermediate-difficulty heuristic. If $r \in \{0,1\}$ and the model solves a prompt with probability $p$, then the reward variance is
\[
\mathrm{Var}(r) = p(1-p),
\]
which is maximized at $p=0.5$. Equivalently, for rewards in $\{-1,1\}$,
\[
\mathrm{Var}(r) = 4p(1-p),
\]
which is also maximized at $p=0.5$. Therefore, prioritizing prompts with high reward variance is equivalent, in the binary-reward case, to prioritizing prompts near a 50\% success rate. These prompts lie between trivial and impossible cases, making them more likely to provide informative credit-assignment signal. This is analogous to curriculum-learning intuitions such as a ``Goldilocks'' level of difficulty or, more historically, the Zone of Proximal Development~\citep{vygotsky1978mind}.

More generally, to also account for \emph{continuous} rewards, we note that reward variance is closely related to the magnitude of the group-relative advantage. In the unnormalized case,
\[
\frac{1}{k}\sum_{j=1}^k A_{i,j}^2
\propto
\frac{1}{k}\sum_{j=1}^k (r_{i,j}-\bar r_i)^2,
\]
so the mean squared advantage is proportional to the empirical reward variance. Likewise, objectives that prioritize the expected absolute advantage are closely related to prioritizing the within-prompt reward standard deviation. This provides a simple explanation for why adaptive curriculum methods based on reward variance~\citep{foster2025learning, wang2025reinforcementlearningreasoninglarge}, intermediate-difficulty heuristics~\citep{qu2025can, shen2025bots, hu2025vadevarianceawaredynamicsampling}, expected absolute advantage~\citep{wang2025dump,chen2025self}, or related notions of information gain~\citep{zhang2025speed, zeng2025cures} can often select similar prompts in practice. While these methods differ in their objectives and implementations, they share a common emphasis on prompts that produce nontrivial rollout-level contrast.

There also exists a more subtle reason for this choice of learning signal. Reward variance is not the full policy gradient: the gradient also depends on the score term $\nabla_\theta \log \pi_\theta(y\mid x)$, and prompts may differ in their gradient geometry. In principle, one could define a more direct learning signal using gradient norms, Fisher information, Hessian approximations, or other curvature-based quantities. However, such quantities are substantially more expensive to compute and may introduce task-dependent scale effects. For example, prompts from different domains may induce systematically different token-level likelihoods, response lengths, or gradient magnitudes, causing a gradient-norm-based scheduler to emphasize some task types for reasons unrelated to their usefulness for learning \citep{wu2025imbalancedgradientsrlposttraining}.

Reward variance provides a cheaper and more normalized alternative. It is computed entirely from rollout rewards, is naturally prompt-local, and places binary and continuous reward signals on a common scale when appropriately normalized. We therefore use reward variance as a practical proxy for the availability of learning signal: it identifies prompts whose sampled responses produce meaningful differences in outcome, and hence prompts for which group-relative policy optimization can assign nontrivial credit.

\subsubsection{Normalizing the Learning Signal}

In traditional multitask reinforcement learning, it is generally considered good practice to normalize rewards across heterogeneous tasks in order to maintain balanced learning dynamics and avoid overemphasizing tasks with naturally larger reward scales~\citep{hafner2024masteringdiversedomainsworld, hansen2024tdmpc2scalablerobustworld}. We believe the same principle applies to RL training for LLMs, particularly when reward magnitudes differ across prompts, tasks, or datasets.

For this reason, while not strictly required, our framework assumes that the learning signal is normalized to the range $[0,1]$ whenever possible. If the underlying rewards are bounded and the bounds are known, this normalization can be implemented with standard rescaling procedures. For example, given a learning signal $s$ with known minimum and maximum values $s_{\min}$ and $s_{\max}$, one may apply:
\[
s_{\mathrm{norm}}
=
\frac{s - s_{\min}}{s_{\max} - s_{\min}}.
\]

More generally, any monotonic transformation that maps heterogeneous reward scales to a shared range may be used. This ensures that the scheduler compares prompts using a consistent notion of learning signal magnitude, rather than inadvertently prioritizing prompts whose rewards are numerically larger for unrelated reasons.

Importantly, normalization is not fundamentally required by our framework. Since the scheduler models learning signals using Gaussian beliefs, it can in principle represent arbitrary continuous quantities, including unbounded signals. However, if learning signals from different prompts or tasks reside on substantially different scales and cannot be normalized appropriately, sampling imbalance issues may arise. In such settings, the scheduler may over-prioritize certain prompts due purely to scale differences rather than genuine informativeness. We leave a more detailed investigation of unbounded and heterogeneously scaled learning signals to future work.

\subsection{Logit-Normal Beliefs and the Sigma Rule}
\label{app:logit-normal-sigma-rule}

The main text presents BMC's prompt-level belief model as Gaussian for clarity. In the implementation, however, the normalized learning signal satisfies $y_i \in [0,1]$. An \emph{unconstrained} Gaussian belief directly over this bounded signal can assign probability mass to impossible values, which makes uncertainty difficult to interpret when beliefs are propagated through the Latent Task Tree. A Beta distribution would naturally respect the interval $[0,1]$, but is most directly suited to Bernoulli-style success probabilities, whereas our learning signal may be a continuous bounded quantity such as normalized reward variance or expected absolute advantage. We therefore use a Logit-Normal parameterization: BMC maintains Gaussian beliefs in logit space, where posterior updates, hierarchical Thompson sampling, and empirical-Bayes aggregation remain convenient. Under the inverse sigmoid map, this Gaussian belief corresponds to a valid bounded belief over the original learning-signal scale.

Before applying the logit transform, we clamp the observed signal to avoid infinite values:
\[
\tilde y_i = \operatorname{clip}(y_i, \epsilon, 1-\epsilon).
\]
We then transform the signal as
\[
r_i = \operatorname{logit}(\tilde y_i)
=
\log \frac{\tilde y_i}{1-\tilde y_i}.
\]
BMC maintains a Gaussian belief over this transformed signal,
\[
r_i \sim \mathcal{N}(\mu_i, \sigma_i^2).
\]
Equivalently, the inverse sigmoid map induces a logit-normal belief over the original bounded learning signal:
\[
y_i = \sigma(r_i)
=
\frac{1}{1+\exp(-r_i)}.
\]
In practice, BMC performs belief updates, aggregation, and sampling directly in logit space. The inverse map is useful for interpretation and intuition: it shows that the underlying belief corresponds to valid learning-signal values in $[0,1]$.

Importantly, because the sigmoid and logit transforms are monotone, operating in logit space does not change the ordering of point estimates used for comparative Thompson sampling. Thus, if one prompt or distribution has a larger estimated learning signal than another in the original space, it also has a larger transformed estimate in logit space. The purpose of the transform is therefore not to change what counts as productive, but to place uncertainty modeling and empirical-Bayes aggregation in a space where Gaussian beliefs are better aligned with the bounded geometry of the learning signal.

The clamping parameter $\epsilon$ is chosen based on the number of rollouts per prompt. In our implementation, with $k$ rollouts, we set
\[
\epsilon = \frac{1}{k^2}.
\]
This avoids treating empirical signals of $0$ or $1$ as infinitely certain endpoints, which would be inappropriate under finite rollout estimates. The corresponding maximum logit magnitude is
\[
r_{\max}
=
\left|\operatorname{logit}(1-\epsilon)\right|.
\]
For example, with $k=8$ rollouts, $\epsilon=1/64$ and $r_{\max}=\log(63)$.

Let $a$ denote the sigma-rule constant, set to $a=3$ in our experiments.\footnote{This corresponds to the standard ``three-sigma'' heuristic for Gaussian distributions, where approximately $99.7\%$ of the mass lies within three standard deviations of the mean; see the \href{https://en.wikipedia.org/wiki/68-95-99.7_rule}{68--95--99.7 rule}.} We choose $\sigma_{\max}$ so that $a$ standard deviations span the effective logit range:
\[
a\sigma_{\max} = r_{\max}.
\]
Thus,
\[
\sigma_{\max}^2
=
\left(\frac{r_{\max}}{a}\right)^2.
\]
This prevents uncertainty from growing without bound while still allowing Thompson samples to cover the effective range of attainable learning signals. Conversely, we also impose a minimum variance to prevent beliefs from collapsing permanently, ensuring that prompts can be revisited if their learning signal changes as the policy evolves.

Overall, the logit-normal parameterization aligns the belief model with the bounded geometry of the learning signal. The sigma rule then provides a simple scale for uncertainty in the transformed space: large enough to represent substantial uncertainty over the full range of possible signals, but bounded enough to avoid unstable empirical-Bayes propagation or Thompson samples dominated by extreme logit-space values.

\subsection{Sampling Without Replacement}
\label{app:sampling_without_replacement}

In LLM training, prompt batches are typically sampled without replacement so that the same prompt does not appear multiple times in a single optimization step. For prompt-level curriculum methods that follow the standard bandit pattern, this is straightforward: the scheduler can select the top-$B$ prompts under the current acquisition score, or sample a batch of distinct prompt-level arms directly.

For BMC, without-replacement sampling is slightly more subtle because sampling decisions are made through hierarchical descent. Each element of the batch is selected by traversing the Latent Task Tree, rather than by independently scoring all prompt-level arms. A naive implementation is to let each batch element descend the tree independently, check for duplicate prompts after the batch is formed, and resample any duplicates from the root until all prompts are unique. We refer to this as the rejection-based implementation.

Although this approach is simple and often sufficient, it has two drawbacks. First, resampled prompts are selected using a tree whose beliefs still include prompts that have already been assigned to the current batch. Thus, the posterior quantities used during descent are not fully conditioned on the remaining available prompts. Second, when the sampler becomes highly concentrated, rejection sampling can lead to long duplicate-resolution loops, increasing wall-clock time. We observed this issue most clearly in BMC-T, where utility-aware sampling can concentrate probability mass on a small set of relevant regions.

To avoid this issue, we use a batch-aware variant of hierarchical descent. Batch elements are selected sequentially. After a prompt is selected, it is marked as unavailable for the remainder of the batch, and the tree statistics used for subsequent descents are temporarily recomputed to exclude unavailable prompts. Thus, each batch element descends a tree conditioned on the prompts already selected in the current batch. This preserves without-replacement sampling while avoiding repeated rejection loops.

Batch-aware descent also has a useful exploration effect. As high-value prompts in a region are exhausted within the current batch, subsequent descents are encouraged to select other available prompts or neighboring regions, rather than repeatedly attempting to select the same prompt. In this sense, batch awareness makes the within-batch sampling process better aligned with the hierarchical structure used by BMC.

Unless otherwise stated, we treat batch-aware descent as the default implementation of BMC. The main BMC experiments used the rejection-based implementation because duplicate conflicts were rare and the simpler procedure was sufficient. The two implementations differ only in how without-replacement constraints are enforced within a batch; the underlying posterior updates, tree structure, and acquisition rules are unchanged.

\subsection{Saturating Estimation Errors}
\label{app:saturate}

The non-stationary belief update functions mentioned in the main section are intended to make BMC responsive to non-stationarity without making belief updates overly reactive to individual noisy observations. Intuitively, we want the scheduler to revise its beliefs when observations contradict prior expectations, but not to ``panic'' in response to a single surprising estimation error.

The squared surprise term (\hyperref[eq:lambda_update]{Equation~\ref*{eq:lambda_update}}):
\[
\lambda_i^{\mathrm{eff}} = \lambda_i^{(t-1)} \cdot \exp\!\left(- (s_i^{(t)})^2 \right)
\]
provides a simple symmetric measure of mismatch. Since both unexpectedly high and unexpectedly low learning signals indicate that the current belief may be stale, the update should depend on the magnitude of the deviation rather than its sign. Squaring the normalized surprise accomplishes this while increasing sensitivity to larger deviations.

For the belief (posterior mean) update (\hyperref[eq:belief_update]{Equation~\ref*{eq:belief_update}}):
\[
\mu_i^{(t)} = \frac{\lambda_i^{\mathrm{eff}} \cdot \mu_i^{(t-1)} + y_i^{(t)}}{\lambda_i^{\mathrm{eff}} + 1}
\]

the exponential decay applied to the effective prior weight implements soft forgetting. For small surprises, the decay is mild, so accumulated evidence is mostly preserved. For large surprises, the effective prior weight decreases rapidly, allowing the belief to adapt to potential model drift. This gives the update a controlled reset-like behavior without requiring a hard threshold or discrete change-point detector.

For the uncertainty (posterior variance) update (\hyperref[eq:uncertainty_update]{Equation~\ref*{eq:uncertainty_update}}):
\[
\sigma_i^{2\,(t)} =
\underbrace{
\sigma_i^{2\,(t-1)}
\frac{\lambda_i^{\mathrm{eff}}}{\lambda_i^{\mathrm{eff}} + 1}
}_{\substack{\text{confidence}\\[-1pt]\text{contraction}}}
+
\underbrace{
\frac{\log\!\left(1 + (s_i^{(t)})^2 + \mathrm{staleness}_i\right)}
{\lambda_i^{\mathrm{eff}} + 1}
}_{\substack{\text{uncertainty}\\[-1pt]\text{injection}}}
\]

the logarithmic uncertainty term serves the complementary purpose of injecting uncertainty without allowing it to grow explosively. Large surprises and stale observations should increase uncertainty, since they suggest the current estimate may no longer be reliable. However, a linear  uncertainty penalty could over-amplify noisy measurements and destabilize sampling. The logarithm provides diminishing growth: uncertainty increases when mismatch is detected, but the increase saturates for very large deviations.

Together, the exponential and logarithmic terms implement a conservative adaptation rule: surprising observations can reduce confidence in old evidence and encourage re-exploration, but their effect is bounded and smooth. These choices are practical heuristics rather than claims of Bayesian optimality. Alternative saturating functions could be used, but we found this parameterization to be simple, coefficient-light, and effective in our structured bandit setting.

\subsection{Staleness \& Uncertainty Growth}
\label{app:uncertainty-growth}
In addition to local surprise from newly observed rewards, BMC increases uncertainty for prompts that have not been sampled recently. This mechanism is intended to prevent uncertainty estimates from collapsing permanently for prompts whose learning signal may have changed due to policy drift.

Let $B$ denote the batch size and $N$ denote the number of available arms. We define the batch coverage coefficient as
\[
c = \frac{B}{N}.
\]
This coefficient measures the fraction of arms that can be sampled in a single batch. If $c$ is small, then many arms will naturally remain unsampled for long periods of time, so staleness should accumulate more slowly. Conversely, if $c$ is large, then failing to sample an arm for several steps is stronger evidence that its belief may be stale relative to the current policy.

For each arm $i$, we maintain an age counter $a_i$ measuring the number of scheduler steps since the arm was last sampled. We define the coverage-adjusted staleness term as
\[
\mathrm{staleness}_i = c \cdot a_i.
\]
The resulting uncertainty injection is then computed as
\[
\Delta \sigma_i^2
=
\frac{
\log\!\left(1 + (s_i^{(t)})^2 + \mathrm{staleness}_i\right)
}{
\lambda_i^{\mathrm{new}}
},
\]
where $\lambda_i^{\mathrm{new}}$ denotes the updated effective evidence weight.

This design makes staleness contribute to uncertainty growth while accounting for the sampling budget. The coverage factor prevents uncertainty from saturating too quickly in settings with many arms and small batches, where long gaps between observations are expected. The logarithmic transformation further ensures diminishing growth, so uncertainty increases for stale arms without becoming dominated by extremely large staleness values.

\subsection{Tree Diagnostics}
\label{app:treediag}

To assess differences between sampling strategies (specifically, strategies that rely on non-uniform sampling), we devise two diagnostics: \emph{rarity-weighted coverage} and \emph{structure gain}.

\subsubsection{Rarity-Weighted Exposure}
\label{app:rarity-weighted-exposure}

To evaluate whether a scheduler explores underrepresented regions of the latent task tree, we compute a rarity-weighted exposure metric over penultimate clusters. Let $\mathcal{C}=\{1,\dots,K\}$ denote the set of penultimate clusters, and let $n_c^{\mathrm{data}}$ be the number of training prompts assigned to cluster $c$. The empirical dataset prior over clusters is
\[
p_{\mathrm{data}}(c)
=
\frac{n_c^{\mathrm{data}}}{\sum_{c'=1}^K n_{c'}^{\mathrm{data}}}.
\]

We assign each cluster a rank-based rarity weight. Let $\mathrm{rank}(c)$ be the rank of cluster $c$ when clusters are sorted in ascending order by $p_{\mathrm{data}}(c)$, so that the rarest cluster has rank $0$ and the most common cluster has rank $K-1$. We define
\[
w_c
=
1 - \frac{\mathrm{rank}(c)}{K-1},
\]
for $K>1$, and $w_c=1$ when $K=1$. Thus, rarer clusters receive larger weights, while the most common clusters receive smaller weights. Because the weights depend only on rank rather than raw inverse frequency, the metric is less sensitive to outlier cluster sizes.

At training step $t$, let $n_c^{(t)}$ denote the number of selected prompts in the batch that belong to cluster $c$. The scheduler-induced exposure distribution over clusters is
\[
p_t(c)
=
\frac{n_c^{(t)}}{\sum_{c'=1}^K n_{c'}^{(t)}}.
\]
The rarity-weighted exposure at step $t$ is then
\[
R_t
=
\sum_{c=1}^K p_t(c) w_c.
\]
This metric is high when the scheduler allocates more batch mass to clusters that are rare under the dataset distribution.

For visualization, we normalize this quantity relative to the expected rarity exposure under the dataset prior. Let
\[
R_{\mathrm{data}}
=
\sum_{c=1}^K p_{\mathrm{data}}(c) w_c,
\qquad
R_{\max}
=
\max_{c} w_c.
\]
We report the normalized rarity-weighted exposure
\[
\widetilde{R}_t
=
\mathrm{clip}
\left(
\frac{R_t - R_{\mathrm{data}}}{R_{\max} - R_{\mathrm{data}}},
0, 1
\right).
\]
Thus, $\widetilde{R}_t=0$ corresponds to exposure no greater than what would be expected under the dataset prior, while larger values indicate disproportionate exposure to rarer clusters.

We emphasize that rarity-weighted exposure is a diagnostic metric, not an objective that BMC directly optimizes. Directly maximizing this quantity would be degenerate: a scheduler could obtain a high score by allocating all samples to the rarest penultimate cluster, regardless of whether those prompts provide useful learning signal. Instead, we use rarity-weighted exposure to characterize sampling behavior. In particular, it helps distinguish BMC from unstructured Thompson Sampling by showing whether the scheduler can prioritize high-learning-signal prompts while still allocating exposure to underrepresented regions of the task space.

\subsubsection{Structure Gain}
\label{app:structure-gain}

To quantify whether the latent task tree captures meaningful structure in the problem space, we measure how much variation in a prompt-level statistic (belief about learning signal) is explained by the penultimate clusters. Let $y_i^{(t)}$ denote a prompt-level scalar at training step $t$, such as the current estimated learning signal for prompt $i$. Let $c_i \in \{1,\dots,K\}$ denote the penultimate cluster assignment for prompt $i$.

We compute the fraction of variance in $y^{(t)}$ explained by the cluster partition:
\[
R^2_{\mathrm{tree}}(t)
=
1 -
\frac{
\sum_{k=1}^K \sum_{i:c_i=k}
\left(y_i^{(t)} - \bar y_k^{(t)}\right)^2
}{
\sum_{i=1}^N
\left(y_i^{(t)} - \bar y^{(t)}\right)^2
},
\]
where
\[
\bar y_k^{(t)}
=
\frac{1}{|\{i:c_i=k\}|}
\sum_{i:c_i=k} y_i^{(t)}
\]
is the mean value within cluster $k$, and $\bar y^{(t)}$ is the global mean across prompts. This quantity measures the proportion of total variance explained by the tree partition. Higher values indicate that prompts within the same penultimate cluster have more similar learning-signal estimates than prompts in different clusters.

Because larger or more imbalanced clusters can affect explained variance, we compare the tree partition against a size-matched random baseline. Specifically, we generate random partitions by permuting the cluster labels across prompts, preserving the exact cluster-size distribution while destroying the relationship between prompts and clusters. For each random partition $m$, we compute the same explained variance score $R^2_{\mathrm{rand},m}(t)$.

We define the structure gain as the ratio between the explained variance of the true tree partition and the average explained variance of these size-matched random partitions:
\[
G_{\mathrm{struct}}(t)
=
\frac{
R^2_{\mathrm{tree}}(t)
}{
\frac{1}{M}
\sum_{m=1}^M
R^2_{\mathrm{rand},m}(t)
}.
\]
Values greater than one indicate that the latent task tree explains more variation in the prompt-level statistic than expected from a random partition with the same cluster sizes. Because the denominator can be small, the ratio should be interpreted as a relative alignment diagnostic rather than as a claim that the tree fully explains the learning signal. In particular, a large structure gain indicates that the tree partition captures substantially more learning-signal variation than size-matched random groupings. In the 8B setting, the true partition explains an average of $R^2 = 0.00875$ of prompt-level learning-signal variation, indicating that the effect is modest in absolute terms but substantially enriched relative to the random baseline.

\subsection{BMC-T}
\label{app:bmct}
\subsubsection{Implementation}

BMC-T extends BMC by adding a target-utility term to hierarchical sampling. Let
$\mathcal{D}_{\mathrm{train}} = \{x_i\}_{i=1}^{N}$ denote the sampleable training problems, and let
$\mathcal{D}_{\mathrm{target}} = \{z_j\}_{j=1}^{M}$ denote an arbitrary target distribution, such as a held-out evaluation set, validation set, or collection of problems representing a desired capability. Target examples are used only to define utilities over regions of the Latent Task Tree; they are never included in the set of arms available to the bandit scheduler. Thus, BMC-T may use target examples to estimate which latent regions are relevant, but it can only sample problems from $\mathcal{D}_{\mathrm{train}}$.

We construct a Latent Task Tree using both $\mathcal{D}_{\mathrm{train}}$ and $\mathcal{D}_{\mathrm{target}}$. Let $\mathcal{T}$ denote the resulting tree. The leaves of $\mathcal{T}$ correspond to individual examples, while internal nodes correspond to latent task regions. For a node $v \in \mathcal{T}$, let $\mathrm{child}(v)$ denote its children and let $\mathcal{L}(v)$ denote the set of examples contained in the subtree rooted at $v$.

For each node $v$, we compute
\begin{align*}
    n_{\mathrm{train}}(v)
    &= \sum_{x \in \mathcal{L}(v)} \mathbf{1}\{x \in \mathcal{D}_{\mathrm{train}}\}, \\
    n_{\mathrm{target}}(v)
    &= \sum_{x \in \mathcal{L}(v)} \mathbf{1}\{x \in \mathcal{D}_{\mathrm{target}}\}, \\
    w_{\mathrm{target}}(v)
    &= \sum_{x \in \mathcal{L}(v)} \omega(x)\mathbf{1}\{x \in \mathcal{D}_{\mathrm{target}}\},
\end{align*}
where $\omega(x) \geq 0$ is an optional target-example weight. In our experiments, these weights can be used to normalize across target sources or to focus on a particular target subset. If no weighting is needed, we use uniform target weights.

A node is sampleable only if it contains at least one training example:
\[
    n_{\mathrm{train}}(v) > 0.
\]
Nodes with $n_{\mathrm{train}}(v)=0$ may still contain target examples, but they are excluded from the scheduler because there are no training problems available to sample from that region.

For each internal node $v$, utilities are defined locally over its sibling set $\mathrm{child}(v)$. Let
\begin{align*}
    N_{\mathrm{train}}(v)
    &= \sum_{c \in \mathrm{child}(v)} n_{\mathrm{train}}(c), \\
    N_{\mathrm{target}}(v)
    &= \sum_{c \in \mathrm{child}(v)} n_{\mathrm{target}}(c), \\
    W_{\mathrm{target}}(v)
    &= \sum_{c \in \mathrm{child}(v)} w_{\mathrm{target}}(c).
\end{align*}
If $N_{\mathrm{target}}(v)=0$ or $W_{\mathrm{target}}(v)=0$, then the children of $v$ receive zero utility:
\[
    u(c) = 0
    \quad \forall c \in \mathrm{child}(v).
\]
Otherwise, we estimate the local target mass assigned to each child. Let $K_v = |\mathrm{child}(v)|$ and define
\[
    \alpha_v = \rho \frac{W_{\mathrm{target}}(v)}{K_v},
\]
where $\rho > 0$ is a smoothing coefficient. The child-level target mass is
\[
    p_{\mathrm{target}}(c \mid v)
    =
    \frac{
        w_{\mathrm{target}}(c) + \alpha_v
    }{
        W_{\mathrm{target}}(v) + K_v \alpha_v
    }.
\]
Continuing the Bayesian interpretation used throughout BMC, this additive smoothing can be viewed as placing a symmetric Dirichlet prior with concentration $\alpha_v$ on the categorical distribution over children. Intuitively, $p_{\mathrm{target}}(c \mid v)$ is not meant to represent a true data-generating probability; it is a normalized local estimate of how the observed target mass under parent $v$ is distributed across its children. This makes target relevance \emph{comparable} within each sibling set during top-down descent, while smoothing prevents finite target samples from assigning overly brittle zero-utility estimates to nearby regions.

We additionally apply a train-support gate to avoid assigning large utility to regions that contain target examples but little sampleable training support. Let
\[
    \lambda_v = \eta N_{\mathrm{train}}(v),
\]
where $\eta$ is the same HDBSCAN minimum-cluster fraction hyperparameter used in constructing the Latent Task Tree (see \hyperref[app:hyperparameters]{Appendix~\ref*{app:hyperparameters}}). We define
\[
    g(c \mid v)
    =
    \frac{
        n_{\mathrm{train}}(c)
    }{
        n_{\mathrm{train}}(c) + \lambda_v
    }.
\]
Intuitively, this gate downweights children whose target overlap is not supported by enough sampleable training examples. Using the same minimum-cluster fraction $\eta$ ties this notion of support to the scale used when constructing the Latent Task Tree: a child receives high utility only when it contains a nontrivial amount of trainable mass relative to the local parent region.

The final utility assigned to child $c$ is then
\[
    u(c)
    =
    p_{\mathrm{target}}(c \mid v) \cdot g(c \mid v).
\]

During hierarchical Thompson Sampling, standard BMC samples a productivity score for each child using the Gaussian belief maintained at that node. For child $c$ of parent $v$, the standard BMC score is
\[
    s_{\mathrm{BMC}}(c)
    \sim
    \mathcal{N}\left(
        \mu_c,\,
        \sigma_c^2 + \tau_v
    \right),
\]
where $\mu_c$ and $\sigma_c^2$ are the child belief parameters, and $\tau_v$ is the parent-level heterogeneity term used by BMC.

BMC-T modifies this score by adding the target-utility bonus. Since utilities are computed locally within each sibling set, we normalize by the maximum child utility under the current parent:
\[
    \bar{u}(c)
    =
    \frac{
        u(c)
    }{
        \max_{c' \in \mathrm{child}(v)} u(c') + \epsilon
    },
\]
where $\epsilon$ is a small constant for numerical stability. The utility-aware score is
\[
    s_{\mathrm{BMC\text{-}T}}(c)
    =
    s_{\mathrm{BMC}}(c)
    +
    \gamma r_{\max} \bar{u}(c).
\]
Here, $\gamma \geq 0$ controls the strength of utility awareness, and $r_{\max}$ places the utility bonus on the same reward scale as the productivity model. Setting $\gamma=0$ removes the utility bonus, but does not necessarily recover standard BMC because target examples may still affect the topology of the Latent Task Tree. Larger values of $\gamma$ increasingly steer the scheduler toward regions with high target overlap.

At each internal node, BMC-T selects the child with the largest utility-aware score:
\[
    c^\star
    =
    \arg\max_{c \in \mathrm{child}(v)}
    s_{\mathrm{BMC\text{-}T}}(c),
\]
subject to the constraint that $c$ has at least one available training example in its subtree. The sampler then descends recursively until reaching a sampleable training problem.

\subsubsection{Alternative Implementations}

The BMC-T implementation above is one concrete way to incorporate utility into hierarchical curriculum learning. It estimates target relevance from overlap between training and target examples within the Latent Task Tree, then adds a normalized utility bonus to the productivity score used during top-down sampling. This design keeps the roles of productivity and utility separated: productivity estimates where the current policy can learn, while utility estimates which learnable regions are relevant to a target distribution.

However, the broader idea is not tied to this particular utility estimator or additive scoring rule. Utility could instead enter multiplicatively, modulating the productivity score so that target relevance acts as a gate rather than a bonus. Utility could also be estimated from other signals, such as gradient-based alignment \citep{yang2026gradaligngradientaligneddataselection} or optimizer-based data valuation \citep{wang2026opusefficientprincipleddata}. These alternatives may be useful when target examples are sparse, unavailable, or difficult to combine directly with the training set in a shared tree.

Thus, BMC-T should be interpreted as a proof of concept for utility-aware curriculum design rather than a unique prescription. Its purpose is to show that evaluation relevance can be explicitly represented alongside productivity and diversity, and that curriculum decisions can be made over target-relevant regions of the task manifold rather than over difficulty-based metrics alone.

\clearpage % end section
%%%%%%%%%%%%%%%%%%%%%%%%%%%%%%%%%%%%%%%%%%%%%%%%%%%%%%%%%%%%%%%%%%%%%%%%%%%%%%%%%%%%%%%%%%%%%%%%%%%%%%%%%%%%%%%%%%%%%
\end{document}